\documentclass[sn-mathphys]{sn-jnl}

\jyear{2021}%

\theoremstyle{thmstyleone}%
\theoremstyle{thmstyletwo}%

\theoremstyle{thmstylethree}%

\usepackage{float}
\usepackage{tabularx}
\usepackage{caption}
\usepackage{subcaption}
\usepackage{natbib}[authoryear]

\usepackage{pdflscape}
\usepackage{afterpage}
\usepackage{xcolor, soul}
\sethlcolor{red}

\newcommand{\graphwidth}{0.8}
\newcommand{\confusionwidth}{0.3}
\newcommand{\histogramwidth}{0.65}

\begin{document}

\title[An Analysis of Discretization Methods for Comm. Learning with MARL]{An In-Depth Analysis of Discretization Methods for Communication Learning using Backpropagation with Multi-Agent Reinforcement Learning}


\author*[1]{\fnm{Astrid} \sur{Vanneste}}\email{astrid.vanneste@uantwerp.be}

\author*[1]{\fnm{Simon} \sur{Vanneste}}\email{simon.vanneste@uantwerp.be}

\author[2]{\fnm{Kevin} \sur{Mets}}\email{kevin.mets@uantwerp.be}

\author[2]{\fnm{Tom} \sur{De Schepper}}\email{tom.deschepper@uantwerp.be}

\author[1]{\fnm{Siegfried} \sur{Mercelis}}\email{siegfried.mercelis@uantwerp.be}

\author[3]{\fnm{Peter} \sur{Hellinckx}}\email{peter.hellinckx@uantwerp.be}

\affil[1]{\orgdiv{IDLab - Faculty of Applied Engineering}, \orgname{University of Antwerp - imec}, \orgaddress{\street{Sint-Pietersvliet 7}, \city{Antwerp}, \postcode{2000}, \country{Belgium}}}

\affil[2]{\orgdiv{IDLab - Department of Computer Science}, \orgname{University of Antwerp - imec}, \orgaddress{\street{Sint-Pietersvliet 7}, \city{Antwerp}, \postcode{2000}, \country{Belgium}}}

\affil[3]{\orgdiv{IDLab - Faculty of Applied Engineering}, \orgname{University of Antwerp}, \orgaddress{\street{Groenenborgerlaan 171}, \city{Antwerp}, \postcode{2020}, \country{Belgium}}}

\abstract{Communication is crucial in multi-agent reinforcement learning when agents are not able to observe the full state of the environment. The most common approach to allow learned communication between agents is the use of a differentiable communication channel that allows gradients to flow between agents as a form of feedback. However, this is challenging when we want to use discrete messages to reduce the message size, since gradients cannot flow through a discrete communication channel. Previous work proposed methods to deal with this problem. However, these methods are tested in different communication learning architectures and environments, making it hard to compare them. In this paper, we compare several state-of-the-art discretization methods as well as a novel approach. We do this comparison in the context of communication learning using gradients from other agents and perform tests on several environments. In addition, we present COMA-DIAL, a communication learning approach based on DIAL and COMA extended with learning rate scaling and adapted exploration. Using COMA-DIAL allows us to perform experiments on more complex environments.
Our results show that the novel ST-DRU method, proposed in this paper, achieves the best results out of all discretization methods across the different environments. It achieves the best or close to the best performance in each of the experiments and is the only method that does not fail on any of the tested environments.
}

\keywords{Communication Learning, Multi-Agent, Reinforcement Learning}



\maketitle

\section{Introduction}
Over the past several years, both single-agent reinforcement learning (RL) \cite{sutton1998introduction, mnih2015human, schrittwieser2020, fawzi2022} and multi-agent RL (MARL) \cite{busoniu2008, gronauer2022} have gained a lot of interest. In (MA)RL, the agents often have to deal with partial observability \cite{hausknecht2015deep, oliehoek2016concise}. The agents can only observe part of the global state, making it hard to choose appropriate actions. For example, an autonomous car cannot look around corners to see oncoming traffic. In MARL, this partial observability can often be alleviated by allowing the agents to share information with each other. By combining this information with their own observation, agents get a more complete view of the environment and can choose better actions \cite{tan1993multi, melo2012, sukhbaatar2016learning, foerster2016learning}. For example, cars can share the information they observe to prevent dangerous situations. 

One of the subfields within MARL is research towards learned communication between agents. The most commonly used approach thus far is allowing gradients to flow between agents as a form of feedback on the received messages. However, in the case of discrete communication messages this raises a problem since gradients cannot flow through a discrete communication channel. Several different approaches have been proposed in the state of the art to discretize messages while allowing gradients to flow through the discretization unit \cite{foerster2016learning, lowe2020multiagent, mordatch2018, lin2021learning}. Each of these methods was tested using different communication learning approaches and applied on different environments, making a fair comparison very difficult. 

Our contributions consist of three parts. First, we present an in-depth comparison of several discretization methods used in the state of the art. In our comparison we focus on using these discretization methods to allow discrete communication when learning communication using the gradients of the receiving agents. We compare each of the approaches on several environments with increasing complexity as well as analyze their performance when the environment introduces errors to the communication messages. In addition, we present one novel discretization method called ST-DRU. Finally, to enable us to compare the different discretization methods in more complex environments, we present COMA-DIAL. COMA-DIAL is a communication learning method based on DIAL \cite{foerster2016learning} and COMA \cite{foerster2018counterfactual}. We further improve the performance of COMA-DIAL by using an explicit exploration strategy and applying learning rate scaling on the actor of the agents.

The remainder of this paper is structured as follows. Section \ref{sec:related_work} gives an overview of work related to our research. Section \ref{sec:background} contains some additional background information. Section \ref{sec:methods} provides a detailed explanation of the discretization methods we compare in this paper and COMA-DIAL. In Section \ref{sec:experiments}, the different experiments that were performed are explained along with their results. We discuss the results of our experiments further in Secion \ref{sec:discussion}. In Section \ref{sec:conclusion} we draw some conclusions from our experimental results.

\section{Related Work}
\label{sec:related_work}
In this section, we review state-of-the-art work relevant to our research. We give an overview of several communication learning methods, focusing on methods that learn discrete communication. Here, we see some alternative approaches for learning discrete communication beside using a differentiable communication channel as well as different discretization techniques used in the state of the art. 

Foerster et al.\cite{foerster2016learning} and Sukhbaatar et al.\cite{sukhbaatar2016learning} proposed the first successful methods for learning inter-agent communication. Foerster et al.\cite{foerster2016learning} proposed two novel approaches, Reinforced Inter-Agent Learning (RIAL) and Differentiable Inter-Agent Learning (DIAL). Both RIAL and DIAL use discrete communication messages, but learn the communication policy in a different way. RIAL learns the communication policy the same way as learning the action policy, by using the team reward. However, the results clearly show that this is not sufficient in most environments. DIAL proved more successful by using gradients originating from the agents receiving the messages which provide feedback on the communication policy. Sukhbaatar et al.\cite{sukhbaatar2016learning} proposed a different approach called CommNet. Messages consist of the hidden state of the agents, resulting in continuous messages. Similar to DIAL, CommNet uses gradients that flow through the communication channel to train the communication. 

A lot of the research that follows these works chooses to use continuous communication like Sukhbaatar et al.\cite{sukhbaatar2016learning}, avoiding the problem of discretizing the communication messages \cite{simoes2020a3c3}. 
Other methods choose a different method to learn communication than using gradients through the communication channel. This also avoids the challenge of discretizing the messages.
Jaques et al.\cite{jaques2019socialinfluence} train the communication policy using the team reward augmented with a social influence reward. This additional reward is based on how much the message changes the action policy of the receiving agents. 
Vanneste et al.\cite{vanneste2021learning} use counterfactual reasoning to directly learn a communication protocol without the need of a differentiable communication channel. 
Freed et al.\cite{Freed_Sartoretti_Hu_Choset_2020} use a randomized encoder at the sender to encode the continuous messages into discrete messages. At the receiver, a randomized decoder is used to approximate the original continuous message. They show that by using this technique they can consider the communication channel equivalent to a continuous channel with additive noise, allowing gradients to flow between the sender and receiver agent.

Lowe et al.\cite{lowe2020multiagent} and Mordatch and Abbeel\cite{mordatch2018} propose Multi-Agent Deep Deterministic Policy Gradients (MADDPG). In their work they evaluate MADDPG on multiple different scenarios, including communication tasks. They do not use a differentiable communication channel to learn communication but they have to make sure the messages are differentiable to allow the MADDPG method to work properly since policies are learned using gradients that originate from the critic. They allow discrete communication by using a gumbel softmax \cite{jang2017categorical, maddison2017concrete}. 
Lin et al.\cite{lin2021learning} use an autoencoder at the sender to compose a representation of the observation that will be used as communication message. To discretize these messages they use a straight through estimator \cite{bengio2013estimating, yin2019understanding} in the autoencoder. 
Both Lowe et al.\cite{lowe2020multiagent}, Mordatch and Abbeel\cite{mordatch2018} and Lin et al.\cite{lin2021learning} have to use differentiable discretization techniques in their methods to allow discrete communication. However, they do not use the techniques in the same way we do in our work. Lowe et al.\cite{lowe2020multiagent} and Mordatch and Abbeel\cite{mordatch2018} use the discretization method in a similar way as our work but in MADDPG the gradients that correct the communication originate from the critic instead of from other agents. Lin et al.\cite{lin2021learning} use the discretization method in a very different way since they train the communication policy entirely using the reconstruction loss of the autoencoder instead of the gradients from the other agents.  

Havrylov and Titov \cite{havrylov2017} investigate the emergence of language in a slightly different context. They have two agents, one sender and one receiver. The sender sends a series of words to the receiver to describe the image the sender observed. The receiver has to use this information to identify the described picture from a set of options. They describe two ways to learn the policy of the sending agent. The first option is to use REINFORCE and the second option is to use the gradients originating from the receiver agent. In the latter case they use the straight through gumbel softmax \cite{jang2017categorical, maddison2017concrete} to discretize the messages. However, their approach is different than ours because the receiver is based on Imaginet \cite{chrupala2015} which learns using the mean square error of the agent prediction and the target answer instead of an RL approach. Carmeli et al. \cite{carmeli2022} use a similar approach, using cross entropy loss between the prediction and target answer. They analyze the difference between continuous communication, quantization with gumbel softmax \cite{jang2017categorical, maddison2017concrete} and with a straight through estimator \cite{bengio2013estimating, yin2019understanding}. In their experiments the straight trough estimator proved to be the best choice. 

Summarized, in the state-of-the-art related to our research, we see that multiple discretization methods have been proposed as can be seen in Table \ref{tab:sota}. But, differences in communication learning approaches and the fact that each of these methods is tested on different environments makes comparing these different discretization methods very hard. 

\afterpage{%
    \clearpage
    \thispagestyle{empty}
    \begin{landscape}
        \begin{table*}[t]
            \centering
            \caption{Comparison of the state of the art and our work \\  {\normalfont (DRU - Discretize Regulize Unit, GS - Gumbel Softmax, STE - Straight Through Estimator, ST-DRU - Straight Through DRU, ST-GS - Straight Through GS)}}
            \label{tab:sota}
            \begin{tabularx}{0.8\paperheight}{llXX}
                    \toprule
                                                                                                & Message Type          & Communication Learning Technique                                          & Discretization Method                     \\
                    \midrule
                    RIAL \cite{foerster2016learning}                                            &  Discrete             & DQN using team reward                                                     &  Discrete policy                          \\
                    DIAL \cite{foerster2016learning}                                            &  Discrete             & Gradients from other agents                                               &  DRU                                      \\
                    CommNet \cite{sukhbaatar2016learning}, A3C3 \cite{simoes2020a3c3}           &  Continuous           & Gradients from other agents                                               &  N/A                                      \\
                    Havrylov and Titov \cite{havrylov2017}                                      &  Discrete             & Gradients from other agents                                               &  Discrete policy / ST-GS                  \\
                    Carmeli et al. \cite{carmeli2022}                                           &  Continuous/Discrete  & Gradients from other agents                                               &  GS / STE                                 \\
                    MADDPG \cite{lowe2020multiagent, mordatch2018}                              &  Continuous/Discrete  & Gradients from the critic                                                 &  GS                                       \\
                    Freed et al\cite{Freed_Sartoretti_Hu_Choset_2020}                           &  Discrete             & Gradients from other agents                                               &  Randomized Encoder/Decoder               \\
                    Jaques et al.\cite{jaques2019socialinfluence}                               &  Discrete             & A3C using team reward augmented with social influence reward              &  Discrete policy                          \\
                    MACC \cite{vanneste2021learning}                                            &  Discrete             & Counterfactual reasoning                                                  &  Discrete policy                          \\
                    Lin et al.\cite{lin2021learning}                                            &  Discrete             & Reconstruction loss                                                       &  STE                                      \\
                    \textbf{This Work}                                                          &  \textbf{Discrete}    & \textbf{Gradients from other agents}                                      &  \textbf{(ST)-DRU / STE / (ST)-GS}        \\
                    \bottomrule
            \end{tabularx}
        \end{table*}
    \end{landscape}
    \clearpage
}

\section{Background}
\label{sec:background}

\subsection{Deep Q-Networks (DQN)}
In single agent RL \cite{sutton1998introduction}, the agent chooses an action $u_t \in U$ based on the state $s_t \in S$ of the environment. As a result of this action, the environment will transition to a new state $s_{t + 1}$ and provide the agent with a reward $r_{t+1} \in R$. This reward is used to train the agent. Q-learning uses this reward to calculate a Q-value for each state action pair $Q(s, u)$. This Q-value represents a value for each state action pair, where a higher Q-value indicates a better action. Therefore, the policy of our agent can be defined by Equation \ref{eq:qlearning_policy}.

\begin{equation}
    \begin{aligned}
        \pi(s) = \underset{u}{argmax} \left( Q(s, u) \right)
    \end{aligned}
    \label{eq:qlearning_policy}
\end{equation}

Deep Q-learning \cite{mnih2015human} uses a neural network with parameters $\theta$ to represent the Q-function. The deep Q-network is optimized at iteration $i$ by minimizing the loss in Equation \ref{eq:dqn_loss}.

\begin{equation}
    \begin{aligned}
        \mathcal{L}^i(\theta^i) = \mathbb{E}_{s_t,u_t,r_t,s_{t+1}} \big[ (r_t + \gamma \max_{u_{t}} Q(s_{t+1},u_{t+1}, \theta^{i-}) - Q(s_t,u_t, \theta^i))^2\big]
    \end{aligned}
    \label{eq:dqn_loss}
\end{equation}
where $\gamma$ is the discount factor and $\theta^{i-}$ are the parameters of the target network \cite{mnih2015human}. This target network will be updated after each training iteration according to Equation \ref{eq:target_network_update}.
\begin{equation}
    \begin{aligned}
        \theta^{(i + 1)-} \leftarrow \tau \theta^i + (1 - \tau) \theta^{i-}
    \end{aligned}
    \label{eq:target_network_update}
\end{equation}
where $\tau$ is a weight that indicates how fast the target network should follow the parameters $\theta$.
In our work, the agent does not receive the full state $s_t$ but only a limited observation $o_t \in O$ of this state. This increases the complexity since the observation might lack important information. 

\subsection{Differentiable Inter-Agent Learning (DIAL)}
\begin{figure*}[t]
    \centering
    \includegraphics[width=0.9\textwidth]{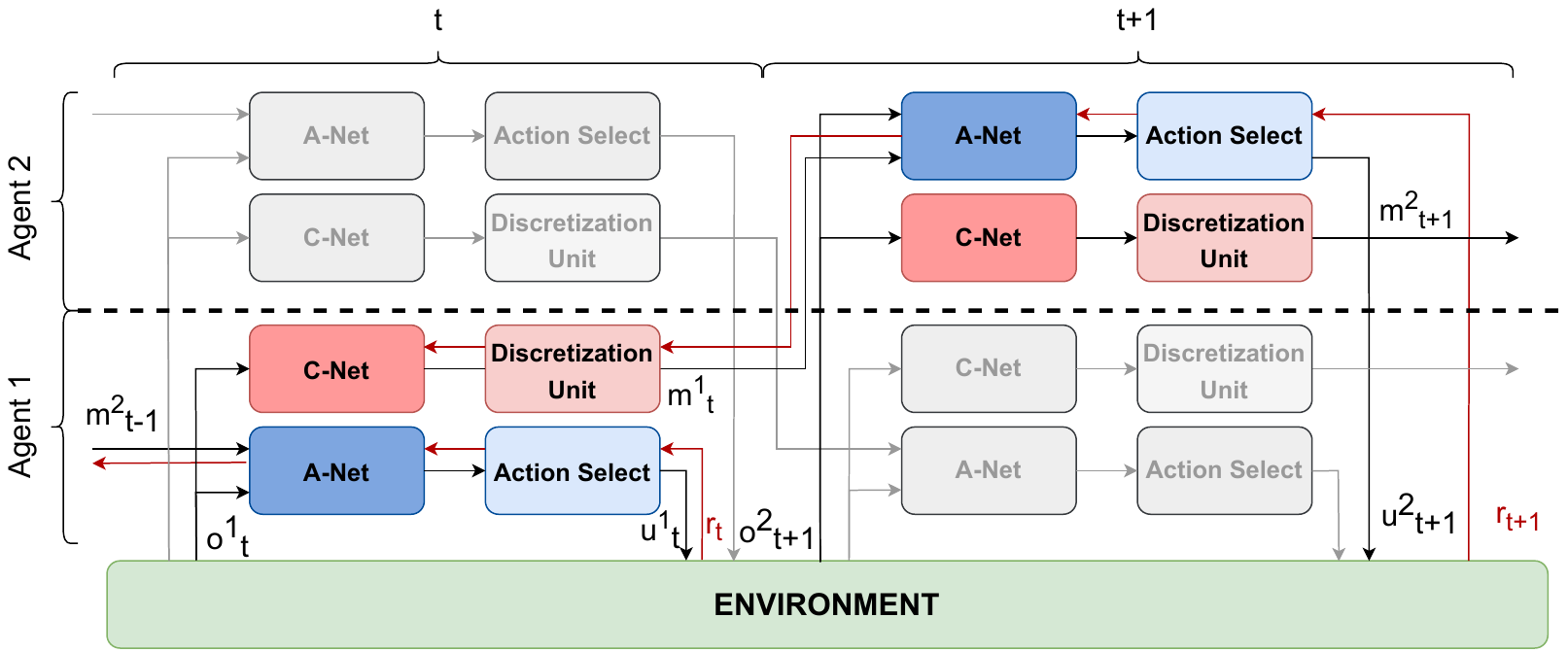}
    \caption{Architecture of DIAL with gradients indicated by the red arrows}
    \label{fig:DIAL}
\end{figure*}

To allow for a fair comparison, we will use the same communication learning approach for each of the different discretization methods. We use DIAL as proposed by Foerster et al.\cite{foerster2016learning} in our experiments since it is the most general and well known architecture to learn discrete communication using gradients from the other agents. The architecture of DIAL can be seen in Figure \ref{fig:DIAL}. We adapted the original DIAL architecture by separating the action and communication network. This allows us to keep the communication network small, making communication learning easier. In our experiments, we examine environments where the agents only need to share and encode part of the observation which allows us to make this adaptation. When the agents are expected to communicate about a strategy, splitting the action and communication network may no longer be possible.

Each agent consists of two networks, the A-Net and the C-Net. The A-Net produces Q-values to determine the action based on the observation and the incoming messages. The C-Net is responsible for calculating the messages based on the observation. It does not receive the incoming messages in our experiments because in these environments the communication policy does not need the incoming messages to determine the output message. Before the messages are broadcast to the other agents, the discretization unit applies one of the discretization techniques that we are comparing in this paper. To train the agents, we apply the team reward provided by the environment on the A-Net according to deep Q-learning. The gradients from the A-Net are propagated to the C-Net of all the agents that sent a message to that agent. This allows us to train the C-Net using the feedback of the agents receiving the messages. 

\subsection{Counterfactual Multi-Agent (COMA) Policy Gradients}
\label{sec:coma}
Foerster et al.\cite{foerster2018counterfactual} proposed a MARL approach that uses a centralized critic to train the policy of their agents. The centralized critic predicts the joint state-action utility $Q(s_t, u_t)$. The centralized critic used in COMA will be learned by minimizing the following loss function: 

\begin{equation}
    \begin{aligned}
        \mathcal{L}(\theta_{Q}^i) = \mathbb{E}_{s_t,u_t,r_t,s_{t+1}}[(r_t + \gamma Q(s_{t+1}, u_{t+1}, \theta_{Q}^{i-}) - Q(s_t, u_t, \theta_{Q}^i))^2] 
    \end{aligned}
    \label{eq:action_critic_loss}
\end{equation}

where $\theta_{Q}^i$ are the parameters of the critic at iteration $i$. Using counterfactual reasoning, this centralized critic can be used to calculate an advantage for each agent. This advantage indicates how good the chosen action is compared to the possible alternative actions while keeping the actions of the other agents fixed. This can be achieved by subtracting the expected utility of the action policy $V^a(s_t, u^{-a}_t)$ from the state-action utility $Q(s_t, u_t)$. The expected utility of the action policy can be calculated using the marginalization of the counterfactual actions multiplied with the joint state-action utility of that set of actions. Equation \ref{eq:action_advantage} and \ref{eq:action_advantage_v} show the detailed calculation of the advantage for COMA. The superscript $-a$ indicates all agents except agent $a$.

\begin{equation}
A^a(s_t, u_t) =  Q(s_t, u_t) - V^a(s_t, u^{-a}_t)
\label{eq:action_advantage}
\end{equation}

\begin{equation}
V^a(s_t, u^{-a}_t) = \sum_{u'^a_t} \Big(Q(s_t, (u'^a_t, u^{-a}_t)) \pi^a_u(u'^a_t \vert o^a_t) \Big)
\label{eq:action_advantage_v}
\end{equation}


\section{Methods}
\label{sec:methods}
In this section, we describe the different discretization units, that we will compare, and COMA-DIAL in more detail. Table \ref{tab:discretization_units} provides an overview of all the discretization methods and their differences. We show the difference between the function used to calculate the output of the discretization unit during training and during evaluation as well as the function that is used for the backward pass.

\subsection{Discretize Regularize Unit (DRU)}
\begin{figure*}
    \centering
    \includegraphics[width=\histogramwidth\linewidth, trim=65 10 80 35, clip]{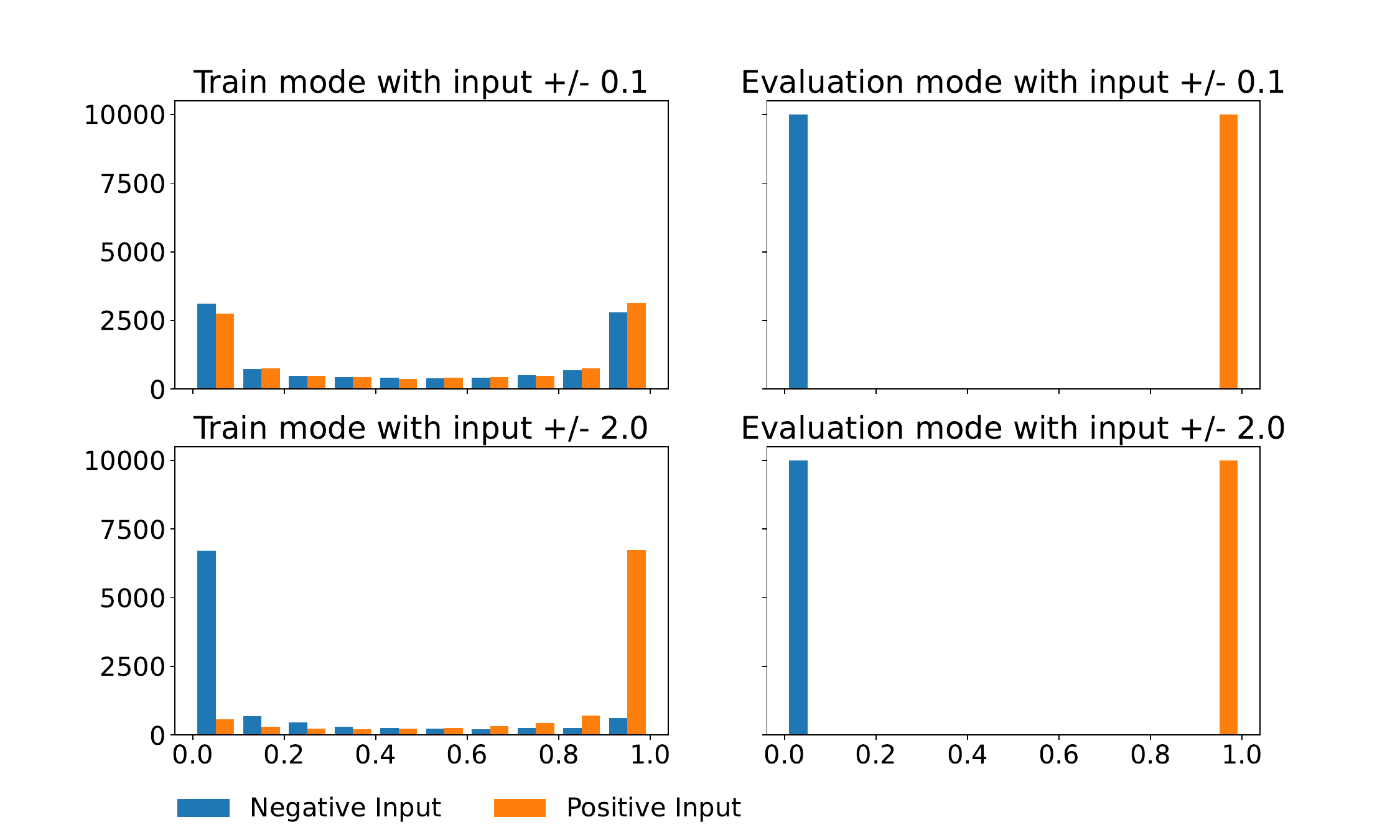}
    \caption{Histogram of the output of the DRU when calculating the output 10k times for different input values (positive and negative) for both training and evaluation mode}
    \label{fig:DRU_response}
\end{figure*}

In the DIAL method, Foerster et al.\cite{foerster2016learning} propose a module called the Discretize Regularize Unit (DRU) to allow gradients to be used for training while learning discrete communication messages. The DRU has two modes, discretization and regularization. The discretization mode is used at execution time and discretizes the input into a single bit using Equation \ref{eq:DRU_discretize}.
\begin{equation}
    \begin{aligned}
        m = H(x)
    \end{aligned}
    \label{eq:DRU_discretize}
\end{equation}
where $H(x)$ is a heaviside function and $x$ is the input of the discretization unit (output of the C-Net). This calculation cannot be used during training because the derivative of the heaviside function is the Dirac function which is zero everywhere except at $x=0$, where the output is infinite. Therefore, the regularization mode is used during training. When using the regularization mode, the agents are allowed to communicate using continuous messages. However, the DRU tries to encourage the communication policy to generate messages that can easily be discretized at execution time. This is achieved by applying Equation \ref{eq:DRU_regularize}. 
\begin{equation}
    \begin{aligned}
        m = \sigma(x + n)
    \end{aligned}
    \label{eq:DRU_regularize}
\end{equation}
where $x$ is the input of the discretization unit (output of the C-Net), $n$ is noise sampled from a Gaussian distribution with standard deviation $\sigma_G$ and $\sigma(x)$ is the sigmoid function. The noise will affect the output of the DRU the most when the input is around zero since the sigmoid is the steepest there. The influence will be much smaller for inputs with high absolute values. The output in those cases will also go towards zero and one, making it very similar to discrete, binary messages. This can be seen in Figure \ref{fig:DRU_response}.

\subsection{Straight Through Estimator (STE)}
A straight through estimator (STE) \cite{bengio2013estimating, yin2019understanding} performs a normal discretization, as in Equation \ref{eq:DRU_discretize}, when calculating the output. However, when performing backpropagation, it uses the gradients of an identity function instead of the gradients of the discretization. The advantage of this technique is that the agent receiving the message will immediately receive binary numbers and can learn how to react to these messages while still being able to use the gradients from the receiving agents to train the communication network. For the STE, the output will always look like the DRU in evaluation mode, shown in Figure \ref{fig:DRU_response}.

\subsection{Gumbel Softmax (GS)}
The Gumbel Softmax (GS) \cite{jang2017categorical, maddison2017concrete} is a method to approximate a sample from a categorical distribution in a differentiable way. Normal sampling techniques are not differentiable and therefore not directly applicable in this context. The GS achieves this desirable property by using Gumbel noise and the gumbel-max trick\cite{gumbel1954statistical}. Using the gumbel-max trick, we can sample from a categorical distribution with class probabilities $\pi$ as described in Equation \ref{eq:gumbel_max_trick}.
\begin{equation}
    \begin{aligned}
        z = one\_hot(\underset{i}{argmax}[g_i + log \pi_i]) 
    \end{aligned}
    \label{eq:gumbel_max_trick}
\end{equation}

\begin{figure*}
    \centering
    \includegraphics[width=\histogramwidth\linewidth, trim=65 10 80 35, clip]{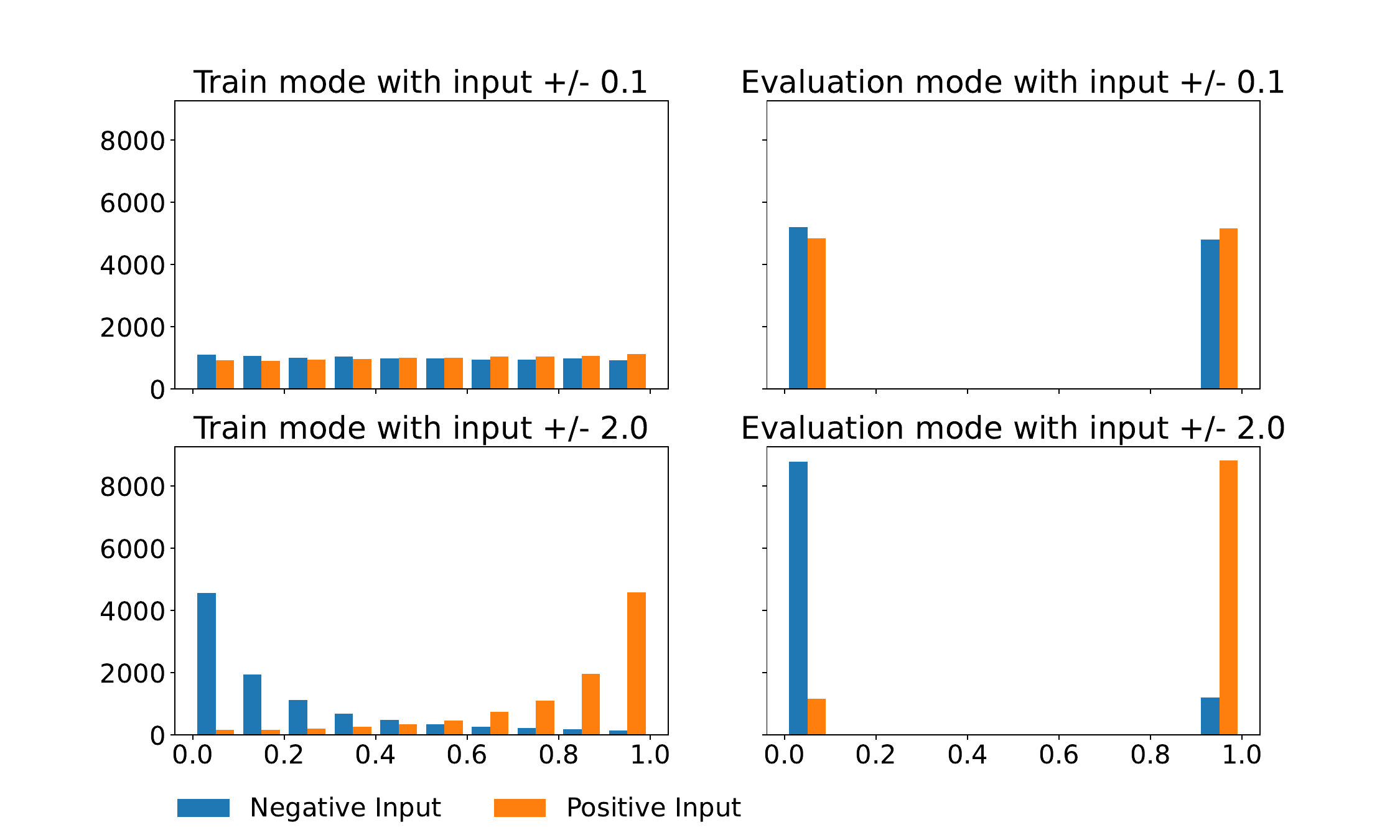}
    \caption{Histogram of the output of the GS when calculating the output 10k times for different input values (positive and negative) for both training and evaluation mode with temparature $\tau = 1.0$}
    \label{fig:GS_response}
\end{figure*}

where $g_1, ... g_k$ are i.i.d. samples drawn from Gumbel(0,1), $i \in \{0, 1\}$, $\pi_0 = \sigma(x)$, $\pi_1 = (1 - \sigma(x))$, $x$ is the input of the discretization unit (output of the C-Net) and $\sigma(x)$ is the sigmoid function. To make this differentiable, we have to approximate the $one\_hot$ and $argmax$ functions with a $softmax$ function. Since we need two probabilities to obtain a categorical distribution for both states of a bit we will obtain the output message by using Equation \ref{eq:gumbel_softmax}.
\begin{equation}
    \begin{aligned}
        m = softmax \left(\frac{log(\sigma(x)) + g_1}{\tau}, \frac{log(1-\sigma(x)) + g_2}{\tau} \right)[0]
    \end{aligned}
    \label{eq:gumbel_softmax}
\end{equation}
where $x$ is the input of the discretization unit (output of the C-Net), $\tau$ is the softmax temperature and $\sigma(x)$ is the sigmoid function. A low temperature will result in an output that closely matches the output of Equation \ref{eq:gumbel_max_trick}. The higher the temperature the more the output will approach a uniform distribution. Figure \ref{fig:GS_response} shows the behaviour of the GS for different inputs in training and evaluation mode with temparature $\tau = 1.0$. 

\subsection{ST-DRU}
\begin{figure*}
    \centering
    \includegraphics[width=\histogramwidth\linewidth, trim=65 10 80 35, clip]{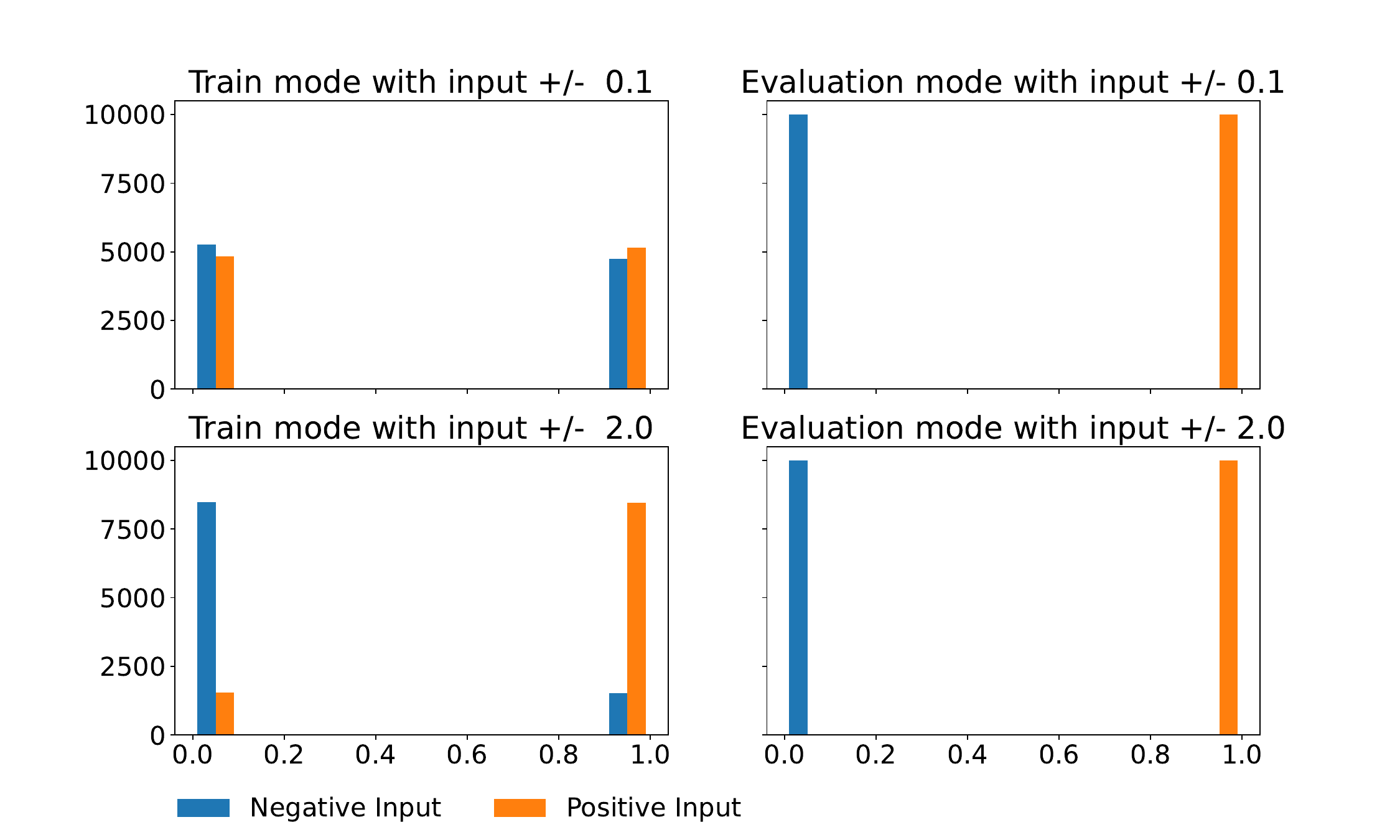}
    \caption{Histogram of the output of the ST-DRU when calculating the output 10k times for different input values (positive and negative) for both training and evaluation mode}
    \label{fig:ST-DRU_response}
\end{figure*}

We propose a novel discretization method (ST-DRU) that combines the DRU and STE methods. During execution, we discretize the input messages in the same way as mentioned in the DRU and STE methods, shown in Equation \ref{eq:DRU_discretize}. During training, we will use a different function for the forward and backward pass. In the forward pass, we add gaussian noise and apply the same discretization, resulting in Equation \ref{eq:STE_discretize}. 
\begin{equation}
    \begin{aligned}
        m = H(x + n)
    \end{aligned}
    \label{eq:STE_discretize}
\end{equation}
where $x$ is the input of the discretization unit (output of the C-Net) and $n$ is noise sampled from a Gaussian distribution with standard deviation $\sigma_G$. However, during backpropagation we use the gradients of Equation \ref{eq:DRU_regularize} instead. 
The advantage of this approach over the original DRU can be seen in Figure \ref{fig:ST-DRU_response}. The agents receiving the messages will receive binary messages from the start of training. The DRU uses continuous messages during training. Even though the agents are encouraged by the DRU to produce outputs with a high absolute value, it will take a while before the output messages will resemble binary messages. The ST-DRU will also encourage the agent to produce outputs that can easily be discretized. But, the receiving agents will immediately receive discrete messages, allowing them to learn to interpret them more quickly.

\begin{figure*}
    \centering
    \includegraphics[width=\histogramwidth\linewidth, trim=65 10 80 35, clip]{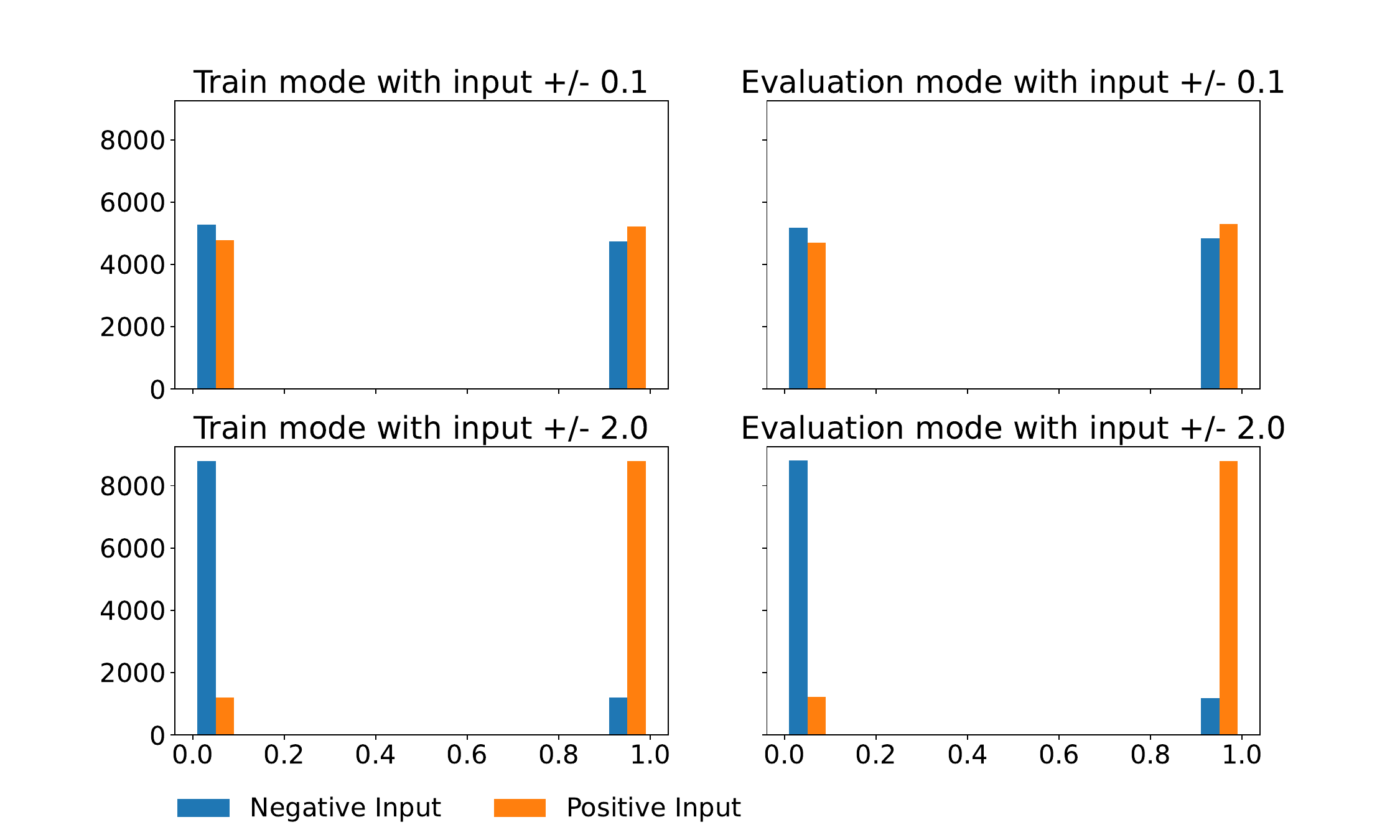}
    \caption{Histogram of the output of the ST-GS when calculating the output 10k times for different input values (positive and negative) for both training and evaluation mode}
    \label{fig:ST-GS_response}
\end{figure*}

\subsection{ST-GS}
Similarly to the ST-DRU, we also test the straight through GS (ST-GS) as proposed by Jang et al.\cite{jang2017categorical}. Here, we use the sampling technique described in Equation \ref{eq:gumbel_max_trick} to calculate the output, while using the gradients of the softmax approximation in Equation \ref{eq:gumbel_softmax} to train the communication network. Similarly to the DRU, the GS produces continuous messages during training. When evaluating the agents the GS will discretize the messages. If the sending agent is not producing outputs with a high enough absolute value, the difference between the messages at training time and at evaluation time will be very large. This prevents the receiving agents from correctly interpreting the message and choosing the appropriate actions. The ST-GS on the other hand will produce discretized messages at evaluation time and at training time, as can be seen in Figure \ref{fig:ST-GS_response}. This way, we make sure that the receiving agent knows how to correctly interpret discrete messages. 

\afterpage{%
    \clearpage
    \thispagestyle{empty}
    \begin{landscape}
        \begin{table*}[ht]
            \caption{Differences between the discretization methods where $x$ is the input of the discretization unit (the output of the C-Net), $H(x)$ is the heaviside function, $n \sim \mathcal{N}(0, \sigma_G^2)$ is Gaussian noise, $i \in \{0, 1\}$, $g_i \sim G$ is Gumbel noise, $\pi_0 = \sigma(x)$, $\pi_1 = (1 - \sigma(x))$, $\tau$ is the softmax temperature and $\sigma(x)$ is the sigmoid function} 
            \label{tab:discretization_units}
            \begin{tabular}{cccc}
                \toprule
                                & Training Output (forward pass)            & Function used for backward pass  & Evaluation Output (forward pass) \\ \midrule
                STE             & $H(x)$                                    & $x $                      & $H(x)$            \\ 
                DRU             & $\sigma(x + n)$                          & $\sigma(x + n)$          & $H(x)$            \\ 
                GS              & $softmax \left(\frac{log(\sigma(x)) + g_1}{\tau}, \frac{log(1-\sigma(x)) + g_2}{\tau} \right)[0]$                          & $softmax \left(\frac{log(\sigma(x)) + g_1}{\tau}, \frac{log(1-\sigma(x)) + g_2}{\tau} \right)[0]$                                  & $one\_hot(\underset{i}{argmax}[g_i + log \pi_i])[0]$                  \\
                ST-DRU          & $H(x + n)$                                & $\sigma(x + n)$           & $H(x)$            \\
                ST-GS           & $one\_hot(\underset{i}{argmax}[g_i + log \pi_i])[0]$             & $softmax \left(\frac{log(\sigma(x)) + g_1}{\tau}, \frac{log(1-\sigma(x)) + g_2}{\tau} \right)[0]$                                  & $one\_hot(\underset{i}{argmax}[g_i + log \pi_i])[0]$                   \\
                \bottomrule
            \end{tabular}
        \end{table*}
    \end{landscape}
    \clearpage
}

\subsection{COMA-DIAL}
\label{sec:coma_dial}
The communication learning principles used in DIAL are not only applicable to Q-learning but can also be used in combination with for example COMA. COMA provides much better results in multi-agent reinforcement learning environments than independent Q-learning because it uses a centralized critic and counterfactual reasoning to learn the action policies. Therefore, we will also perform a series of experiments using a combination of COMA and DIAL, which we will call COMA-DIAL. This method learns the communication policy in the same way as DIAL. The gradients obtained by training the action policy are propagated through the communication channel to update the communication policy.  The action policy of the agents will be trained using the COMA approach as explained in Section \ref{sec:coma}. COMA without these adaptations can only learn a communication policy by including it in the action policy. By doing this the action space grows significantly since the size is determined by the number of action and message combinations. The increased size of the action space makes it very hard for the agents to learn a communication policy. This was shown in the results of Vanneste et al.\cite{vanneste2021learning} where COMA was unable to learn a functional communication protocol in any of their experiments. By including the DIAL communication learning method in COMA, we are able to achieve significantly better results in our experiments.


Since the actor will not be able to learn a proper policy until the critic has learned to estimate the Q-value of a state-action pair, we adapt the learning rate of the actor based on the performance of the critic. This way the actor only adapts its policy when the critic can properly estimate the Q-values, making the policy learning more stable. The final learning rate for the actor ($\alpha$) is calculated by scaling the original learning rate ($\alpha_{max}$) depending on the critic loss and two hyperparameters $\eta_{min}$ and $\eta_{max}$ as can be seen in Equation \ref{eq:lr_scaling}.

\begin{equation}
    \alpha = \begin{cases}
        0, &if \ \eta_{max} < \mathcal{L}(\theta_{Q}^i) \\
        \left( 1 - \frac{\mathcal{L}(\theta_{Q}^i) - \eta_{min}}{\eta_{max} - \eta_{min}} ) \alpha_{max} \right), &if \ \eta_{min} < \mathcal{L}(\theta_{Q}^i) < \eta_{max} \\
        1, &if \ \mathcal{L}(\theta_{Q}^i) < \eta_{min}
    \end{cases} 
    \label{eq:lr_scaling}
\end{equation}

Exploration is often a challenge in MARL since the joint action and state space increases exponentially with the number of agents. Therefore, we use an explicit form of exploration in our COMA-DIAL experiments. During training, we modify the policy of our agent by taking an average of the agent policy and a uniform random policy weighted by weight $\epsilon$. This is similar to the use of $\epsilon$-greedy in Q-learning. In our experiments we use a constant value for $\epsilon$.
\begin{equation}
    \pi^a_b(u_t^a \vert o^a_t) = (1 - \epsilon) \pi^a(u_t^a \vert o^a_t) + \epsilon\frac{1}{\vert U \vert} 
    \label{eq:behaviour_policy}
\end{equation}

Since policy gradients is an on-policy method and the behaviour policy ($\pi_b^a$) and the agent policy ($\pi^a$) will be different, we need to use an off-policy variant of the policy gradient update rule \cite{degris2012}. 

\begin{equation}
\mathcal{L}^i(\theta^i_{\pi}) = -A(s_t, u_t) \frac{\pi^a(u^a \vert o^a)}{\pi^a_b(u^a \vert o^a)} \ln \pi^a(u^a \vert o^a) + \vert \theta^i_{\pi} \vert
\label{eq:actor_loss}
\end{equation}

where $\theta^i_{\pi}$ are the parameters of the actor at iteration $i$.

\section{Experiments}
\label{sec:experiments}
In this section, we explain each of our experiments and analyze the results. For each experiment we show the average performance over five different runs for each of the discretization methods. The hyperparameters and network architectures are identical for each of the discretization methods since our hyperparameter search showed that the best hyperparameters and network architecture were not influenced by the choice in discretization method. All of our experiments are run using the RLlib framework \cite{liang2018rllib}. 

In our experiments, we use several different environments. First, we perform experiments on two configurations of the matrix environment. In this environment, we can easily scale the amount of information that should be contained within a message and use a larger number of agents. Second, we modify this matrix environment to investigate the effect of communication errors. Next, we analyze the performance of the discretization units on two scenarios in the multiple particle environment (MPE)\cite{lowe2020multiagent, mordatch2018}. The MPE is a collection of benchmark scenarios that are often used in the state of the art in MARL \cite{lowe2020multiagent, mordatch2018, papoudakis2020, vanneste2020, vanneste2021learning}. In these environments, learning the action policy based on the observation and the incoming messages is more complex than in the matrix environment. In addition, unlike in the matrix environment, the agents do not have to communicate their entire observation but only a part. They need to be able to learn which information is useful for the other agents. In the speaker listener scenario, agents are either a speaker or a listener. This means that they have only an action policy or only a communication policy. The simple reference scenario changes this. Here the agents are both speaker and listener. Therefore the agents have both an action and a communication policy. Finally, we analyze the results in a modified speaker listener scenario of the MPE, the parallel speaker listener scenario \cite{vanneste2021learning}. In this scenario, there are multiple listeners which results in multiple feedback gradients going through the discretization units. This scenario is more complex because, with a higher number of agents, credit assignment and dealing with non-stationarity becomes more difficult. With these environments, we are able to investigate many environment properties and their effect on the performance of each discretization method.

\subsection{Matrix Environment}
\label{sec:matrix_env}
The matrix environment is inspired by the class of environments presented by Lowe et al.\cite{lowe2019measuring} called the Matrix Communication Games. In the matrix environment $N$ agents receive a natural number in $[0, M-1]$ in one-hot encoding. The values for $N$ and $M$ can be chosen independent of each other. The agents are allowed to broadcast one message to the other agents before they have to indicate whether all agents received the same number or not. The odds of the agents receiving the same number are 50\% regardless of the value of $N$ and $M$. The minimum number of bits required to be able to represent each possible input number can easily be determined by $log_2(M)$ and rounding up. The team reward in this environment is equal to the number of agents that correctly determined whether all agents got the same number or not. Therefore, the maximum reward is equal to $N$. Table \ref{tab:matrix_env} shows the reward matrices that correspond with a matrix environment with $N = 2$ and any value for $M$.

\begin{table}[t]
    \caption{Reward matrices for the matrix environment ($N = 2$ and any value for $M$). The agents have two possible actions, indicating they got the same number (S) or a different number (D)}
    \label{tab:matrix_env}
    \begin{subtable}[h]{0.45\linewidth}
        \centering
        \caption{The agents received the same number}
        \begin{tabular}{c|cc}
                & S     & D\\ \midrule
            S   & 2, 2  & 1, 1     \\
            D   & 1, 1  & 0, 0    
        \end{tabular}
    \end{subtable}
    \hfill
    \begin{subtable}[h]{0.45\linewidth}
        \centering
        \caption{The agents received a different number}
        \begin{tabular}{c|cc}
                & S     & D\\ \midrule
            S   & 0, 0  & 1, 1     \\
            D   & 1, 1  & 2, 2     
        \end{tabular}
    \end{subtable}
\end{table}

\begin{table}[t]
\centering
\caption{Different configurations of the matrix environment}
\label{table:matrix_configs}
\begin{tabular}{ccc}
\toprule
                           & N & M \\ \midrule
Simple Matrix Environment  & 3 & 4 \\
Complex Matrix Environment & 5 & 256 \\
\bottomrule
\end{tabular}
\end{table}

We examine the results for two different configurations of this environment which can be seen in Table \ref{table:matrix_configs}. The simple matrix environment is a small version of this environment while in the complex matrix environment there are more agents and they have to be able to represent more possible labels in their messages. In each of these experiments, we show the evaluation reward of our agents, measured by performing 100 evaluation episodes after each 100 training iterations. During the evaluation episodes, the agents do not explore and the discretization methods are applied in evaluation mode. In this environment, all of the agents are identical. Therefore, we can use parameter sharing between the agents, which improves their performance significantly as shown in the results of Foerster et al.\cite{foerster2016learning}.

\begin{figure*}
        \centering
        \includegraphics[width=\graphwidth\linewidth]{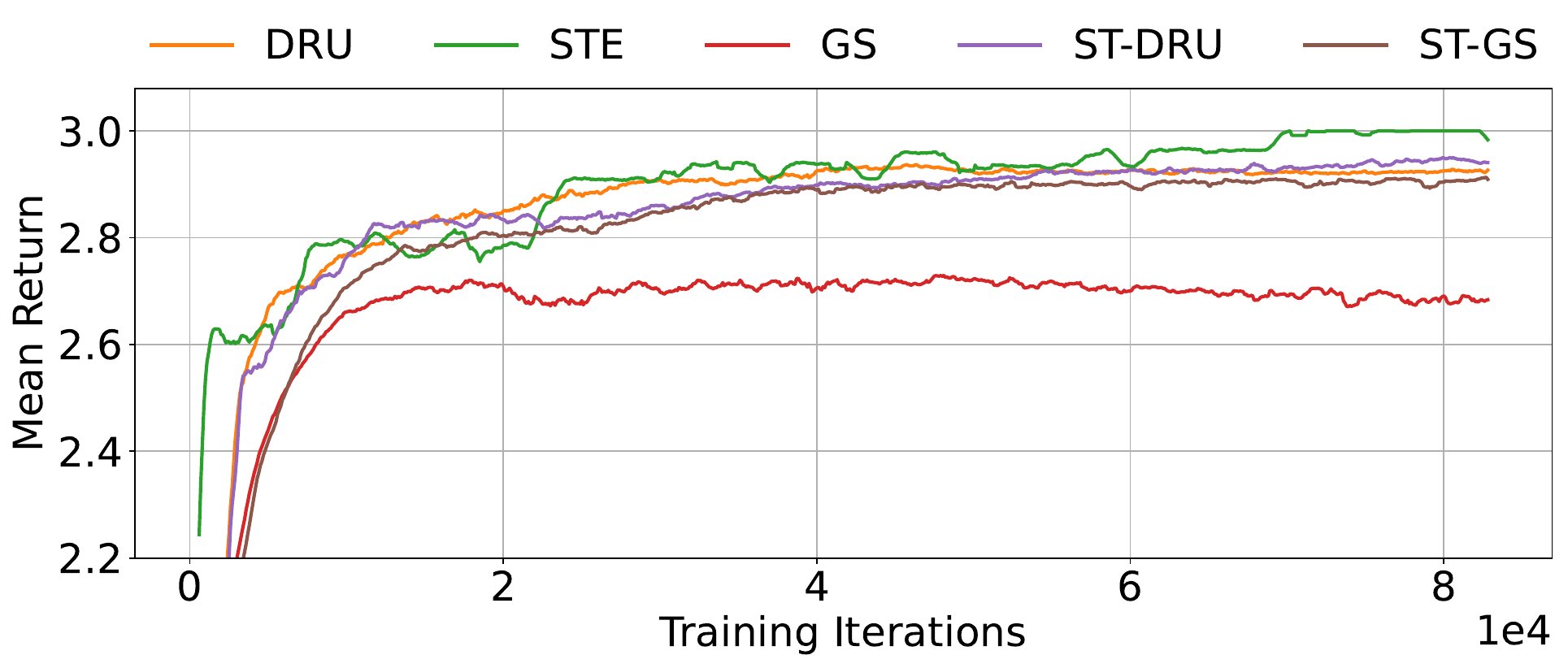}
        \caption{Results in the simple matrix environment}
        \label{fig:matrix_simple_eval_reward}
\end{figure*}

\subsubsection{Simple Matrix Environment}
In Figure \ref{fig:matrix_simple_eval_reward} and Table \ref{tab:conclusion_comparison}, the results for the different discretization methods in the simple matrix environment are shown. The maximum reward the agents can achieve in this scenario is a reward of 3. We can see that most of the methods are able to achieve a reward very close to this maximum except for the GS. The ST-GS does not have the same issue as the GS to achieve the maximum reward. We also see that the STE method is faster at the beginning of the training but this difference disappears rather quickly. However, at the end of the experiments the STE outperforms the other methods again. Due to the limited complexity of this environment, the differences between the methods are still small. 

\subsubsection{Complex Matrix Environment}

\begin{figure*}
        \centering
        \includegraphics[width=\graphwidth\linewidth]{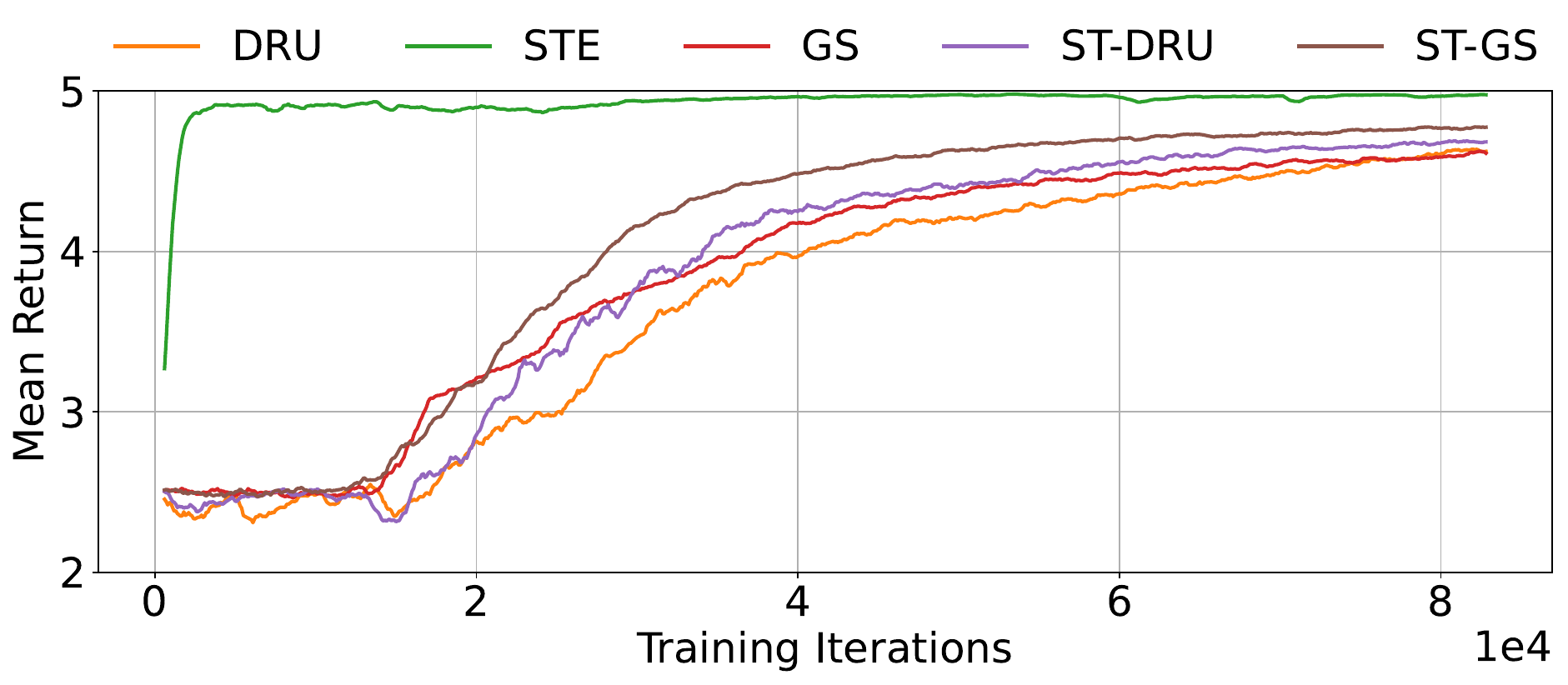}
        \caption{Results in the complex matrix environment}
        \label{fig:matrix_complex_eval_reward}
\end{figure*}

\begin{figure*}
        \centering
        \includegraphics[width=\graphwidth\linewidth]{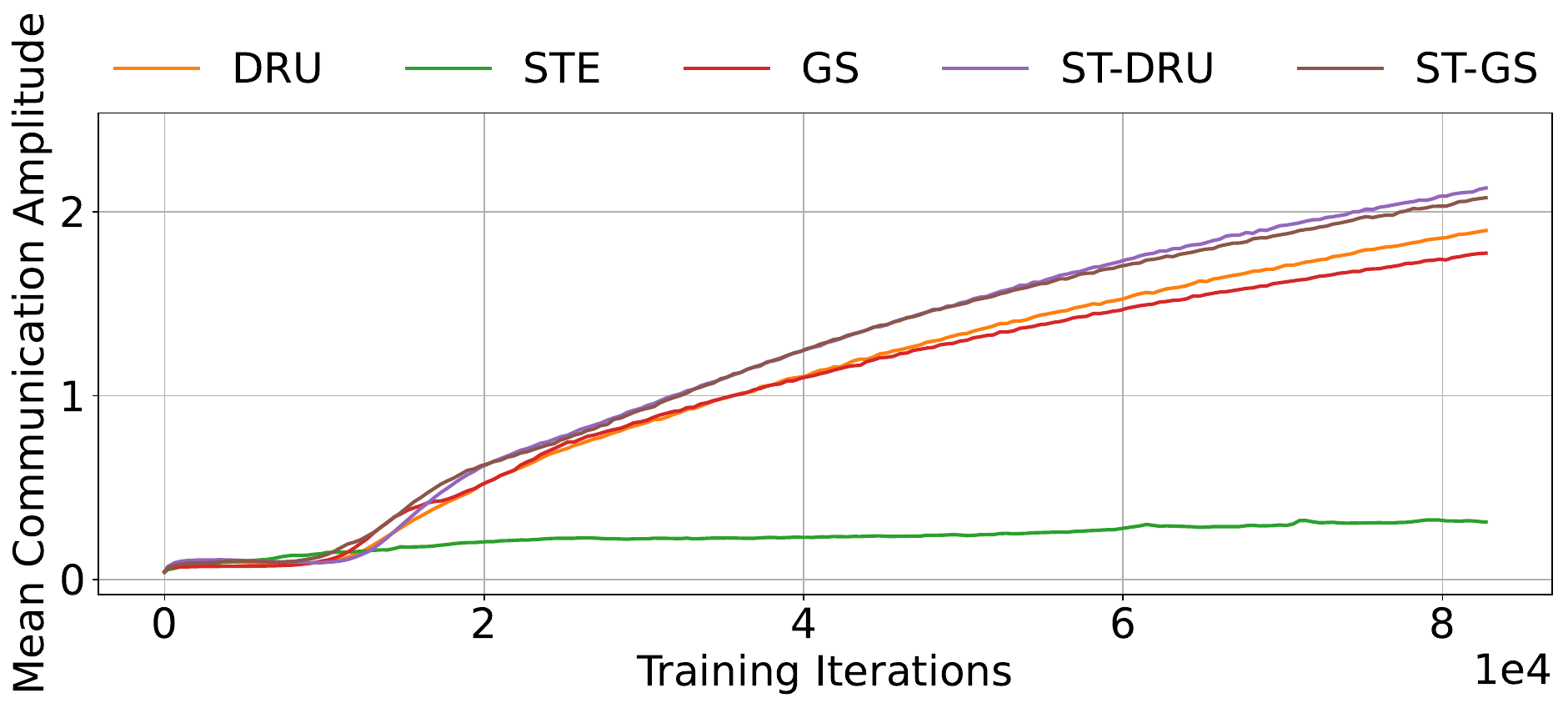}
        \caption{Communication amplitude in the complex matrix environment}
        \label{fig:matrix_complex_comm_amplitude}
\end{figure*}

The complex matrix environment has more agents ($N = 5$) as well as more possible input numbers ($M = 256$). The agents need a message consisting of a full byte to be able to encode each of the possible input numbers. Figure \ref{fig:matrix_complex_eval_reward} and Table \ref{tab:conclusion_comparison} show the results of this experiment. The maximum reward in this configuration is 5. We see that the difference between the methods is larger than in the simple matrix environment due to the added complexity. We see that the STE method is the only one that is able to reach the maximum reward in this training period. It reaches a reward close to the maximum reward after only 5k training iterations. The other methods only start improving after 15k training iterations and take over 60k training iterations to reach their maximal performance. We can also see that the adapted versions of the DRU and GS which include the STE technique perform better than the version without the STE technique. The ST-DRU has an average reward that is 0.072 higher than the DRU and the ST-GS has an average reward that is 0.176 higher than the GS during the final 10\% of training iterations. The communication amplitude in Figure \ref{fig:matrix_complex_comm_amplitude} provides an explanation for the training speed of the STE method. The communication amplitude is the mean absolute value of the input of the discretization unit. We can see a clear difference between the STE and the other methods. The communication amplitude of the STE stays below 0.5 while the communication amplitude of the DRU and GS approaches 1.8 and the communiation amplitude of the ST-DRU and the ST-GS exceeds 2.0. This is caused by the noise that is included in all of the discretization methods except for the STE. For a low communication amplitude the output of each of these discretization methods is still very random. The output during training will be determined by the noise instead of by the sign of the input which is done during evaluation. This encourages the agent to produce outputs with a higher communication amplitude. However, this puts a delay on the speed at which the agents can discover communication protocols. In Figure \ref{fig:matrix_complex_comm_amplitude} we see that the communication amplitude starts rising more quickly at around 15k training iterations. Once the communication amplitude starts rising we can see in Figure \ref{fig:matrix_complex_eval_reward} that the rewards that the agents receive also starts rising, indicating that they are starting to learn how to communicate with each other. 

The amplitude of the message output of our agents needs to reach a certain level to be able to consistently communicate, undisturbed by the noise. This slows down the agents compared to the STE approach which does not need to reach a certain communication amplitude to communicate effectively. A possible way to counteract this effect for the DRU and DRU-STE approaches is to reduce the standard deviation of the gaussian noise that is used in these methods while also increasing the slope of the sigmoid function to achieve the same effect as the original. However, by doing this the DRU(-STE) gets assymptotically closer to being a threshold function. In order to reach the same results as the STE, the standard deviation of the noise should be zero and the slope should be infinitely steep. This reintroduces the problems we have with normal discretization since the gradients will be the output of a Dirac function which are unusable. Therefore, it is not possible to reach the same training speed using the DRU(-STE) as we are able to achieve using the STE method, making the STE the best choice in this environment. 

\subsection{Error Correction}
\begin{figure*}
        \centering
        \includegraphics[width=\graphwidth\linewidth]{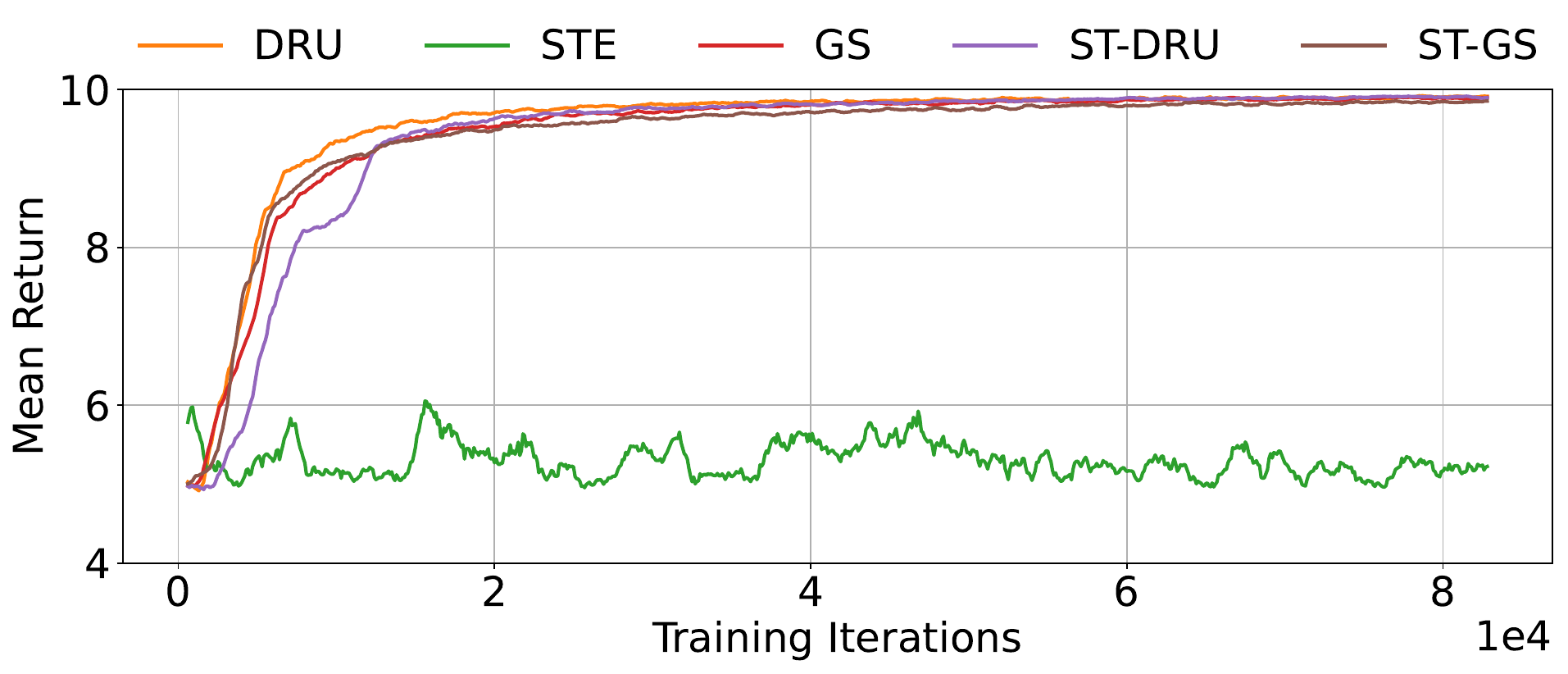}
        \caption{Results in the matrix environment with a 50\% chance that a bit error occurs at a random location in the message}
        \label{fig:matrix_1_bitflip}
\end{figure*}

In addition to comparing the different discretization methods in ideal circumstances, we also want to make this comparison in a situation with more uncertainty. Therefore, we perform some additional experiments on the matrix environment as discussed in Section \ref{sec:matrix_env}. However, instead of perfect communication circumstances without any errors as done before, we flip a certain amount of random bits with a certain probability. This causes the receiver to receive different information than intended by the sender. Depending on the maximum amount of bits that can be flipped, the agents need more message bits to be able to counteract the errors that are introduced. We use a simple matrix environment with $N = 10$ and $M = 2$. In this experiment, we show the evaluation reward of our agents, measured by performing 100 evaluation episodes after each 100 training iterations. During the evaluation episodes, the agents do not explore and the discretization methods are applied in evaluation mode. In this environment, all of the agents are identical. Therefore, we can use parameter sharing between the agents, which improves their performance significantly as shown in the results of Foerster et al.\cite{foerster2016learning}. 

We perform a test where there is 50\% chance that an error will occur. Normally, the agents would be able to represent both possible incoming numbers using a single bit. However, if they have to be able to correct the errors that get introduced, they require three bits. The results of this experiment can be seen in Figure \ref{fig:matrix_1_bitflip} and Table \ref{tab:conclusion_comparison}. We see that the STE is not able to correct the errors that occur. Therefore, it is not able to achieve good results. However, the other discretization methods are able to detect and correct the errors. Our hypothesis is that these methods are better at discovering communication policies to tackle errors due to the noise that is used within these discretization methods. The noise helps explore possible communication policies in addition to helping with the discretization. The STE method does not have any way to explore the communication space to find a communication policy that is unaffected by errors, therefore, using the STE, the agents are not able to achieve the goal. This means that in environments where the ideal communication policy is less straightforward, the STE method will not be the ideal choice to discretize the messages.

\begin{table}[t]
\centering
\begin{tabularx}{0.8\linewidth}{Xcc}
                                & \textbf{Before Errors}        & \textbf{After Errors} \\
\raisebox{0.6cm}{\textbf{DRU}}    & \includegraphics[width=\confusionwidth\linewidth, trim=30 90 45 100, clip]{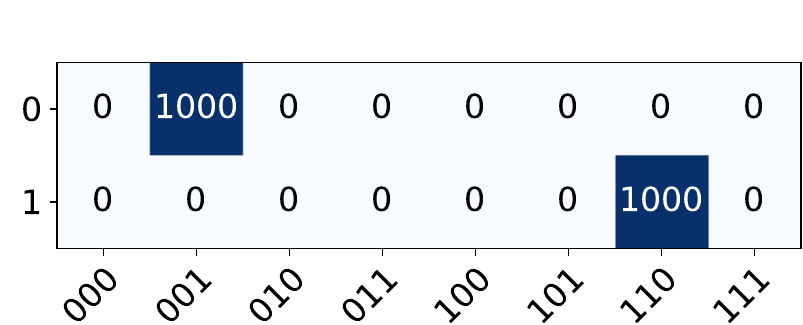}  &  \includegraphics[width=\confusionwidth\linewidth, trim=30 90 45 100, clip]{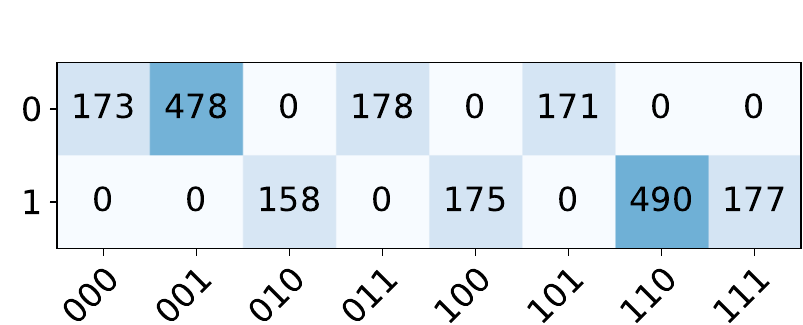}\\
\raisebox{0.6cm}{\textbf{STE}}    & \includegraphics[width=\confusionwidth\linewidth, trim=30 90 45 100, clip]{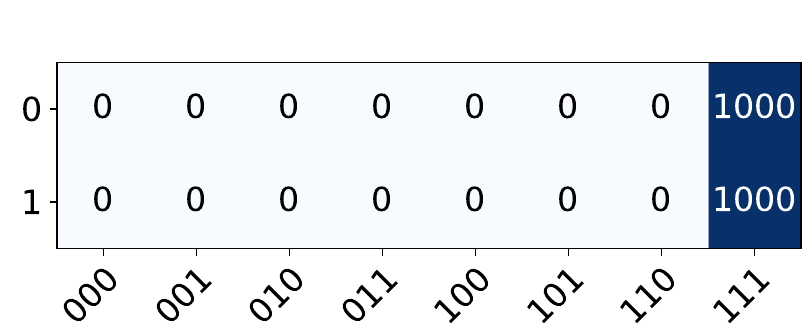}  &  \includegraphics[width=\confusionwidth\linewidth, trim=30 90 45 100, clip]{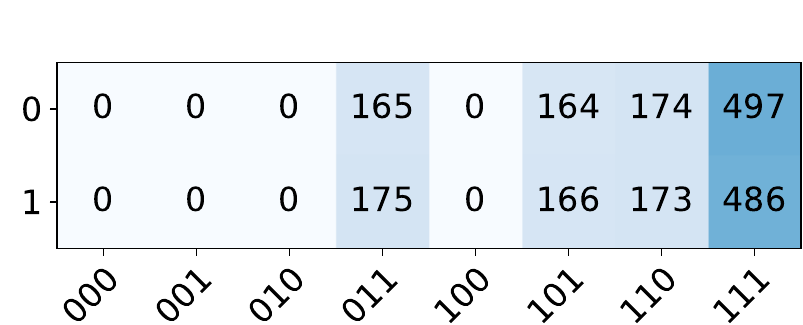}\\
\raisebox{0.6cm}{\textbf{GS}}     & \includegraphics[width=\confusionwidth\linewidth, trim=30 90 45 100, clip]{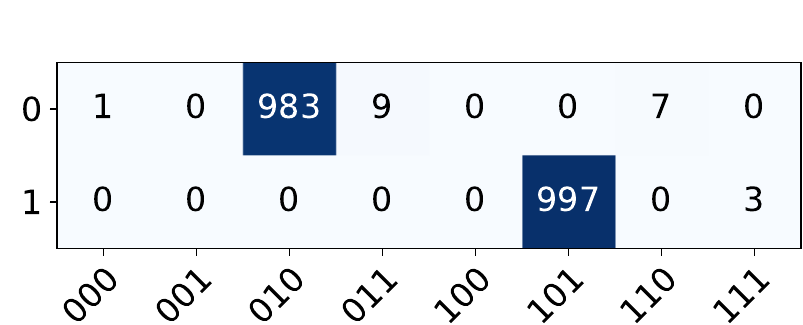}  &   \includegraphics[width=\confusionwidth\linewidth, trim=30 90 45 100, clip]{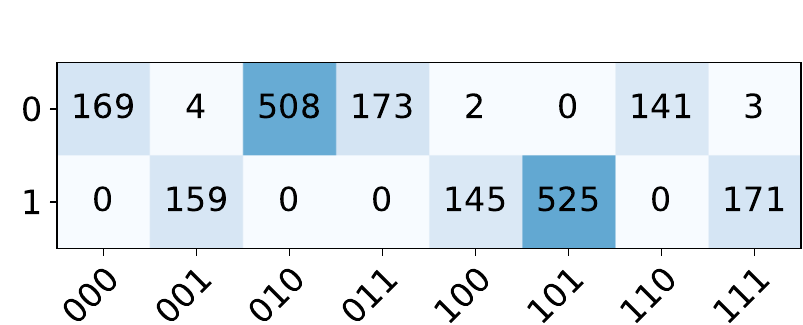}\\
\raisebox{0.6cm}{\textbf{ST-DRU}} & \includegraphics[width=\confusionwidth\linewidth, trim=30 90 45 100, clip]{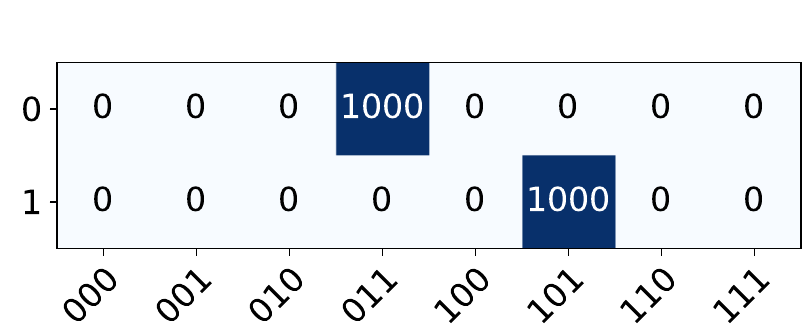} & \includegraphics[width=\confusionwidth\linewidth, trim=30 90 45 100, clip]{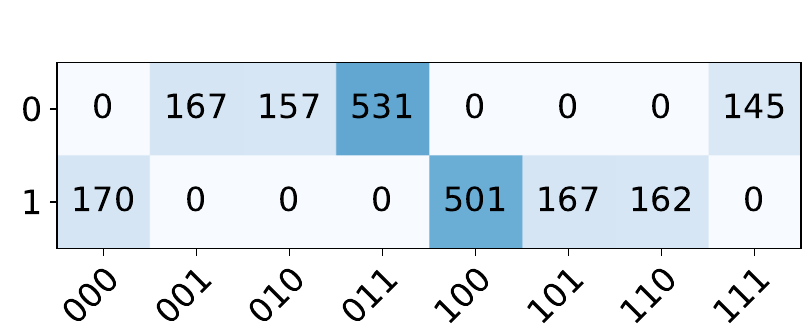} \\
\raisebox{0.6cm}{\textbf{ST-GS}}  & \includegraphics[width=\confusionwidth\linewidth, trim=30 90 45 100, clip]{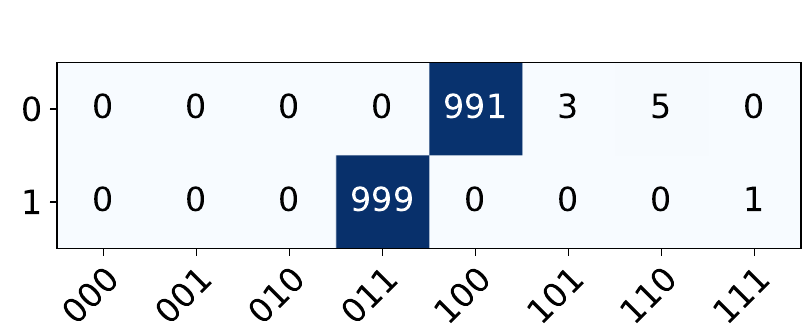}  & \includegraphics[width=\confusionwidth\linewidth, trim=30 90 45 100, clip]{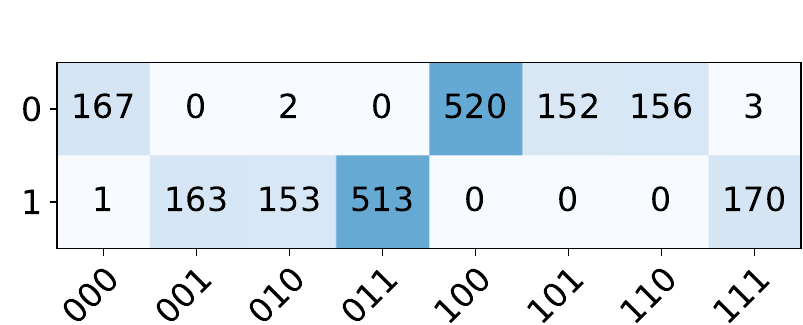}
\end{tabularx}
\captionof{figure}{The communication protocol for the matrix environment ($N = 10$, $M = 2$) for each of the discretization units before and after introducing errors. On the y-axis the different possible input numbers are displayed. On the x-axis the different possible output messages are displayed.}
\label{fig:confusion_matrices}
\end{table}

To see how the agents are able to correct the introduced errors, we examine which message the agents choose for which incoming number. Figure \ref{fig:confusion_matrices} shows the different communication policies for the different discretization methods. We can see the output message depending on which input number was given to the agent before and after the errors are introduced. We see that the agents with the DRU, GS, ST-DRU or ST-GS have chosen messages where the possible messages after the introduction of errors do not overlap between the possible input numbers. This way the agents make sure that the messages are still comprehensible, even if errors occur. For the GS we see that there are 9 messages that overlap with the output messages for a different input number after error introduction. The same thing can be observed for the ST-GS in 6 cases. We see that when we use the STE, the agents are not able to find this communication protocol. Even before the errors are introduced, the messages for both possible inputs are the same. This indicates that the agents did not find a useful communication protocol. 

In summary, we can state that the STE is not able to deal with environments where communication errors occur. Each of the other methods is able to deal with this and learn a functional communication policy in this scenario. The communication policy of the DRU and ST-DRU is able to avoid any information loss due to introduced errors. While the communication policy of the GS and ST-GS avoids information loss in the vast majority of cases, there are still some cases where information loss occurs. This is due to the fact that the (ST-)GS still uses the Gumbel noise during evaluation which results in stochasticity in the output. Given these results we can conclude that the DRU and ST-DRU are best suited to deal with environments where communication errors can occur.

\subsection{Speaker Listener Environment}

As a more complex environment, we use the speaker listener scenario from the MPE by OpenAI \cite{lowe2020multiagent, mordatch2018}. In this environment there are two agents and three landmarks. One of the agents, the speaker, observes which landmark is the target during this episode. The speaker then has to communicate this information to the other agent, the listener. Next, the listener has to navigate to the target landmark. Both agents are rewarded using a team reward that is composed based on the distance of the listener to the target landmark. Contrary to the matrix environment, the agents are not the same in this environment. The speaker will only consist of a communication policy while the listener will only consist of an action policy. In this experiment, we show the evaluation reward of our agents, measured by performing 10 evaluation episodes after each 50 training iterations. During the evaluation episodes, the agents do not explore and the discretization methods are applied in evaluation mode. Since the agents are not identical in this environment, we cannot apply parameter sharing.

\begin{figure*}
    \centering
    \includegraphics[width=\graphwidth\linewidth]{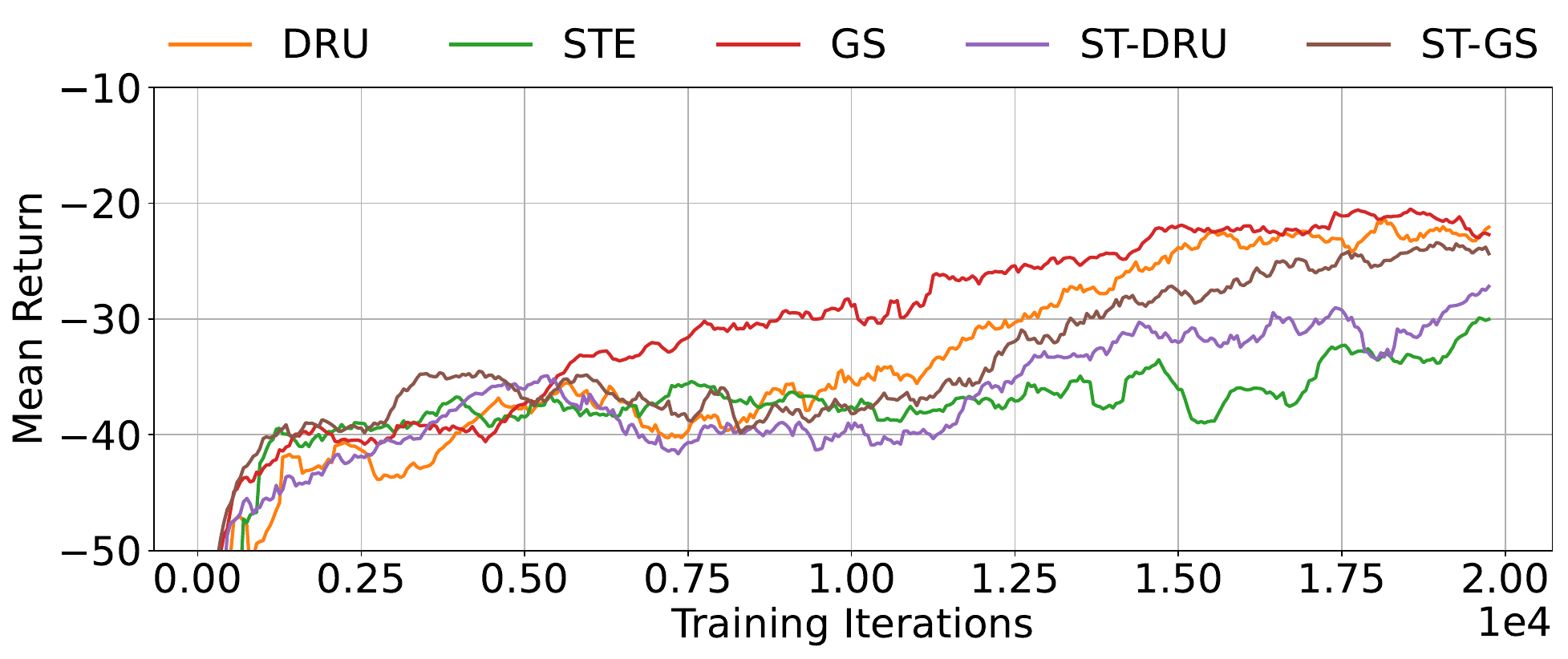}
    \caption{Results in the speaker listener environment using DIAL}
    \label{fig:speaker_listener_eval_reward}
\end{figure*}
\begin{figure*}
    \centering
    \includegraphics[width=\graphwidth\linewidth]{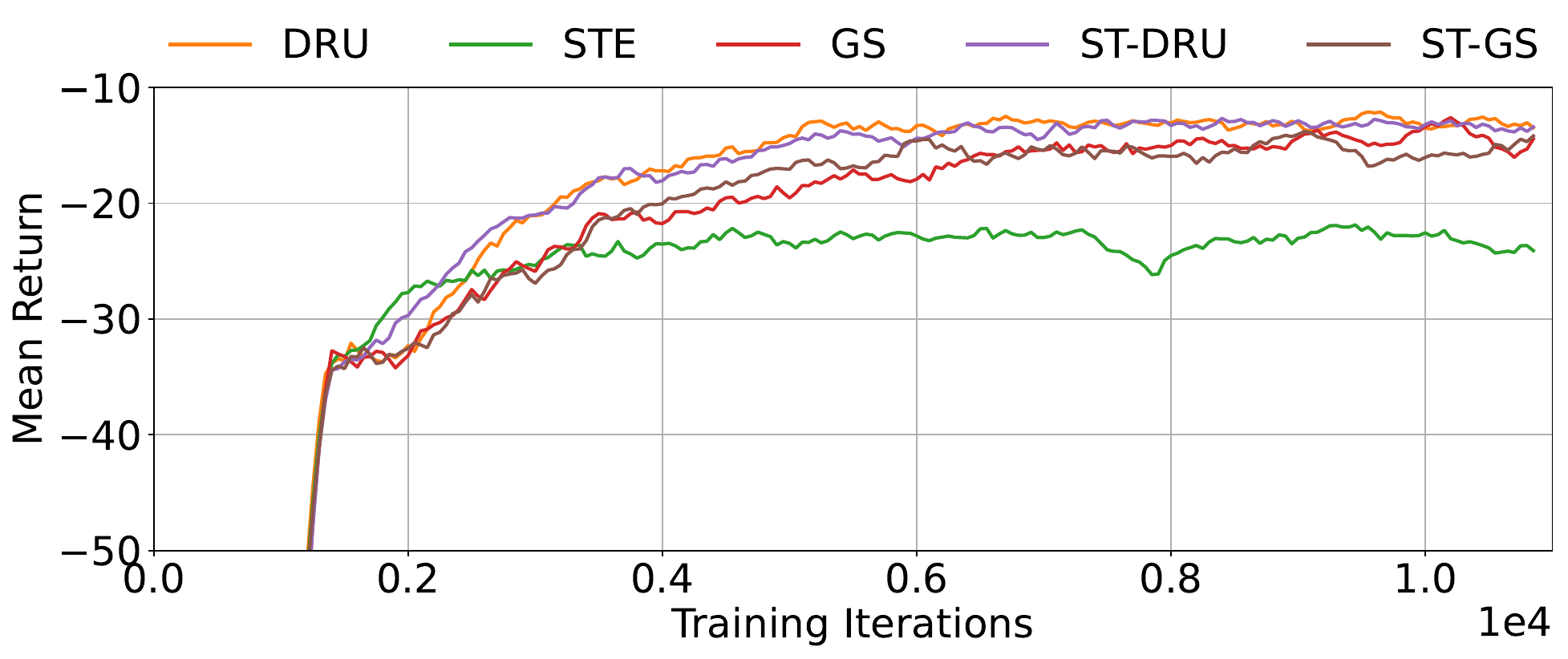}
    \caption{Results in the speaker listener environment using COMA-DIAL}
    \label{fig:speaker_listener_eval_reward_coma_dial}
\end{figure*}

Figure \ref{fig:speaker_listener_eval_reward} and Table \ref{tab:conclusion_comparison} show the results in the speaker listener environment. We see that the STE is no longer performing as good as in the matrix environment. It results in the worst result while the GS and DRU provide the best results. The delay that is caused by the noise in the DRU, GS, ST-DRU and ST-GS is no longer the determining factor in the training speed. The exploration that is done in the communication policy by the DRU, GS, ST-DRU and ST-GS due to the noise appears to have a beneficial result.

In this environment, we see that the original Q-learning variant of DIAL starts to have trouble achieving the goal. Therefore, we will switch to the COMA variant of DIAL that we presented in Section \ref{sec:coma_dial}. COMA is much better equiped to tackle more complex environments. The results of the different discretization methods used in COMA-DIAL can be seen in Figure \ref{fig:speaker_listener_eval_reward_coma_dial} and Table \ref{tab:conclusion_comparison}. When we compare the results of Figure \ref{fig:speaker_listener_eval_reward_coma_dial} with the results achieved with the Q-learning DIAL variant, we see that COMA-DIAL is able to achieve better results. The DRU and ST-DRU achieve the best results in this experiment. The STE is not able to achieve the goal in each of the experiment runs which is why the average performance is lower and the standard deviation is higher. The GS and ST-GS are not able to match the performance of the DRU and ST-DRU. This is due to the fact that during evaluation the (ST-)GS still uses noise. This causes stochasticity in the output of the (ST-)GS as can be seen in Figure \ref{fig:GS_response} and \ref{fig:ST-GS_response} which can result in mistakes.

\subsection{Simple Reference Environment}

\begin{figure*}
    \centering
    \includegraphics[width=\graphwidth\linewidth]{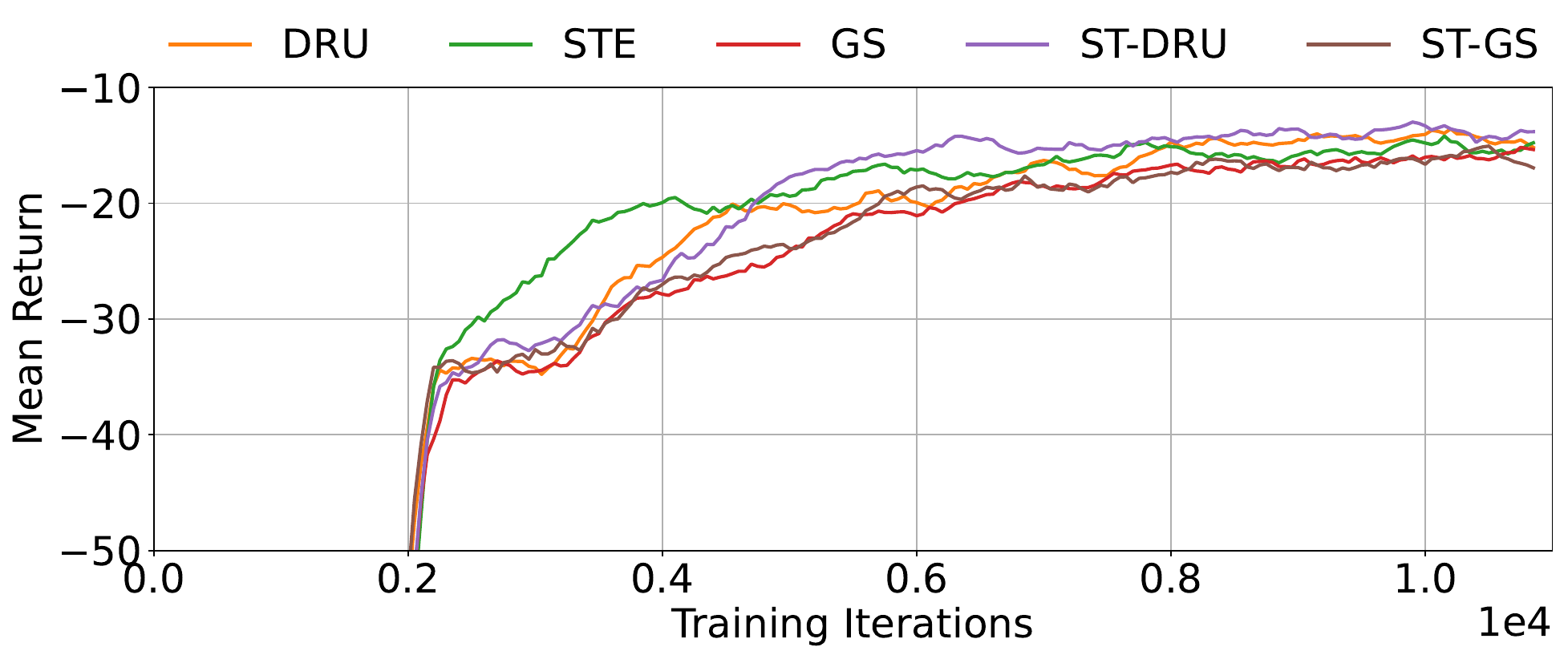}
    \caption{Results in the simple reference environment}
    \label{fig:simple_reference}
\end{figure*}

The simple reference environment is similar to the speaker listener environment. Here, both agents are able to navigate in the environment and observe the target landmark of the other agent. Both agents are both speaker and listener in this scenario. This means they both have an action and communication policy. In this experiment, we show the evaluation reward of our agents, measured by performing 10 evaluation episodes after each 50 training iterations. During the evaluation episodes, the agents do not explore and the discretization methods are applied in evaluation mode. In this environment, all of the agents are identical. Therefore, we can use parameter sharing between the agents, which improves their performance significantly as shown in the results of Foerster et al.\cite{foerster2016learning}. The results can be seen in Figure \ref{fig:simple_reference} and Table \ref{tab:conclusion_comparison}. Each of the discretization methods is able to learn a communication protocol, allowing the agents to achieve the goal. However, we can still see some difference between each of the methods. In the beginning of training we see that the STE learns a lot faster. After a while though the other methods reach the same performance and the ST-DRU and DRU reach the best performance among all methods. The GS and ST-GS achieve the worst results in this environment.

\subsection{Parallel Speaker Listener Environment}

\begin{figure*}
    \centering
    \includegraphics[width=\graphwidth\linewidth]{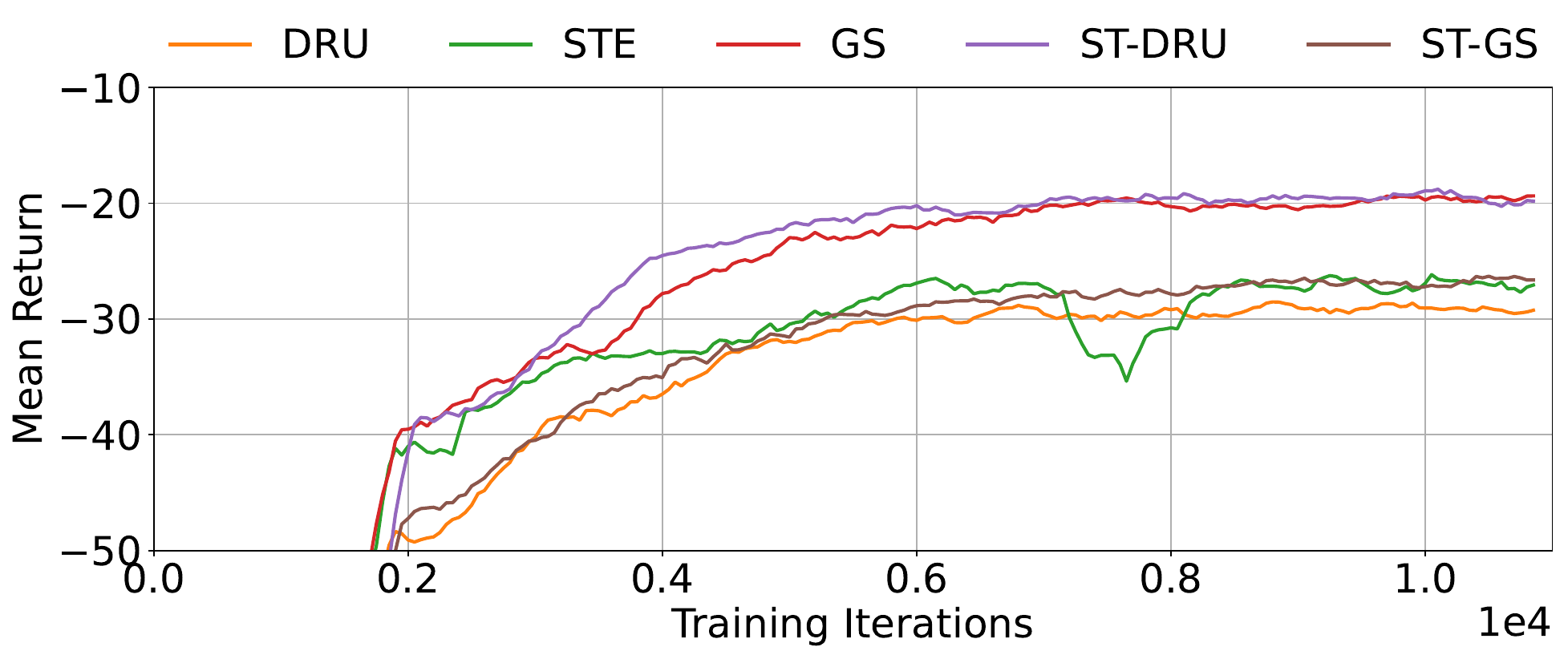}
    \caption{Results in the parallel speaker listener environment}
    \label{fig:parallel_speaker_listener_eval_reward}
\end{figure*}

The parallel speaker listener environment \cite{vanneste2021learning} is a modification of the speaker listener environment. In the parallel speaker listener environment there is one speaker and multiple listeners. The speaker communicates the same target landmark to each of the listeners. In this experiment, we show the evaluation reward of our agents, measured by performing 10 evaluation episodes after each 50 training iterations. During the evaluation episodes, the agents do not explore and the discretization methods are applied in evaluation mode. Since each of the listeners agents are identical, we can use parameter sharing for these agents but not for the speaker agent. The results can be seen in Figure \ref{fig:parallel_speaker_listener_eval_reward} and Table \ref{tab:conclusion_comparison}. There is a clear difference between the different methods. The GS and ST-DRU methods perform significantly better than the other methods. The standard deviation on the average return of the other methods shows that these methods have runs that manage to get a lot closer to the goal but also runs that fail entirely. The return threshold that the agents are able to reach without needing communication is around -30. The DRU, ST-GS and STE do not manage to get a return that is much higher than that on average. The STE also shows instability between 7k and 8k training iterations.

\afterpage{%
    \clearpage
    \thispagestyle{empty}
    \begin{landscape}
        \begin{table*}[t]
            \centering
            \caption{The average return and the standard deviation for each of the experiments during the final 10\% of training iterations}
            \label{tab:conclusion_comparison}
            \begin{tabularx}{0.8\paperheight}{XXXXXX}
                \toprule
                Experiment Name                             & DRU                  &  STE                 & GS                    & ST-DRU                   & ST-GS                \\
                \midrule
                Simple Matrix (DIAL)                        & $2.924 \pm 0.152$    &  $2.999 \pm 0.005$   & $2.685 \pm 0.275$     & $2.944 \pm 0.112$         & $2.906 \pm 0.123$   \\
                Complex Matrix (DIAL)                       & $4.600 \pm 0.118$    &  $4.972 \pm 0.018$   & $4.588 \pm 0.048$     & $4.672 \pm 0.134$         & $4.764 \pm 0.047$   \\
                Error Correction (DIAL)                     & $9.904 \pm 0.051$    &  $5.140 \pm 0.413$   & $9.892 \pm 0.051$     & $9.906 \pm 0.048$         & $9.842 \pm 0.059$   \\
                Speaker Listener (DIAL)                     & $-22.330 \pm 4.514$  &  $-32.224 \pm 6.833$ & $-21.533 \pm 4.758$   & $-29.966 \pm 9.430$       & $-24.211 \pm 5.177$ \\
                Speaker Listener (COMA-DIAL)                & $-13.139 \pm 4.062$  &  $-23.486 \pm 7.366$ & $-14.074 \pm 3.401$   & $-13.295 \pm 3.610$       & $-15.181 \pm 4.519$ \\
                Simple Reference (COMA-DIAL)                & $-14.428 \pm 2.918$  &  $-15.054 \pm 2.966$ & $-15.600 \pm 3.356$   & $-13.752 \pm 2.758$       & $-16.196 \pm 3.440$ \\
                Parallel Speaker Listener (COMA-DIAL)       & $-29.275 \pm 12.593$ &  $-26.799 \pm 7.477$ & $-19.581 \pm 4.129$   & $-19.499 \pm 3.300$       & $-26.826 \pm 14.691$\\
                \bottomrule
            \end{tabularx}
        \end{table*}
    \end{landscape}
    \clearpage
}

\section{Discussion}
\label{sec:discussion}

In our experiments, we compared different discretization techniques in different environments where the agents need to learn a communication protocol to achieve their goal. In this section, we discuss some general trends that we saw accross the experiments. Table \ref{tab:conclusion_comparison} shows how each of the different methods performed in each experiment. It shows the average return and standard deviation during the last 10\% of training iterations. 

To allow us to extend our analysis to more complex environments, we presented COMA-DIAL, a novel communication learning method. In our results, we saw that it was able to clearly outperform the Q-learning version of DIAL in the speaker listener environment. Using COMA-DIAL, we were able to include the simple reference and parallel speaker listener scenario in our experiments which gave us more insight in the performance of the different discretization methods.

In our results, we saw only small differences between the different discretization methods in the simple matrix environment. However, the differences become more apparent when using more complex environments. Initially, the STE appeared to perform best. It allowed the agents to learn significantly faster in the complex matrix environment. When we analyse the performance in the other environments however, we can see that the STE no longer outperforms the other methods. In the error correction task, the agents using the STE are not able to learn a valid communication policy due to the lack of exploration in the communication policy. 

When we compare the difference in performance between the DRU and GS and their straight through versions, we can see that the use of the straight through method can make a significant difference. In the parallel speaker listener environment, the GS significantly outperforms the ST-GS while in the simple matrix environment, the agents using the GS perform significantly worse than all the other methods. Therefore, we cannot conclusively determine which of these methods is the best. The difference between the ST-DRU and the DRU is more clear. Both methods perform very well in almost all environments. The only environment which causes issues for the DRU is the parallel speaker listener. Here, the ST-DRU achieves the best performance of all methods, while the DRU performs the worst. In the other environments, even though the DRU is able to achieve good performance, the ST-DRU performs similar or better than the DRU. We can conclude that in the case of the DRU the use of the straight through method has proven to be beneficial. We also observed that, in several environments, the GS and ST-GS perform slightly worse than the DRU and ST-DRU. The Gumbel noise used in the GS and ST-GS during evaluation makes the output stochastic which results in mistakes. Therefore, in most environments the GS and ST-GS are not the best choice.

When comparing all of the methods, the best performing method is not the same in all experiments. However, the ST-DRU is the only method that consistently ranks among the best performing methods. The ST-DRU is also the only method that does not fail to achieve the goal. The other methods all have trouble achieving the goal in at least one environment. Therefore, for most applications, the ST-DRU is the best choice as a discretization method. In some environments, other methods can achieve a slightly higher return or faster training speed, for example the STE in the complex matrix environment. However, in most cases this does not outweigh the risk of failing to achieve the goal in other environments. Therefore, in general, the ST-DRU is the best choice. 

\section{Conclusion}
\label{sec:conclusion}

In this paper we compared several discretization methods in different environments with different complexities and challenges. We focus on the situation where these discretization methods are used to discretize communication messages between agents that are learning to communicate with each other while acting in an environment. Along with several approaches used in the state-of-the-art of communication learning we also present the ST-DRU approach, which combines the DRU and STE approaches. We also introduce COMA-DIAL, a novel communication learning approach that combines COMA \cite{foerster2018counterfactual} and DIAL \cite{foerster2016learning} and uses an explicit exploration strategy and learning rate scaling on the actor, allowing us to include more complex environments in our comparisons. 

The results showed that the choice of discretization method can have a big impact on the performance of the agents. Across all of the experiments, the ST-DRU showed the best results. In all of the environments, the ST-DRU either achieved the best or close to the best performance and, unlike the other approaches, the ST-DRU did not fail in any of the environments. 

In the future, it can be interesting to investigate strategic communication where, in addition to sharing observation information, the agents also communicate about the strategy to reach the goal.

\begin{appendices}

\section{Detailed Results}
In this section, we provide detailed results for all experiments in this work. We show the average return along with the 95\% confidence interval calculated using the Bootstrap Method and bootstrapping 25 times.
\subsection{Matrix Environment}
\subsubsection{Simple Matrix Environment}
\begin{figure*}[h]
    \centering
    \includegraphics[width=\graphwidth\linewidth]{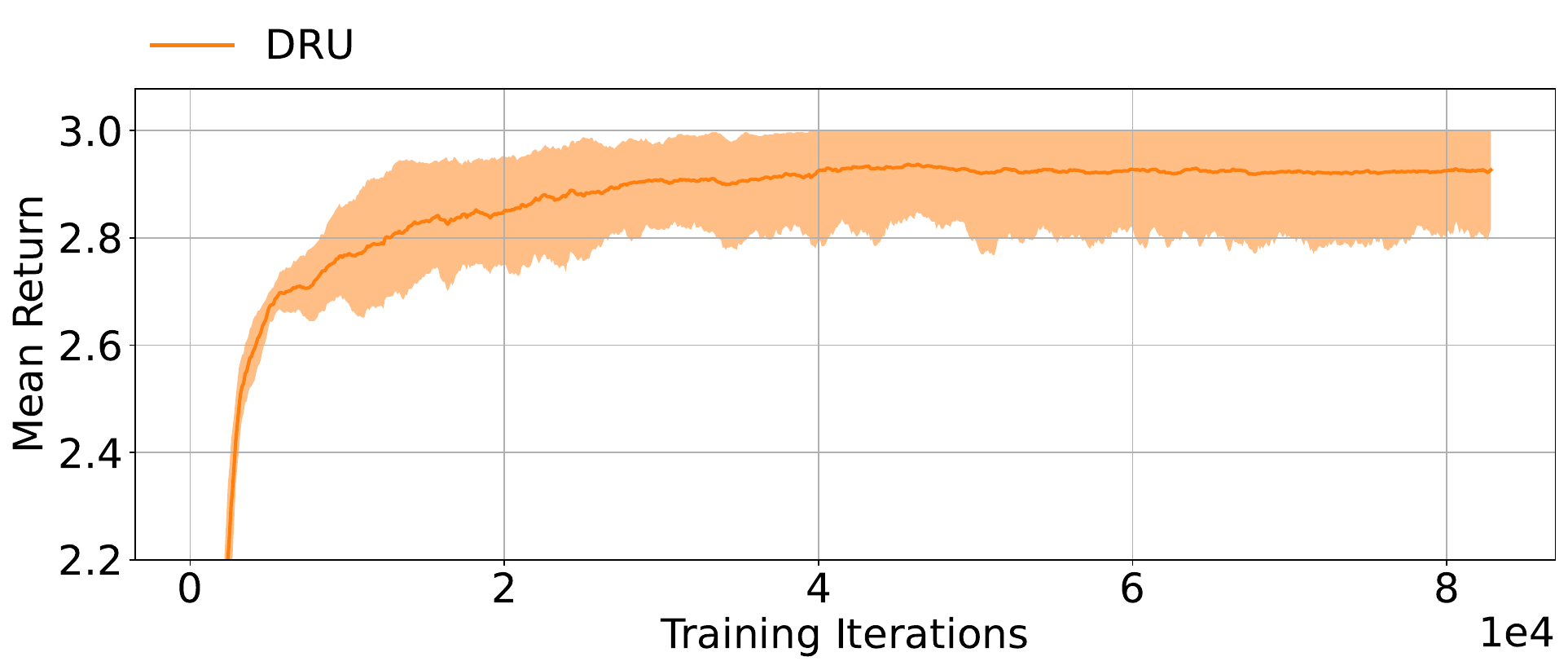}
    \caption{DRU in the simple matrix environment}
\end{figure*}
\begin{figure*}[h]
    \centering
    \includegraphics[width=\graphwidth\linewidth]{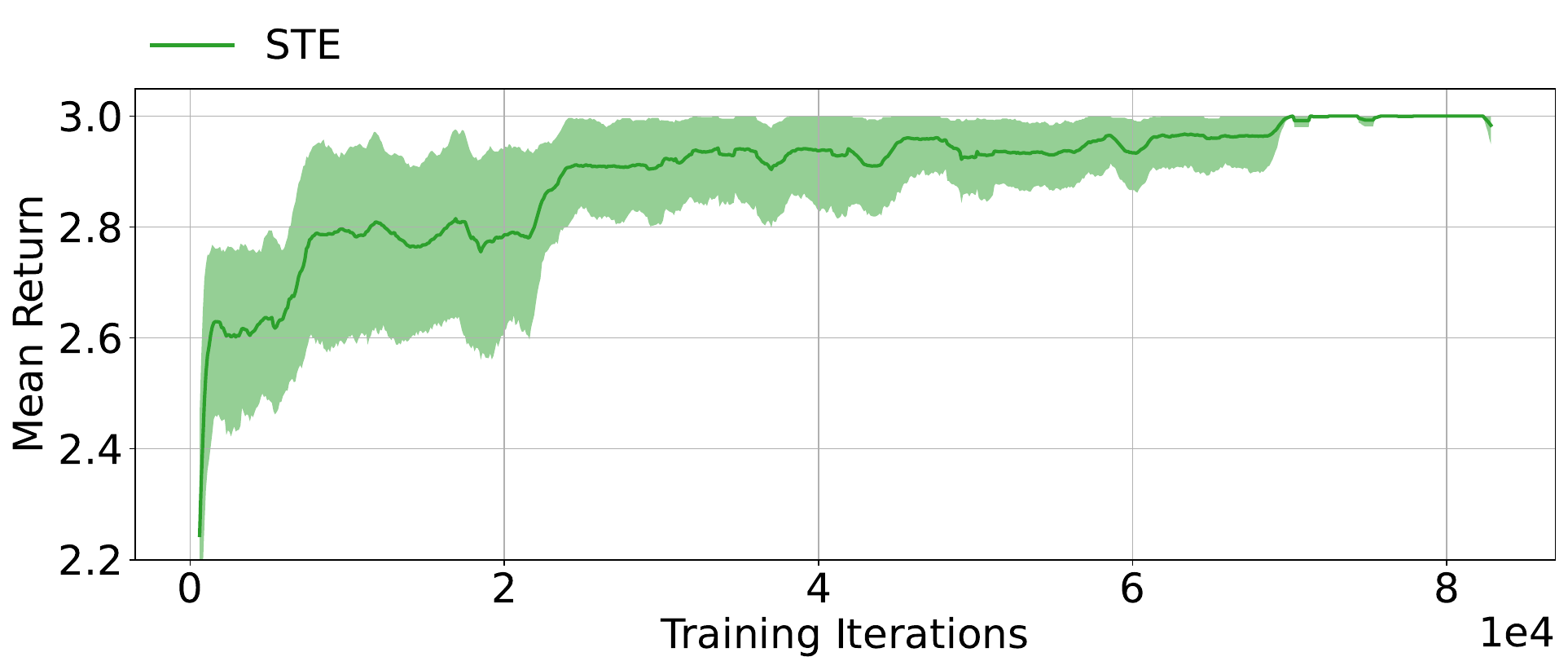}
    \caption{STE in the simple matrix environment}
\end{figure*}
\begin{figure*}[h]
    \centering
    \includegraphics[width=\graphwidth\linewidth]{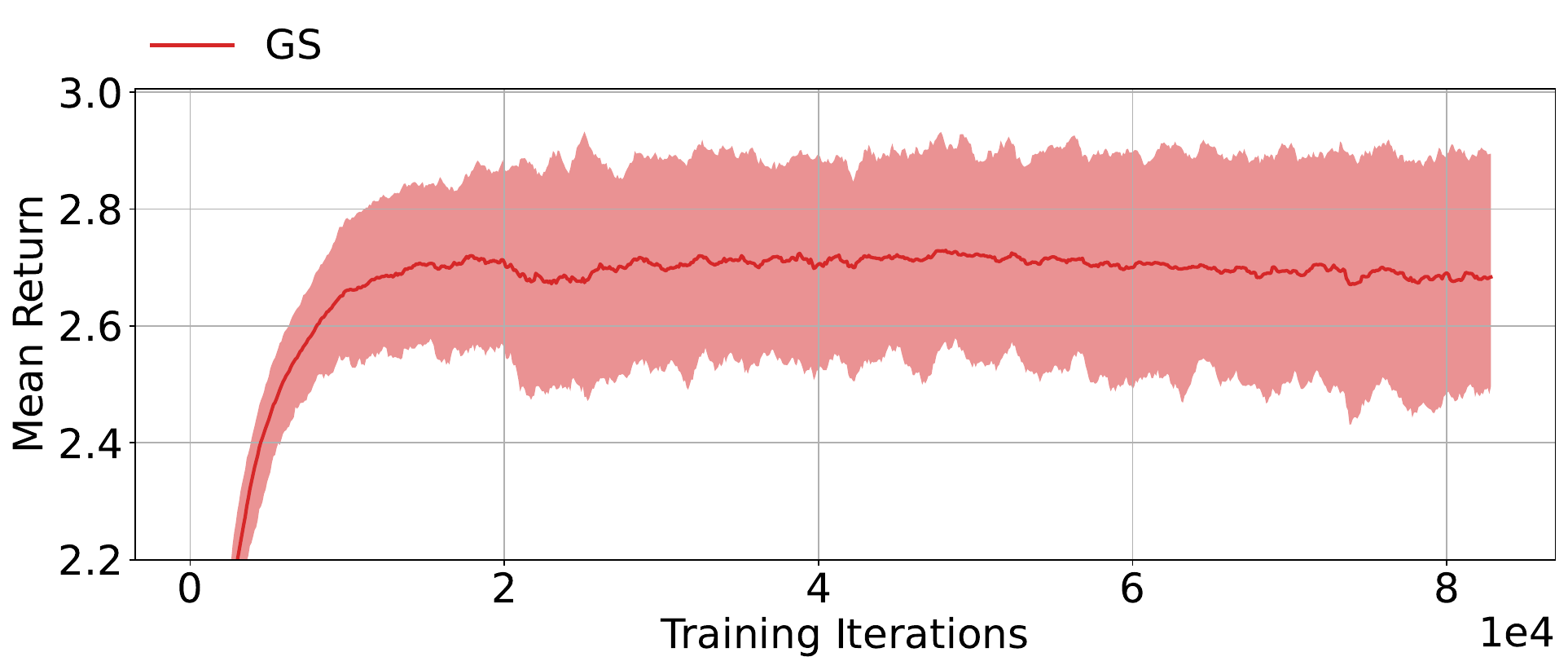}
    \caption{GS in the simple matrix environment}
\end{figure*}
\begin{figure*}[h]
    \centering
    \includegraphics[width=\graphwidth\linewidth]{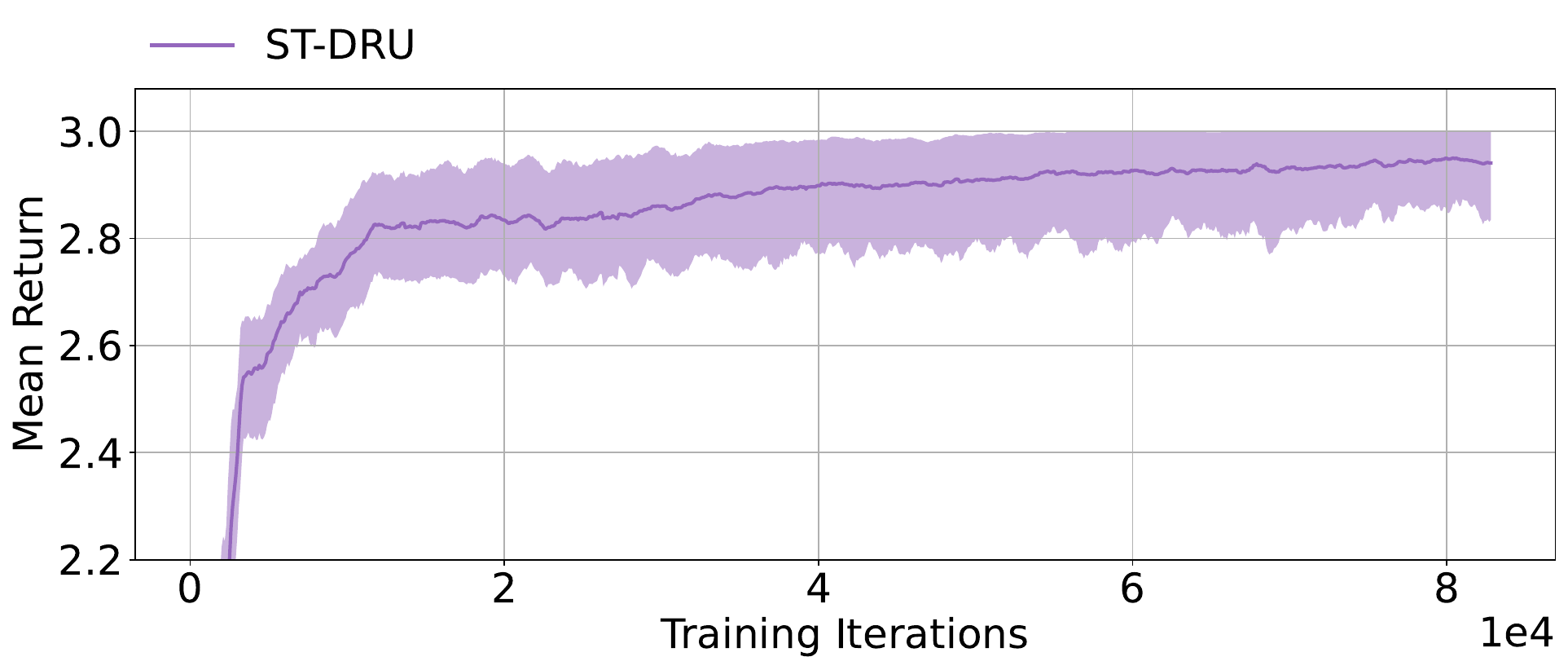}
    \caption{ST-DRU in the simple matrix environment}
\end{figure*}
\begin{figure*}[h]
    \centering
    \includegraphics[width=\graphwidth\linewidth]{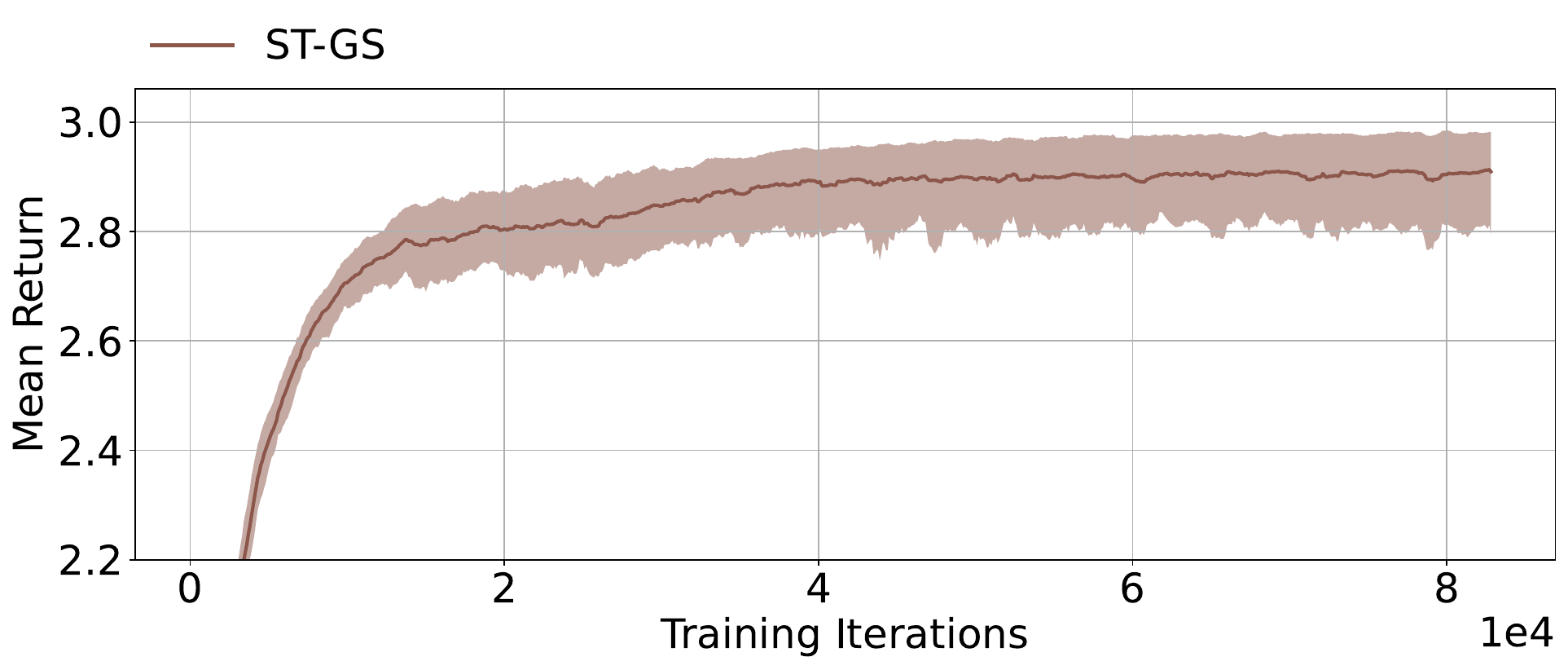}
    \caption{ST-GS in the simple matrix environment}
\end{figure*}

\clearpage
\subsubsection{Complex Matrix Environment}
\begin{figure*}[h]
    \centering
    \includegraphics[width=\graphwidth\linewidth]{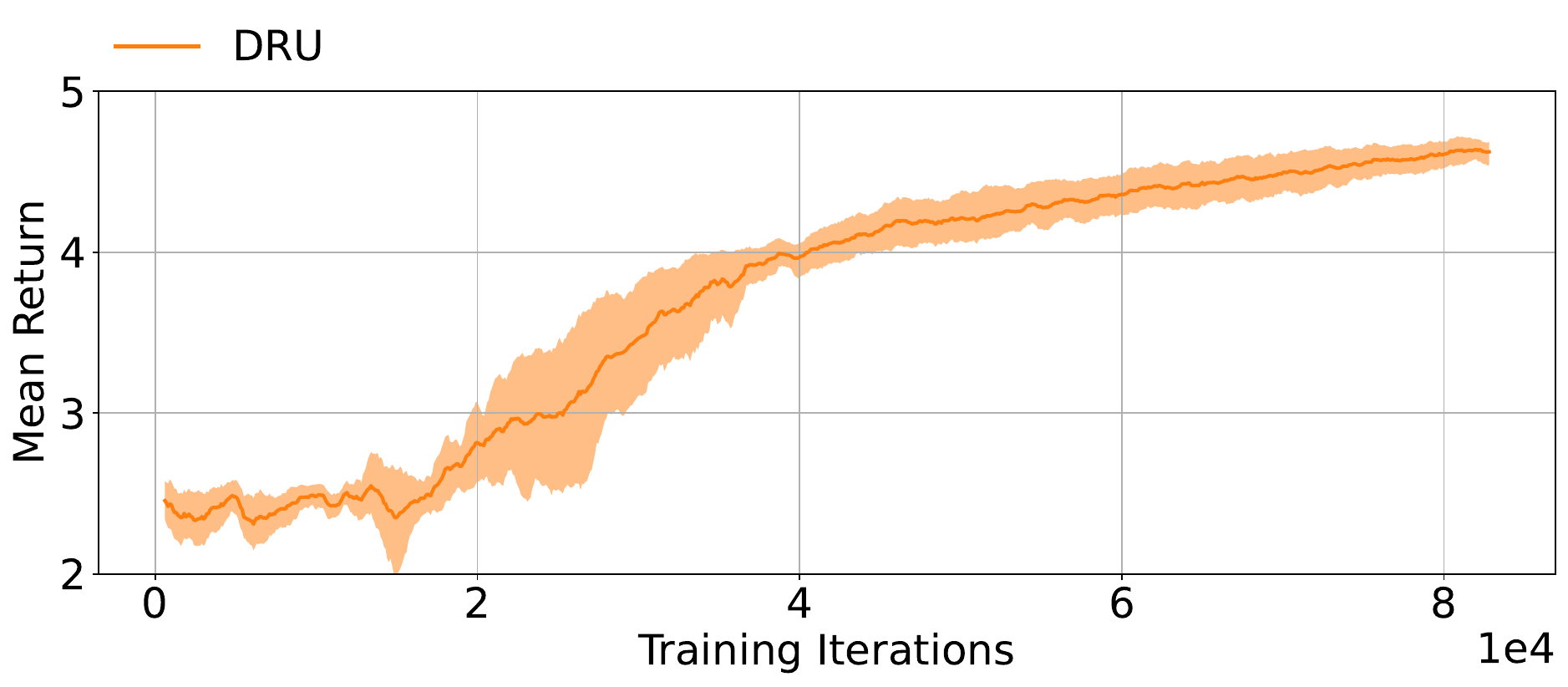}
    \caption{DRU in the complex matrix environment}
\end{figure*}
\begin{figure*}[h]
    \centering
    \includegraphics[width=\graphwidth\linewidth]{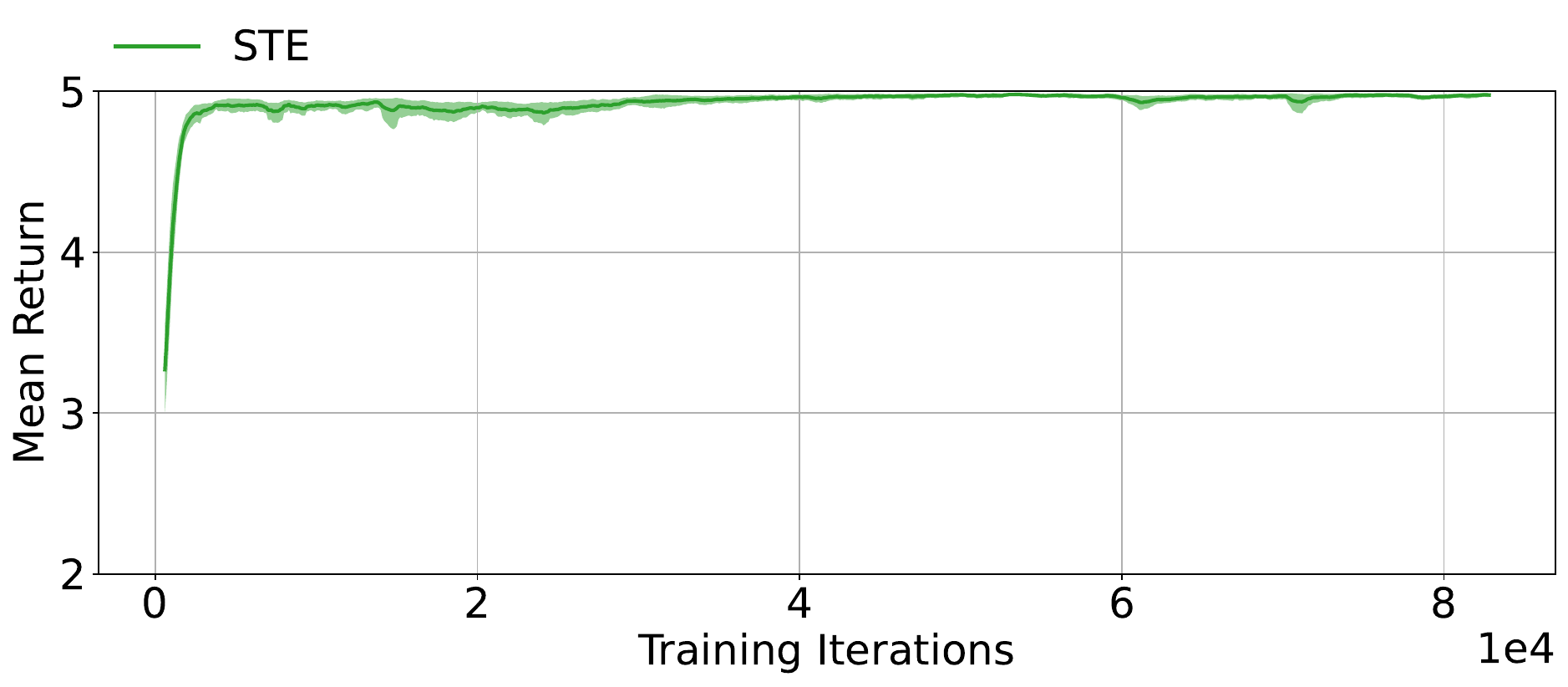}
    \caption{STE in the complex matrix environment}
\end{figure*}
\begin{figure*}[h]
    \centering
    \includegraphics[width=\graphwidth\linewidth]{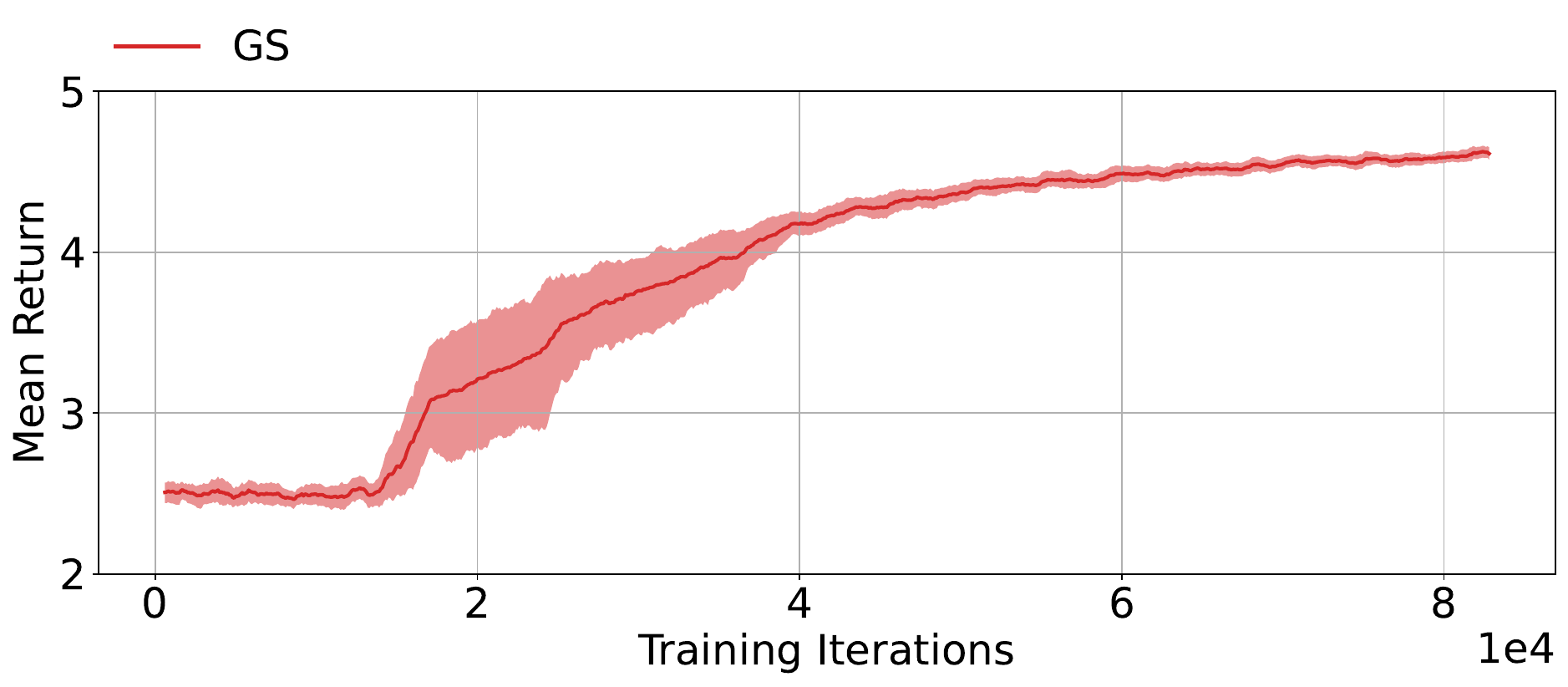}
    \caption{GS in the complex matrix environment}
\end{figure*}
\begin{figure*}[h]
    \centering
    \includegraphics[width=\graphwidth\linewidth]{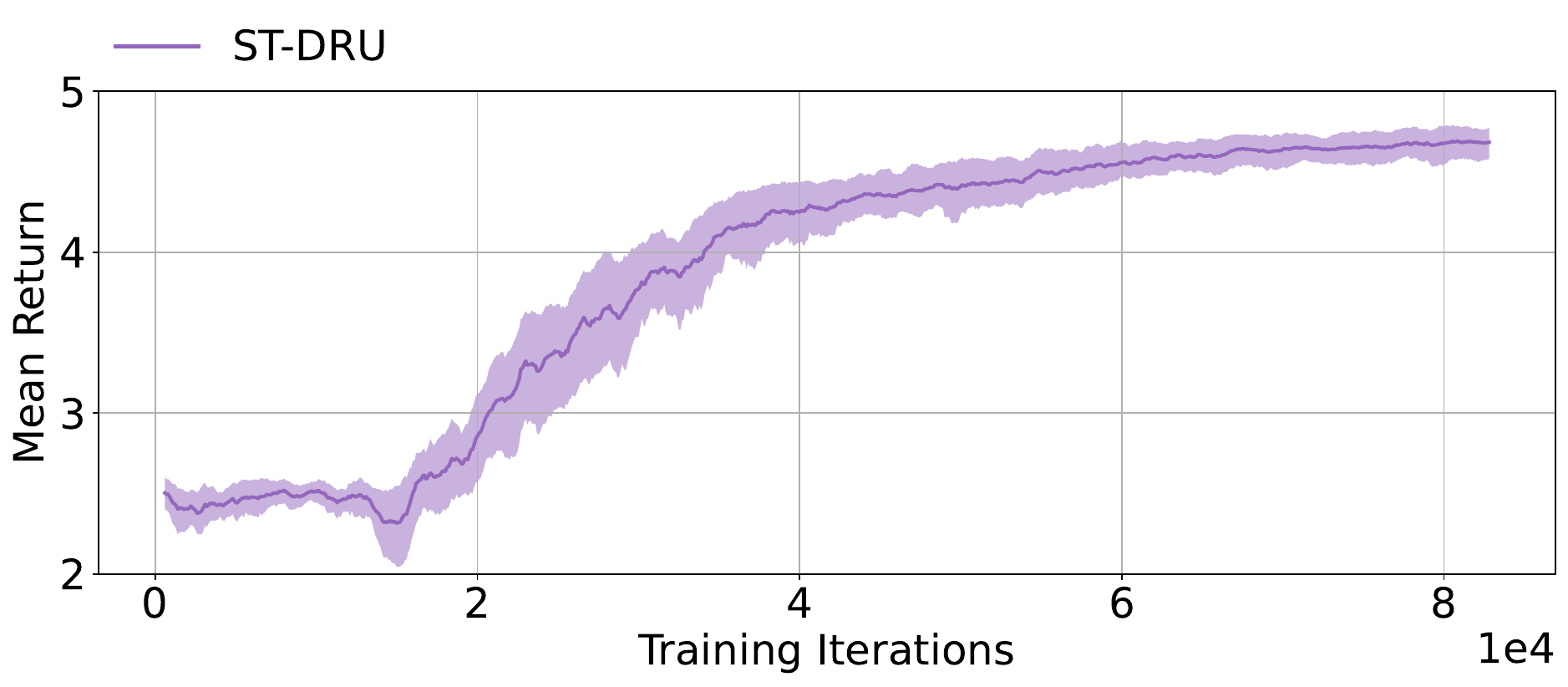}
    \caption{ST-DRU in the complex matrix environment}
\end{figure*}
\begin{figure*}[h]
    \centering
    \includegraphics[width=\graphwidth\linewidth]{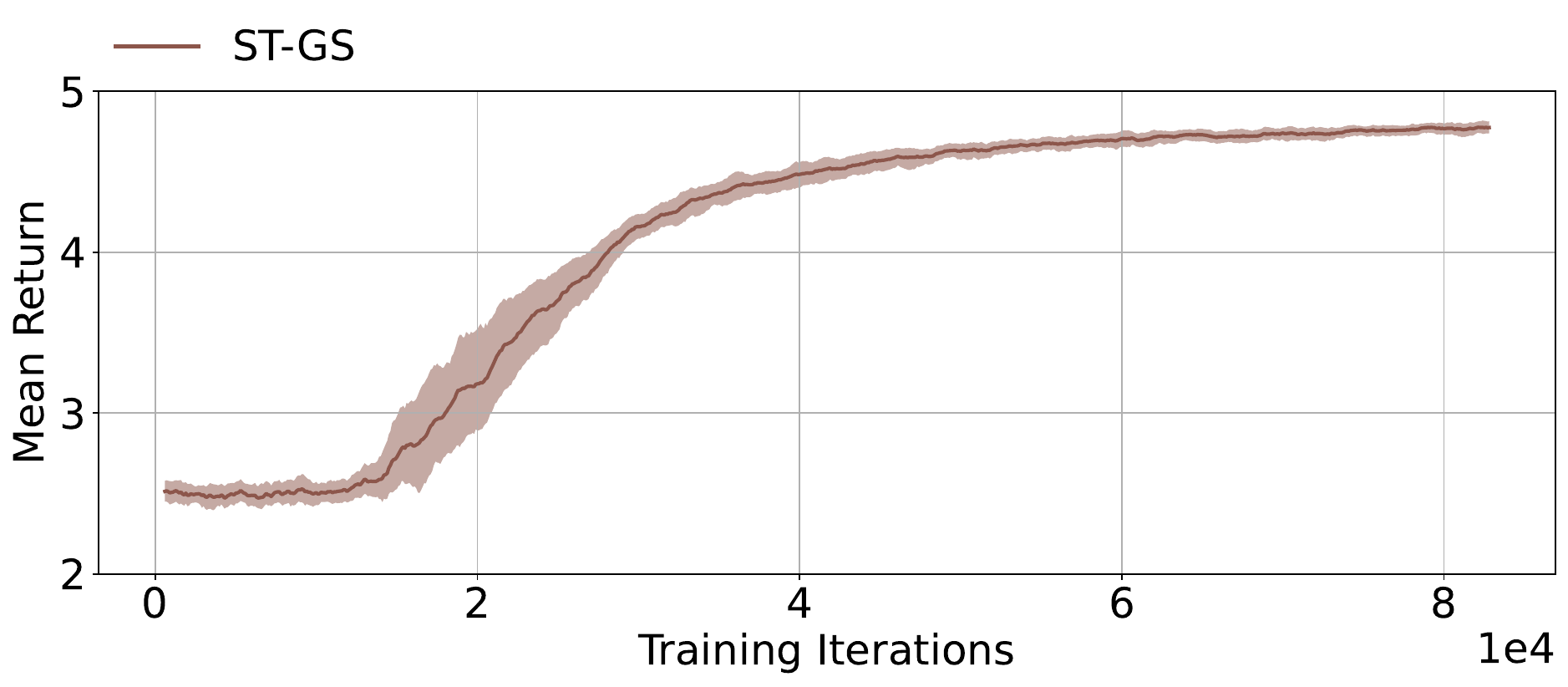}
    \caption{ST-GS in the complex matrix environment}
\end{figure*}

\clearpage
\subsection{Error Correction}
\begin{figure*}[h]
    \centering
    \includegraphics[width=\graphwidth\linewidth]{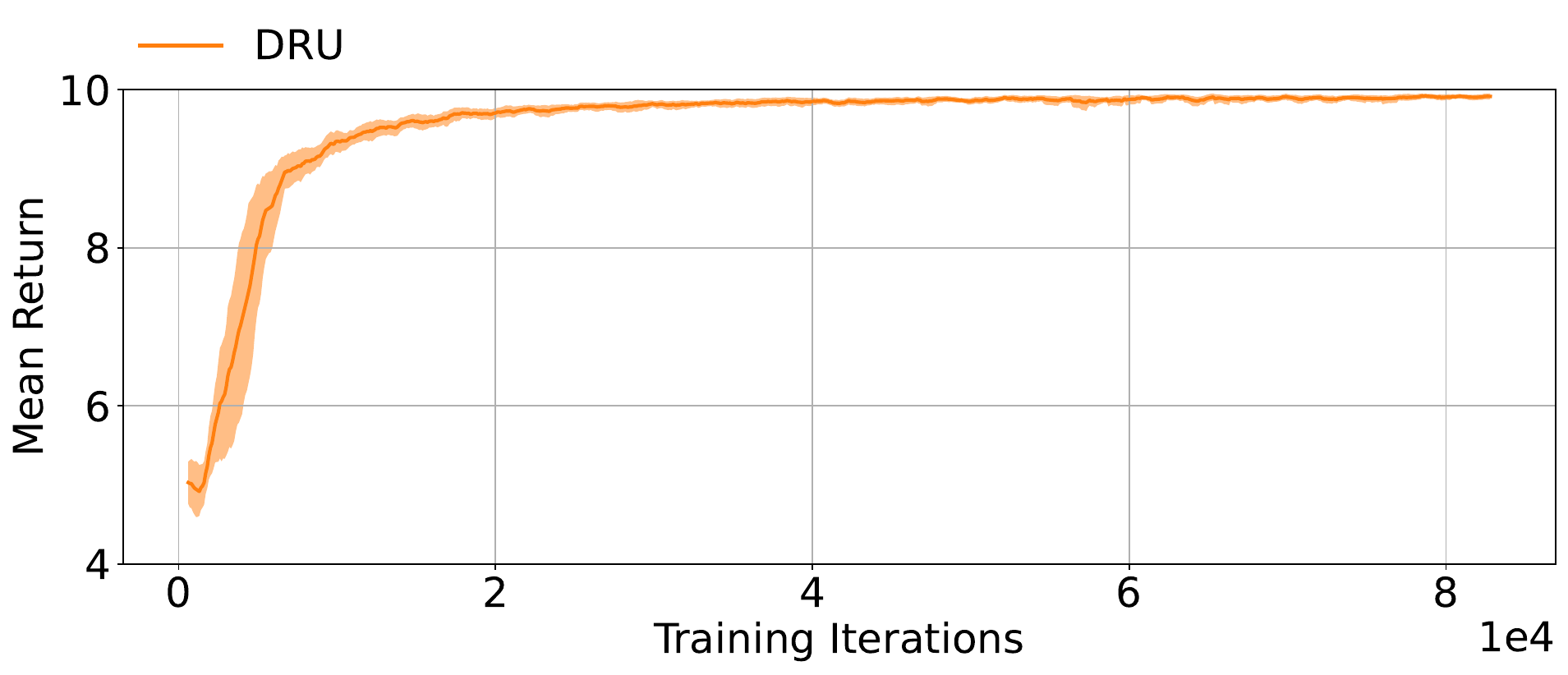}
    \caption{DRU in the matrix environment with a 50\% chance that a bit error
occurs at a random location in the message}
\end{figure*}
\begin{figure*}[h]
    \centering
    \includegraphics[width=\graphwidth\linewidth]{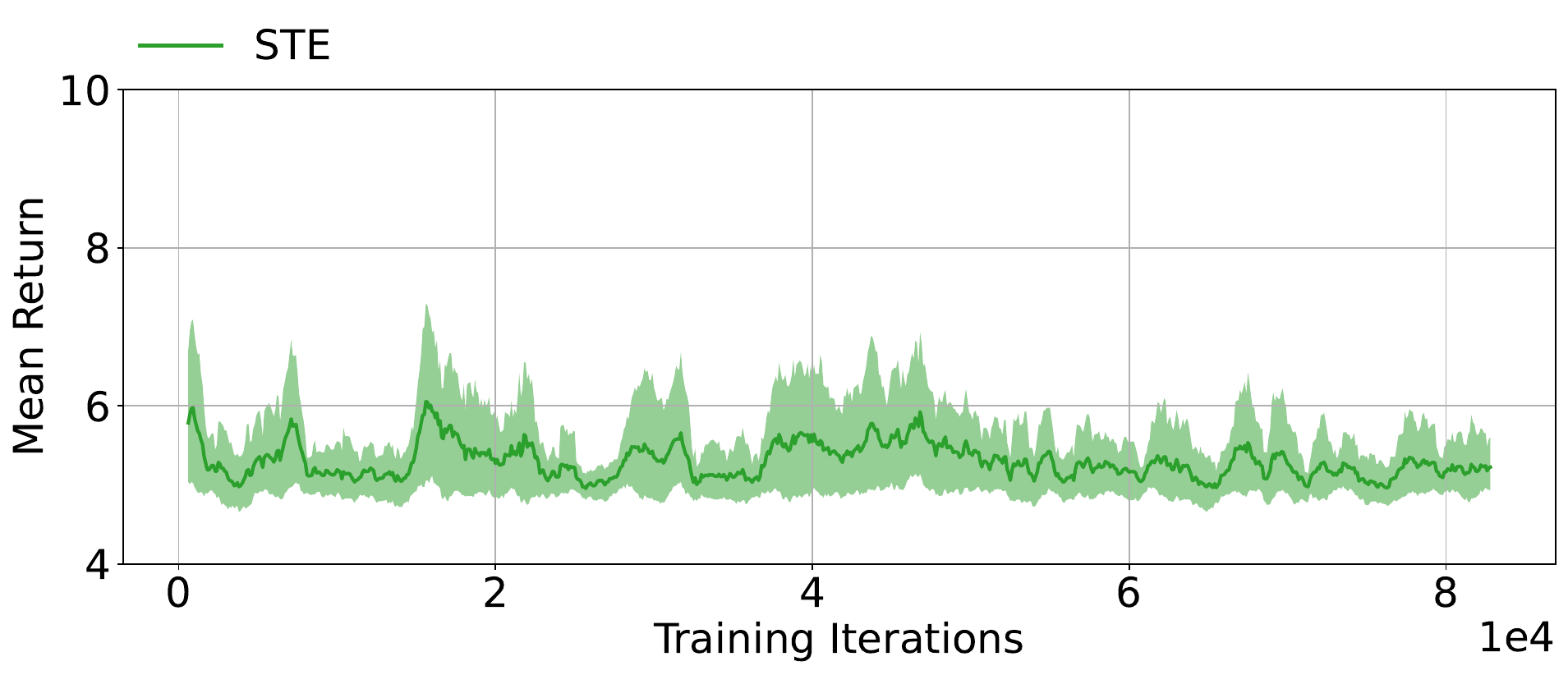}
    \caption{STE in the matrix environment with a 50\% chance that a bit error
occurs at a random location in the message}
\end{figure*}
\begin{figure*}[h]
    \centering
    \includegraphics[width=\graphwidth\linewidth]{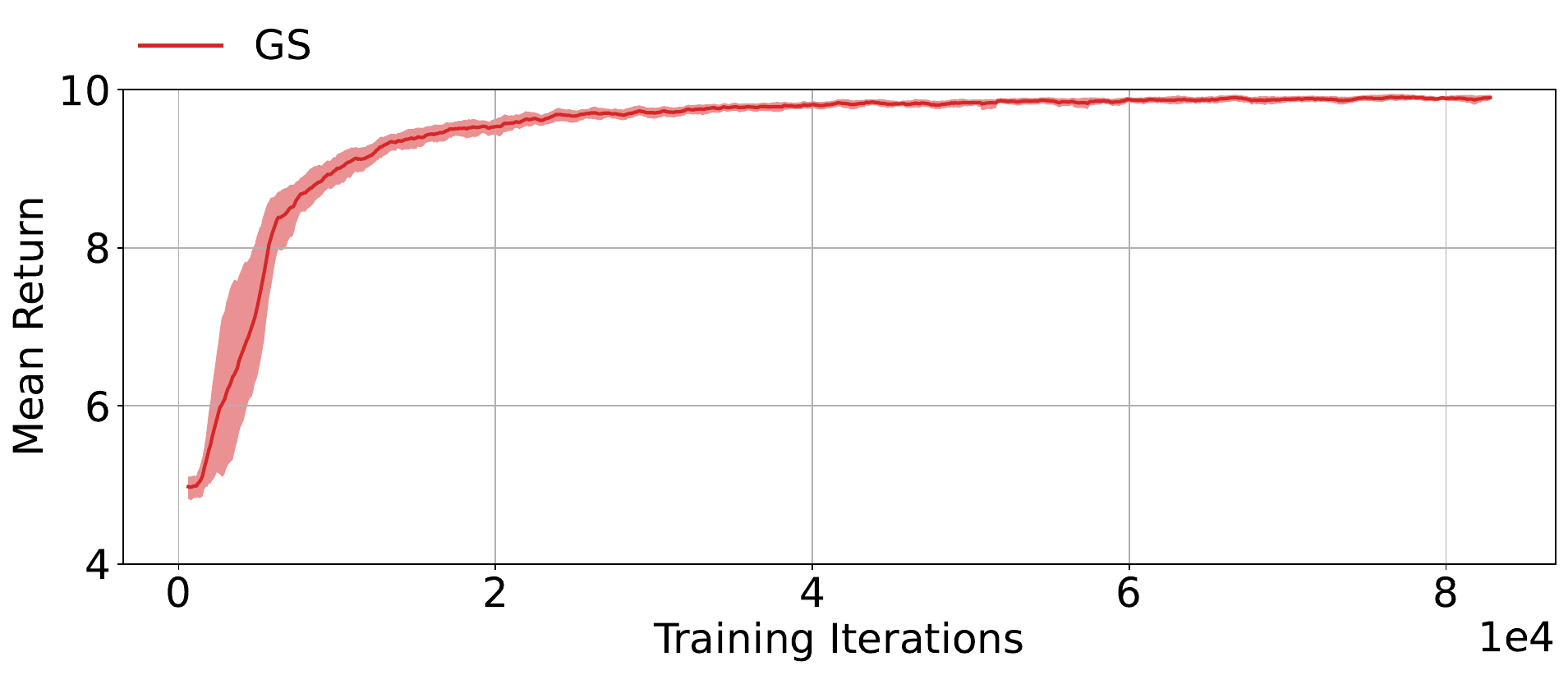}
    \caption{GS in the matrix environment with a 50\% chance that a bit error
occurs at a random location in the message}
\end{figure*}
\begin{figure*}[h]
    \centering
    \includegraphics[width=\graphwidth\linewidth]{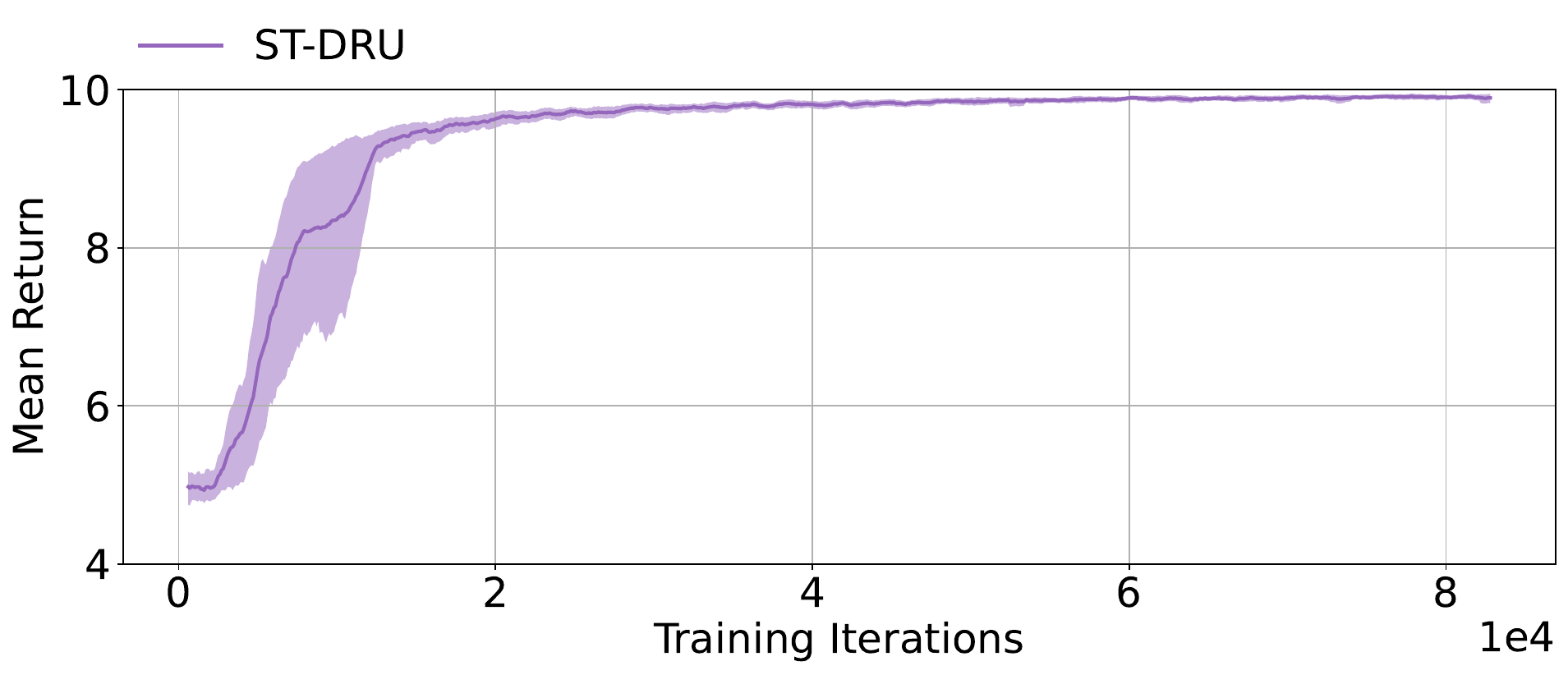}
    \caption{ST-DRU in the matrix environment with a 50\% chance that a bit error
will occur at a random location in the message}
\end{figure*}
\begin{figure*}[h]
    \centering
    \includegraphics[width=\graphwidth\linewidth]{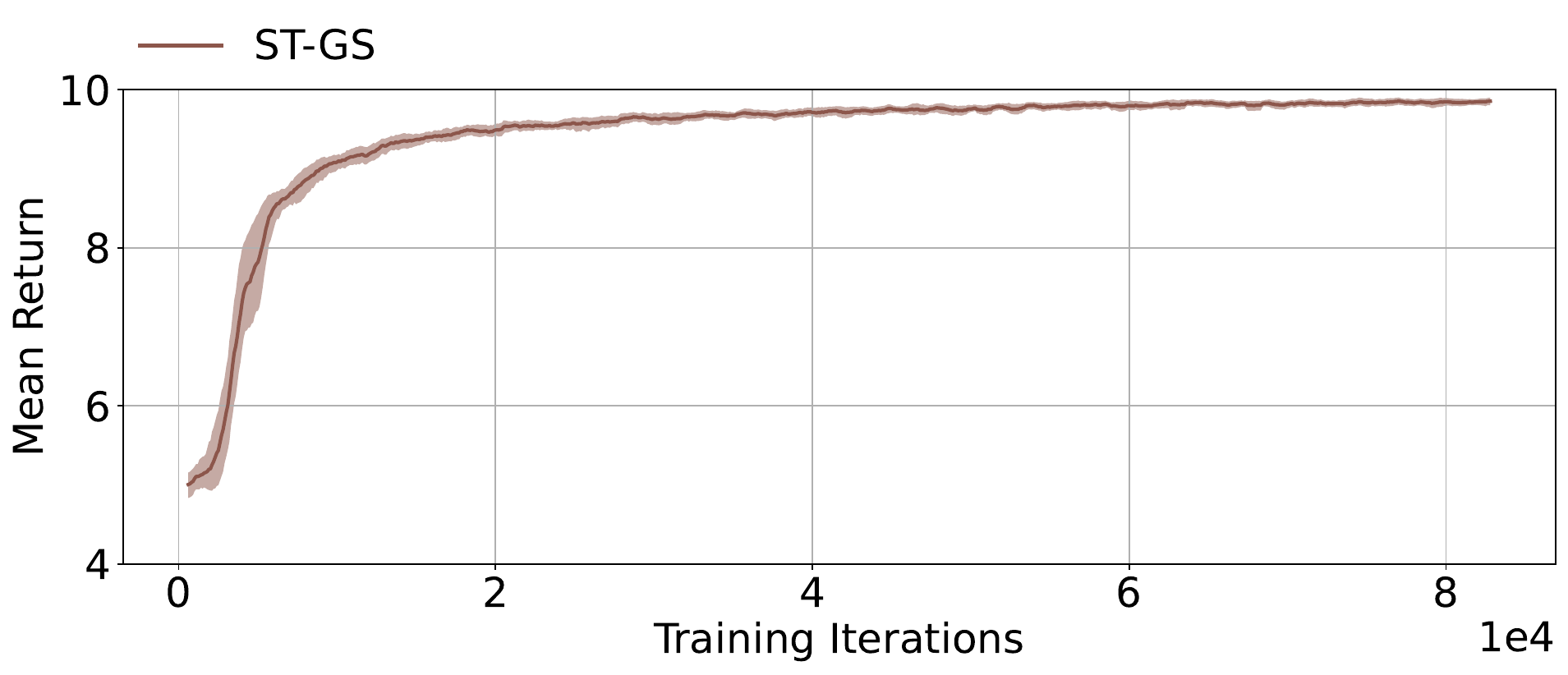}
    \caption{ST-GS in the matrix environment with a 50\% chance that a bit error
occurs at a random location in the message}
\end{figure*}

\clearpage
\subsection{Speaker Listener Environment}

\subsubsection{DIAL}
\begin{figure*}[h]
    \centering
    \includegraphics[width=\graphwidth\linewidth]{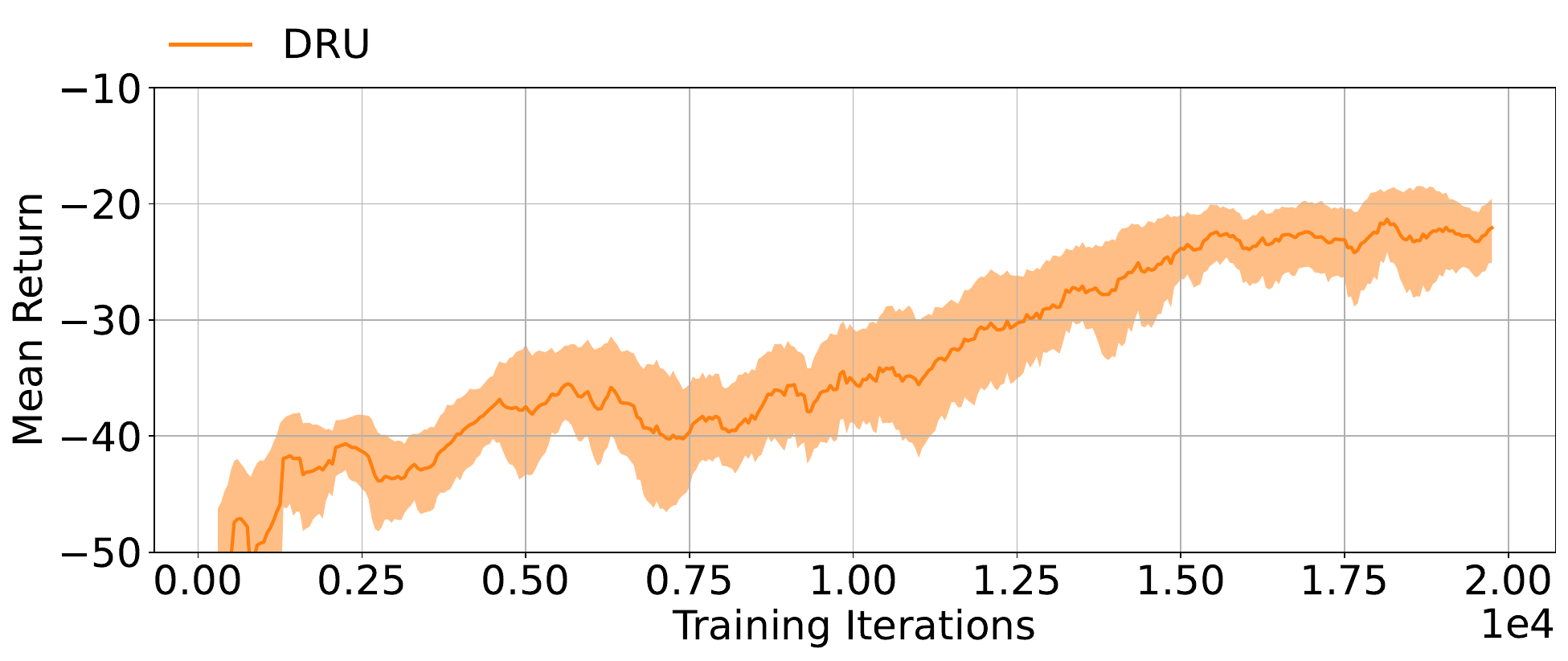}
    \caption{DRU in the speaker listener environment using DIAL}
\end{figure*}
\begin{figure*}[h]
    \centering
    \includegraphics[width=\graphwidth\linewidth]{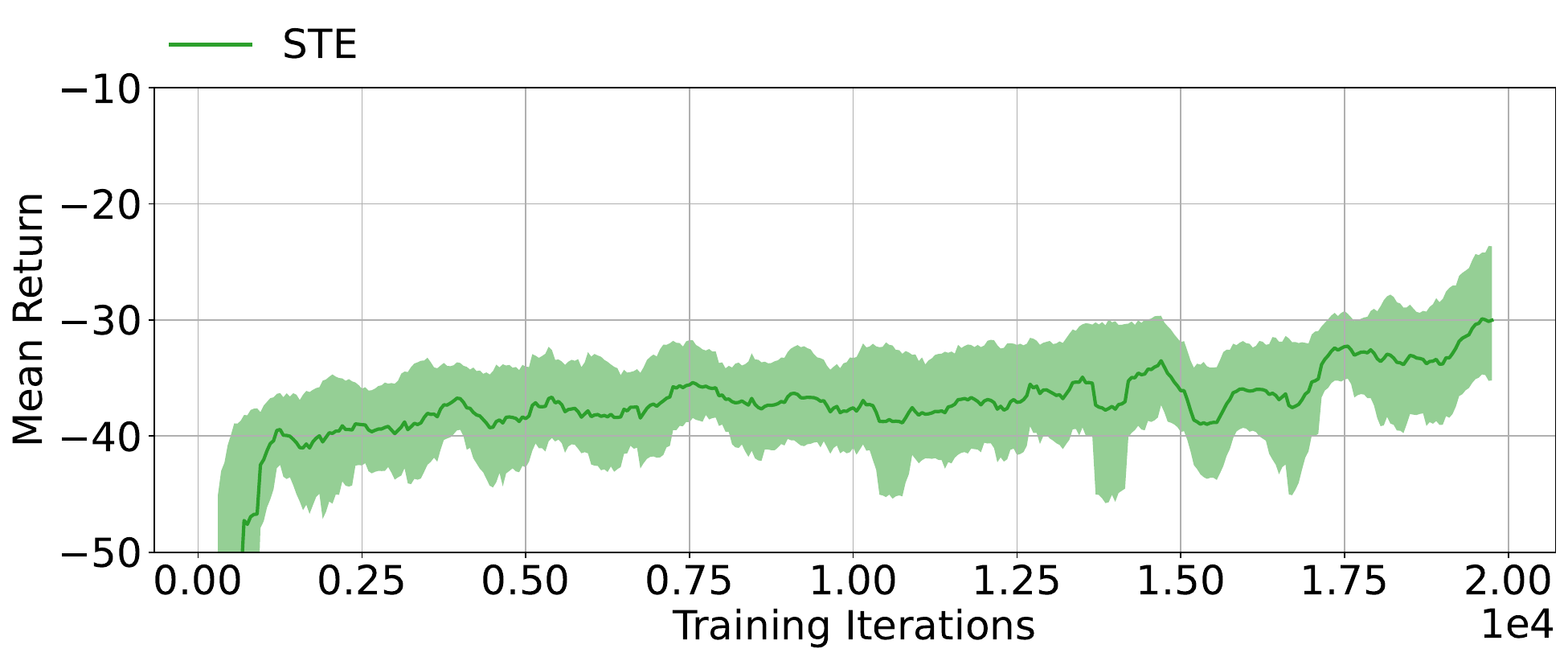}
    \caption{STE in the speaker listener environment using DIAL}
\end{figure*}
\begin{figure*}[h]
    \centering
    \includegraphics[width=\graphwidth\linewidth]{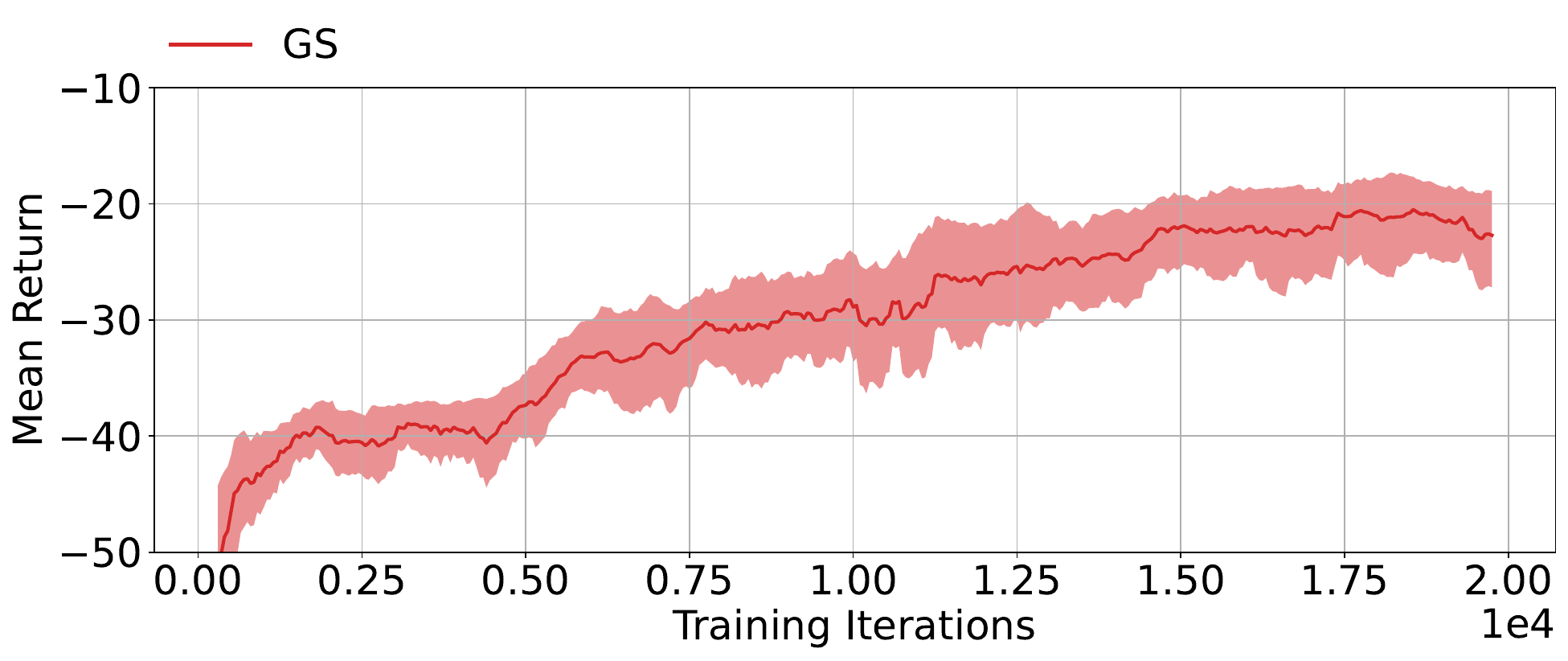}
    \caption{GS in the speaker listener environment using DIAL}
\end{figure*}
\begin{figure*}[h]
    \centering
    \includegraphics[width=\graphwidth\linewidth]{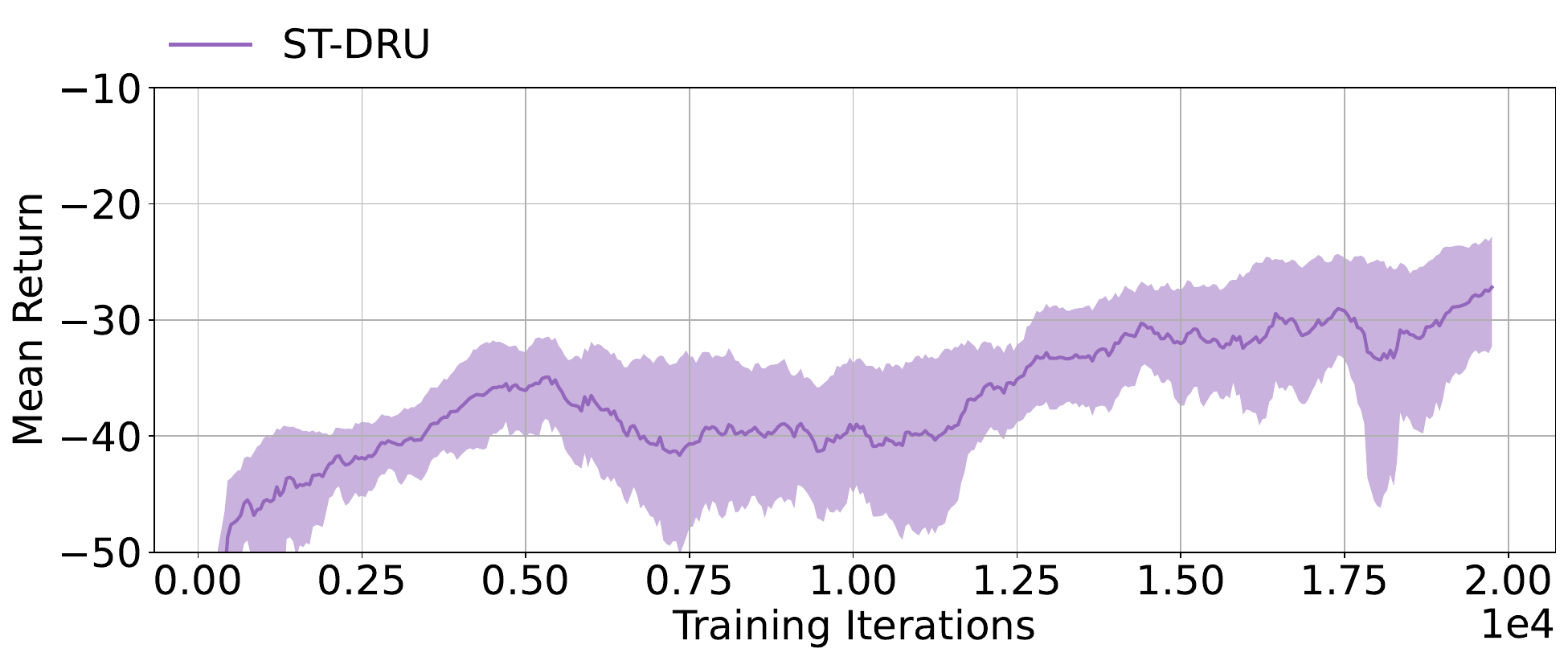}
    \caption{ST-DRU in the speaker listener environment using DIAL}
\end{figure*}
\begin{figure*}[h]
    \centering
    \includegraphics[width=\graphwidth\linewidth]{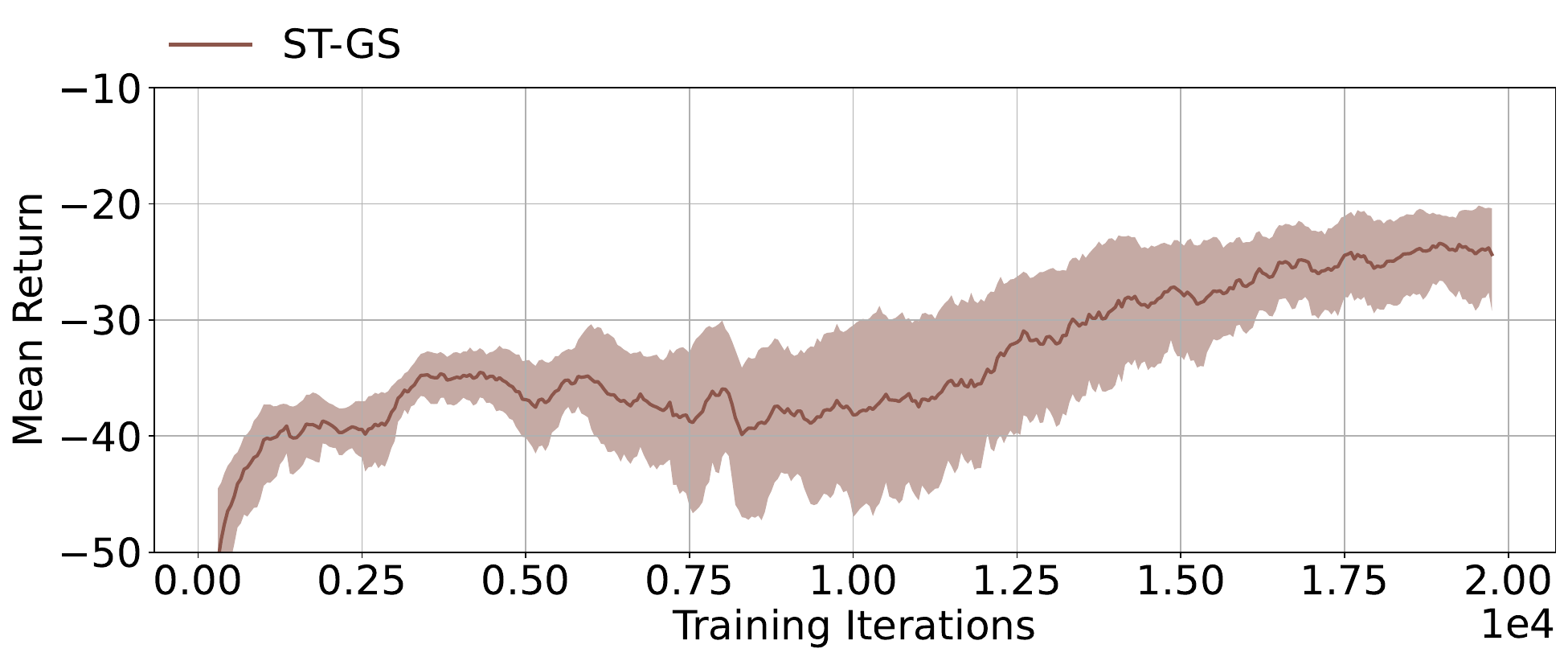}
    \caption{ST-GS in the speaker listener environment using DIAL}
\end{figure*}

\clearpage
\subsubsection{COMA-DIAL}
\begin{figure*}[h]
    \centering
    \includegraphics[width=\graphwidth\linewidth]{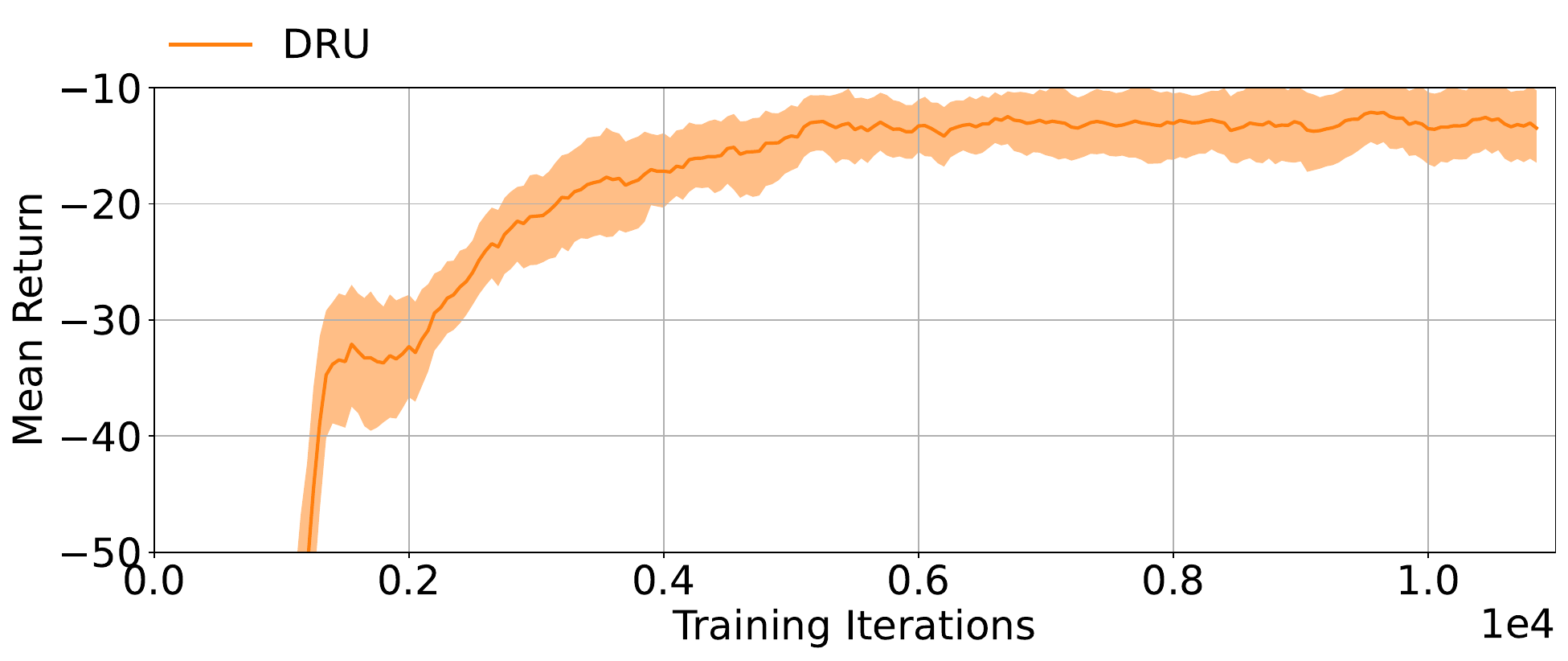}
    \caption{DRU in the speaker listener environment using COMA-DIAL}
\end{figure*}
\begin{figure*}[h]
    \centering
    \includegraphics[width=\graphwidth\linewidth]{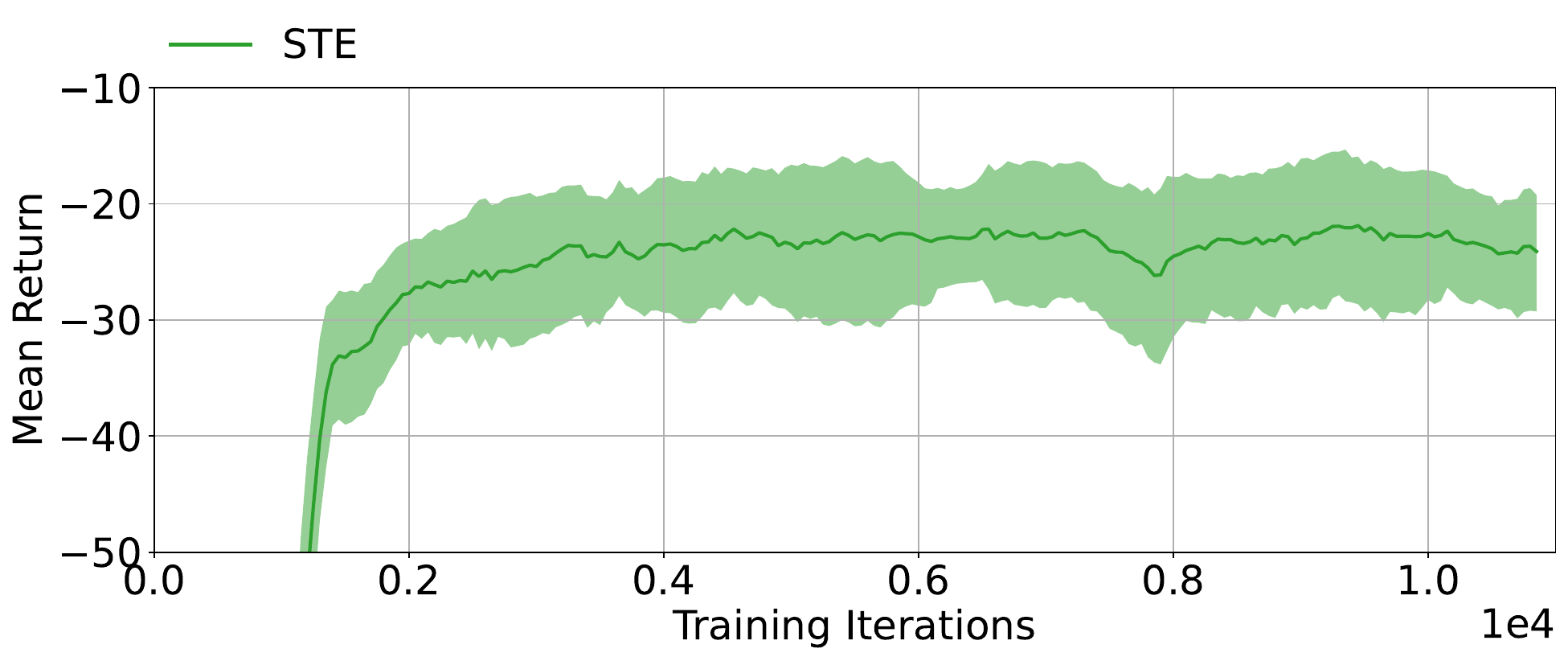}
    \caption{STE in the speaker listener environment using COMA-DIAL}
\end{figure*}
\begin{figure*}[h]
    \centering
    \includegraphics[width=\graphwidth\linewidth]{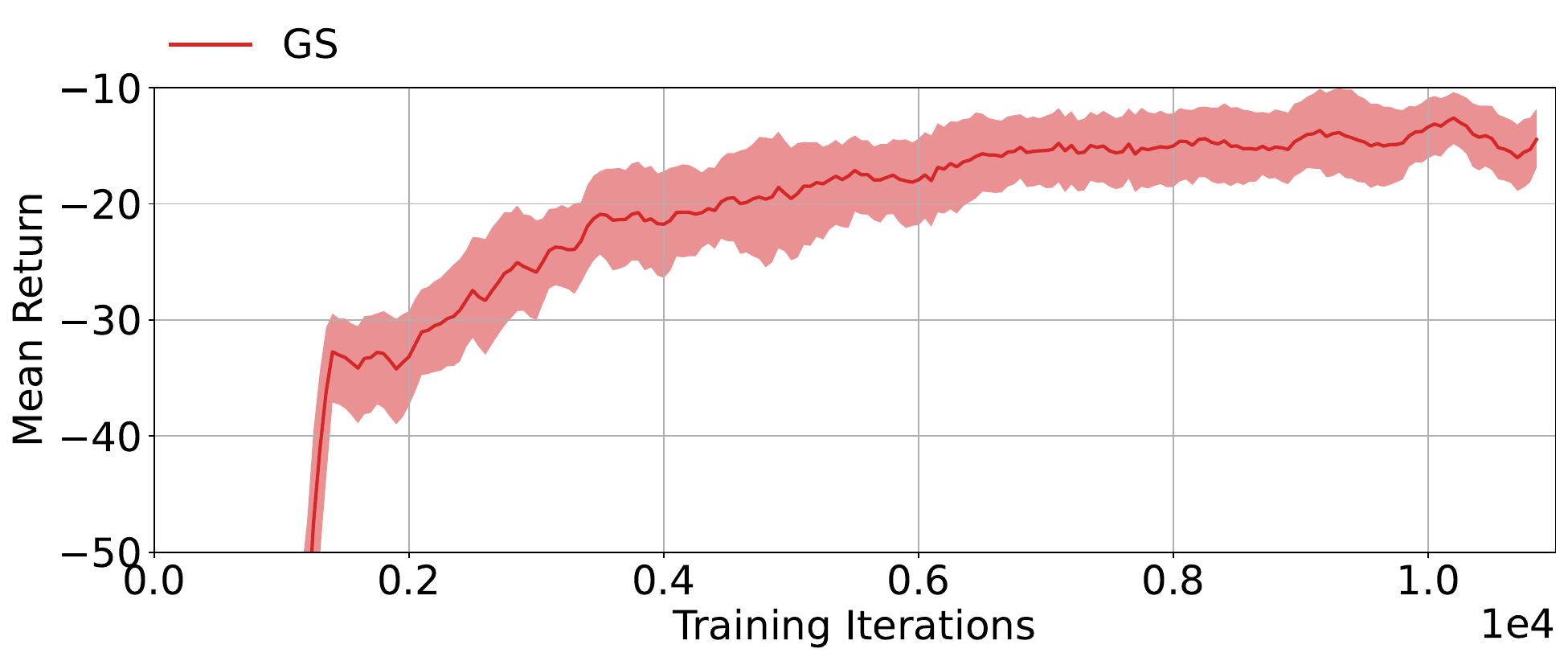}
    \caption{GS in the speaker listener environment using COMA-DIAL}
\end{figure*}
\begin{figure*}[h]
    \centering
    \includegraphics[width=\graphwidth\linewidth]{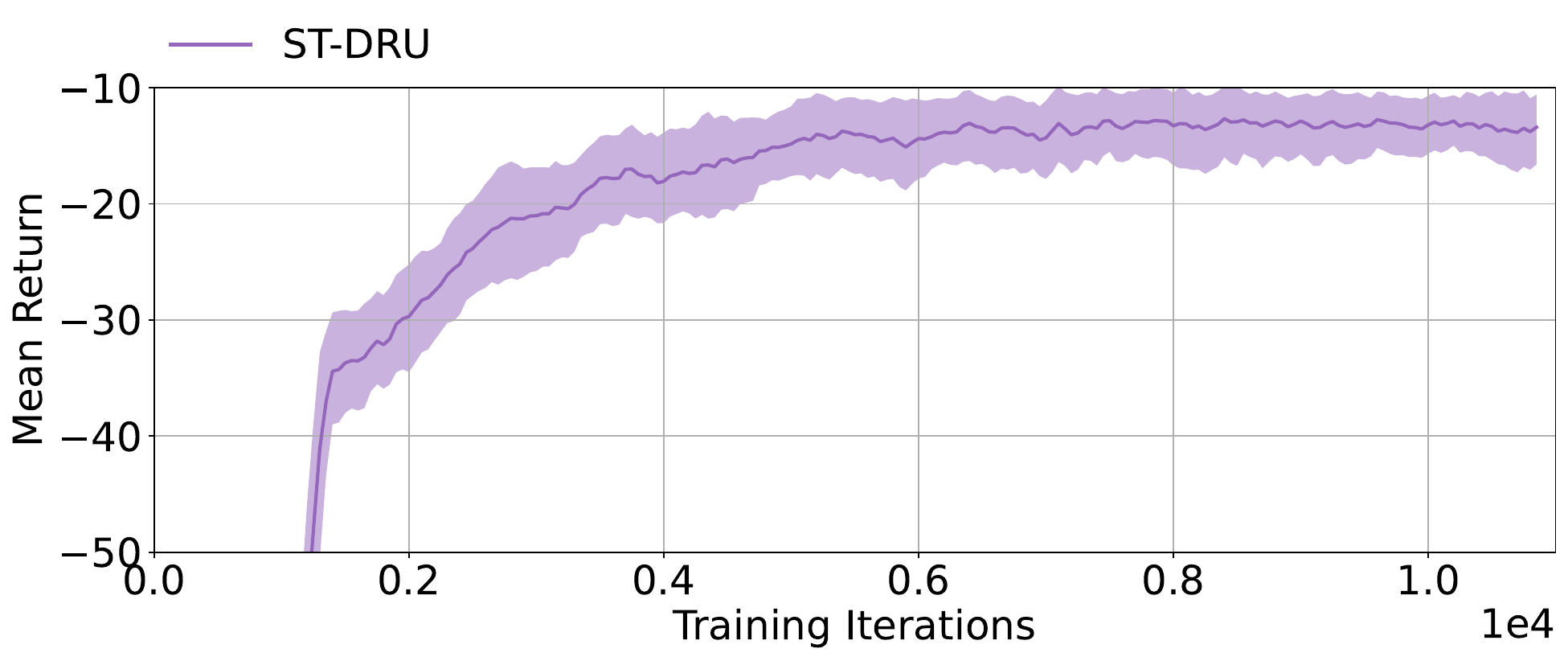}
    \caption{ST-DRU in the speaker listener environment using COMA-DIAL}
\end{figure*}
\begin{figure*}[h]
    \centering
    \includegraphics[width=\graphwidth\linewidth]{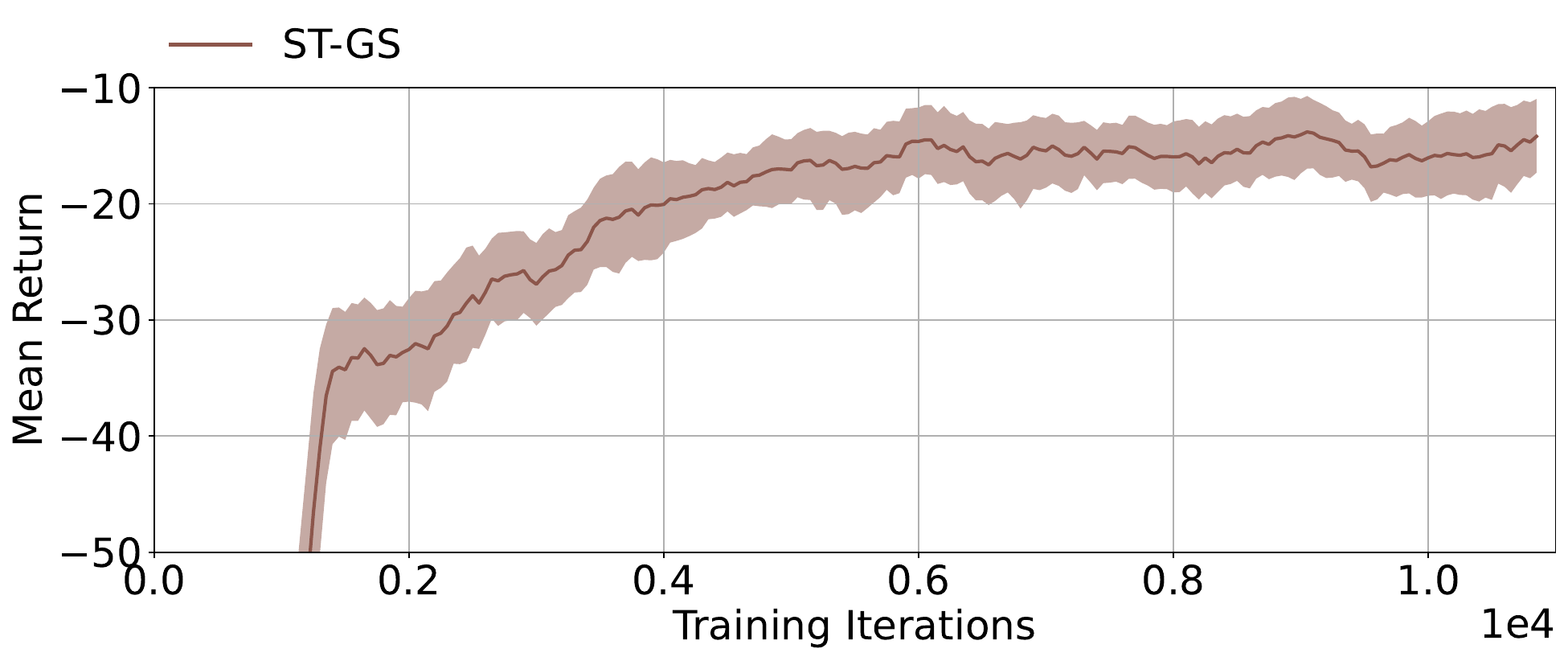}
    \caption{ST-GS in the speaker listener environment using COMA-DIAL}
\end{figure*}

\clearpage
\subsection{Simple Reference Environment}
\begin{figure*}[h]
    \centering
    \includegraphics[width=\graphwidth\linewidth]{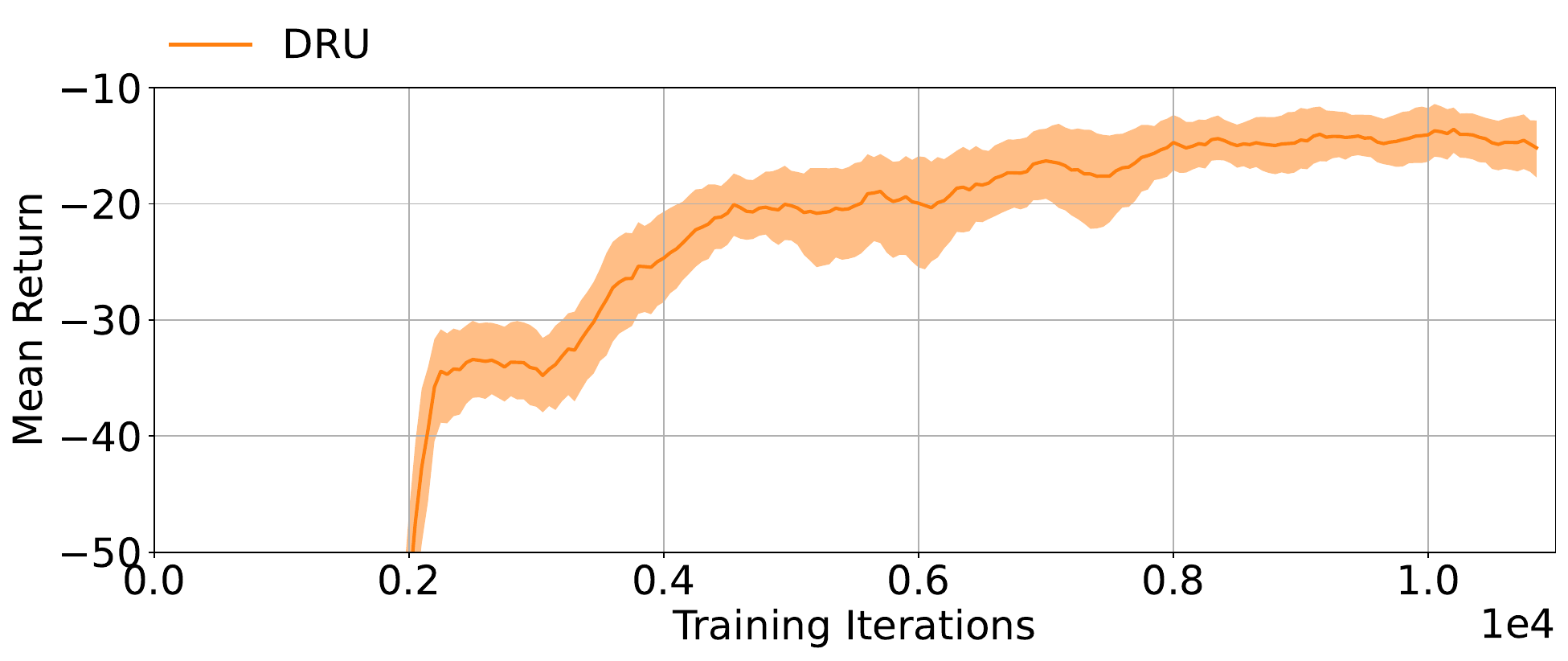}
    \caption{DRU in the simple reference environment}
\end{figure*}
\begin{figure*}[h]
    \centering
    \includegraphics[width=\graphwidth\linewidth]{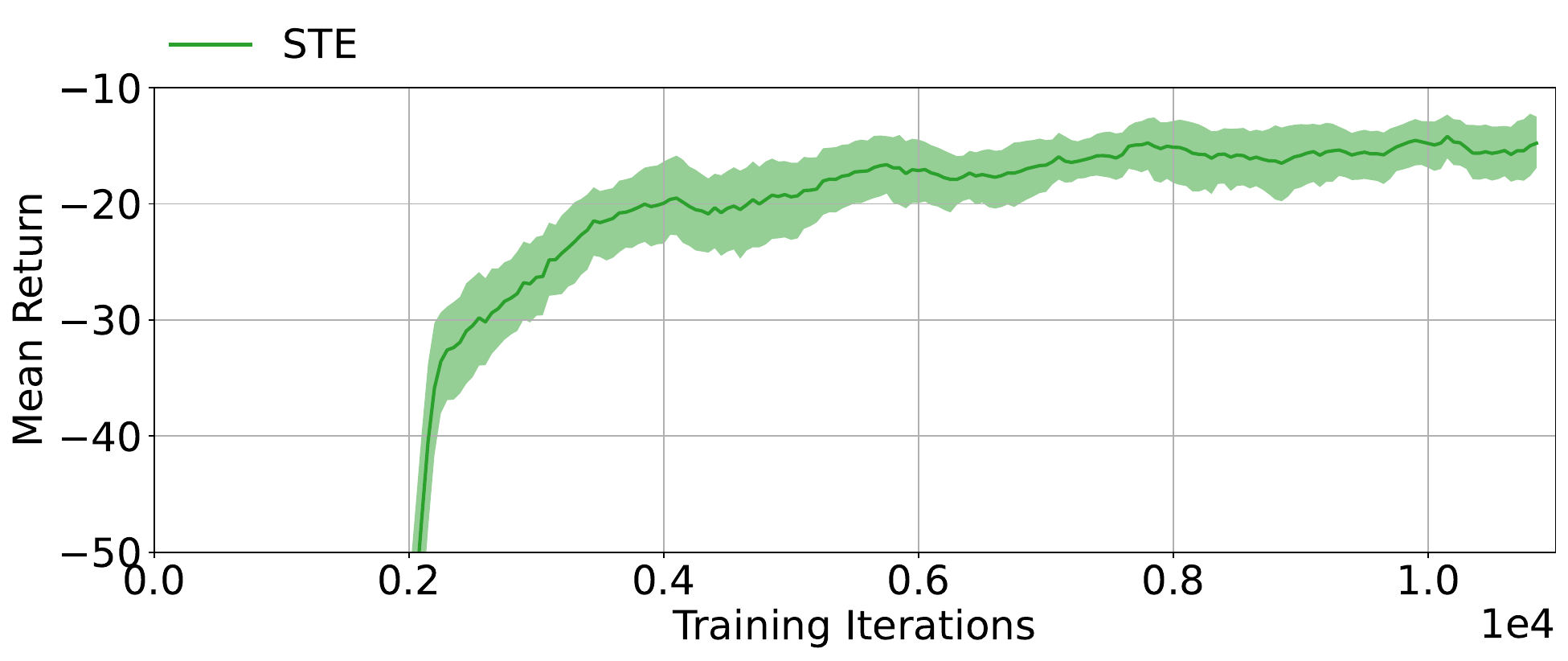}
    \caption{STE in the simple reference environment}
\end{figure*}
\begin{figure*}[h]
    \centering
    \includegraphics[width=\graphwidth\linewidth]{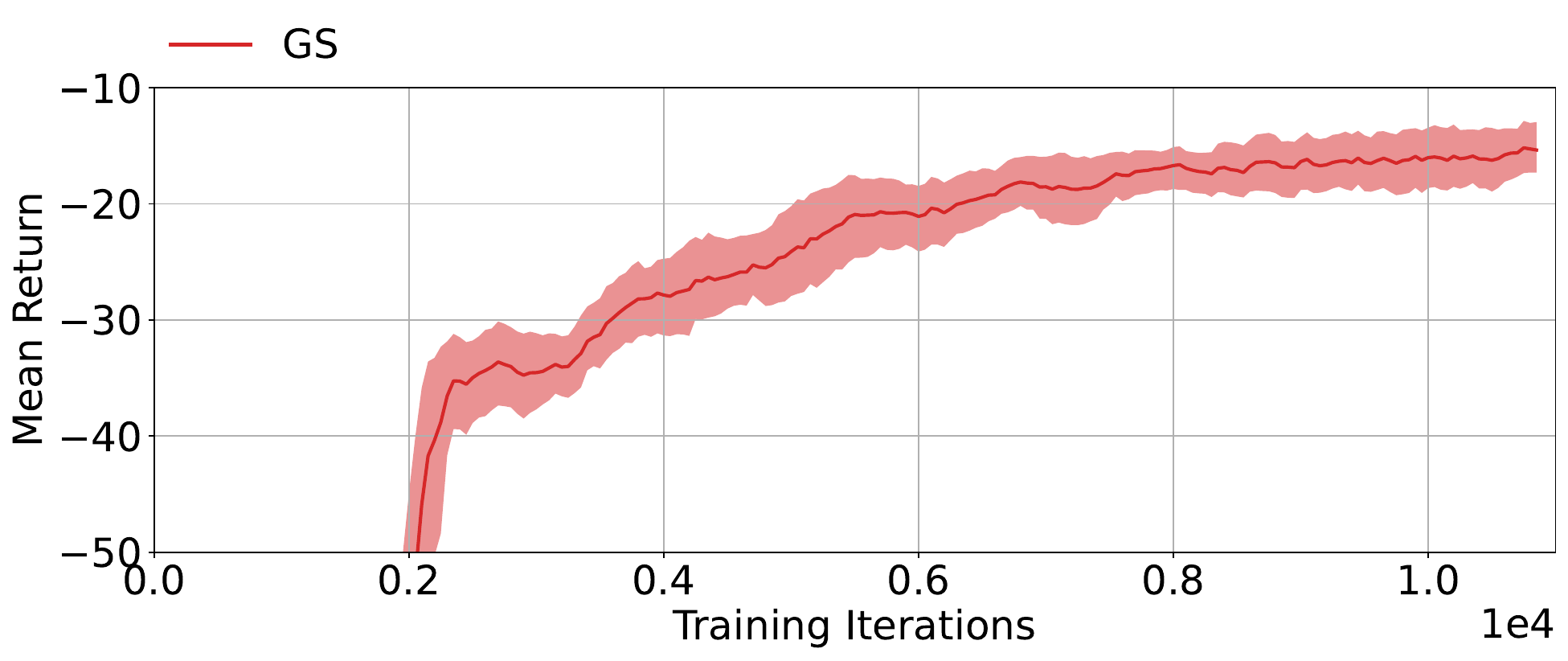}
    \caption{GS in the simple reference environment}
\end{figure*}
\begin{figure*}[h]
    \centering
    \includegraphics[width=\graphwidth\linewidth]{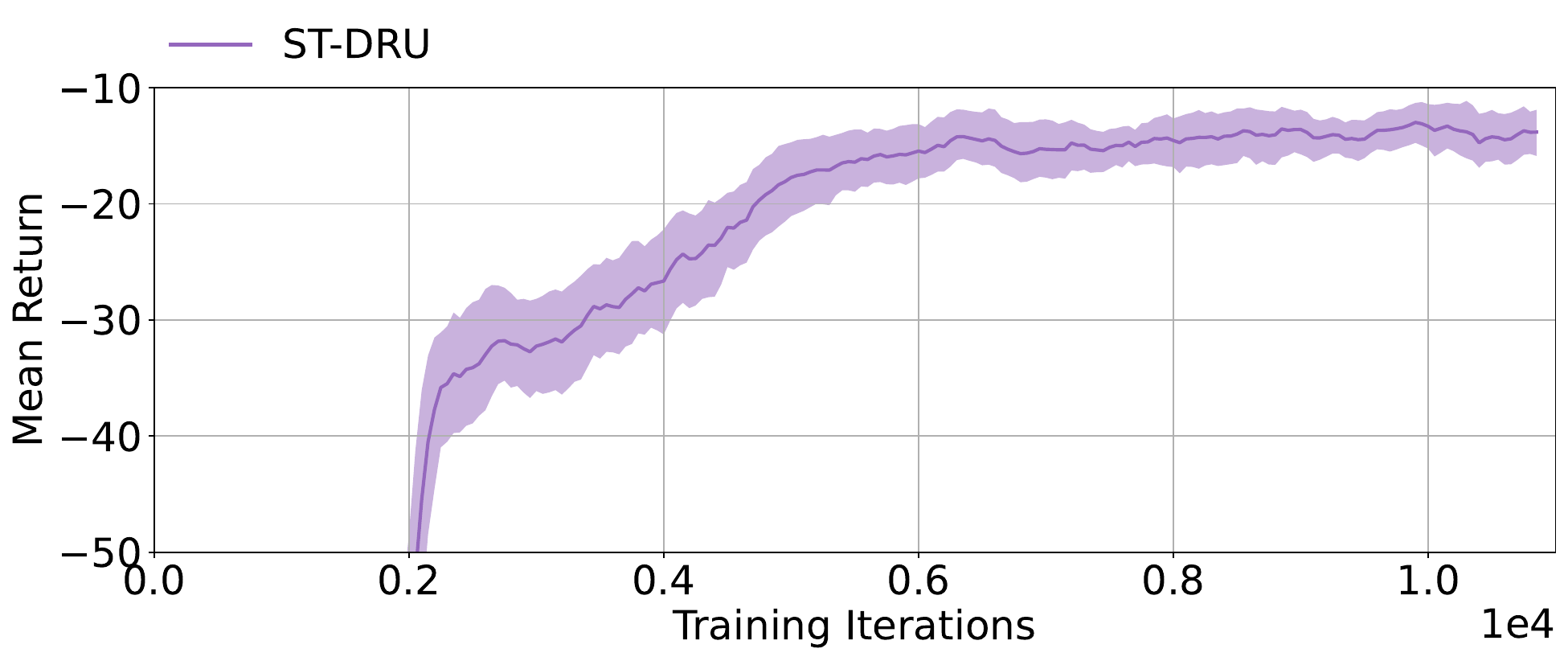}
    \caption{ST-DRU in the simple reference environment}
\end{figure*}
\begin{figure*}[h]
    \centering
    \includegraphics[width=\graphwidth\linewidth]{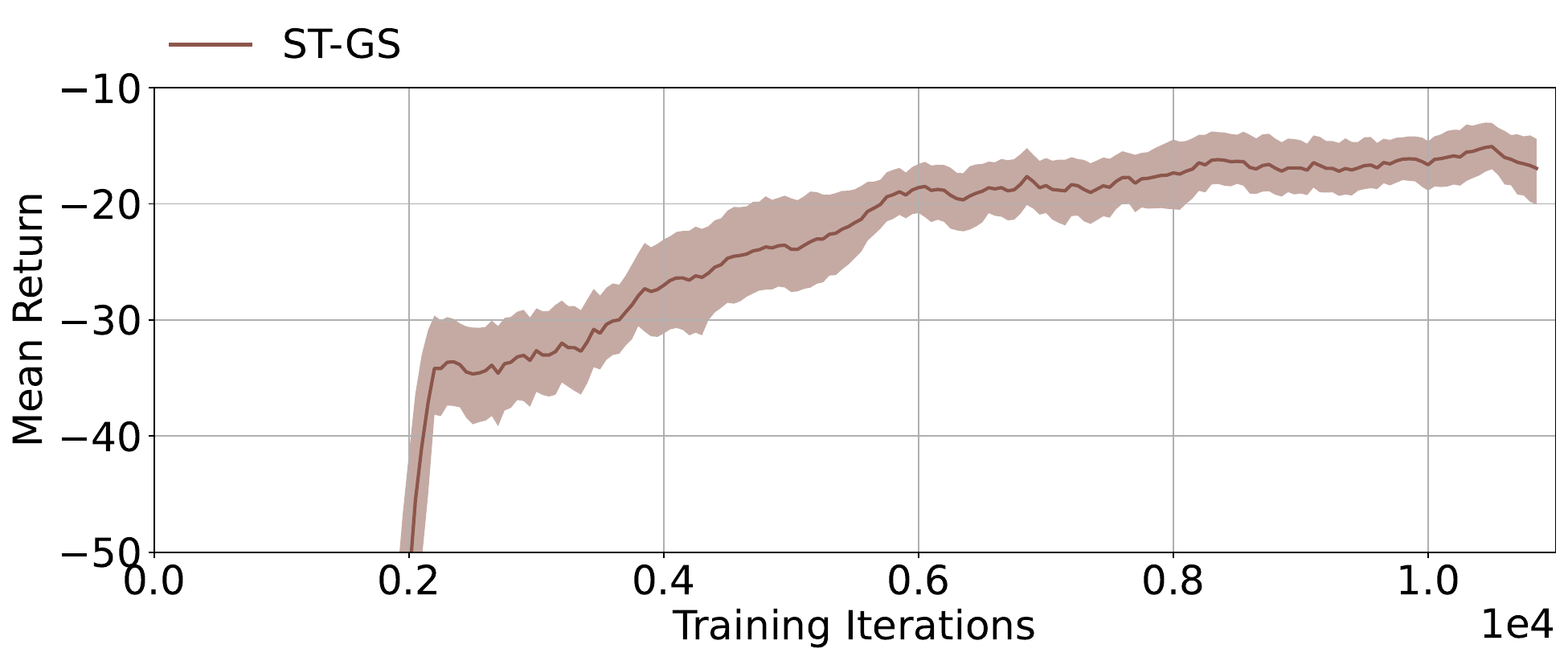}
    \caption{ST-GS in the simple reference environment}
\end{figure*}

\clearpage
\subsection{Parallel Speaker Listener Environment}
\begin{figure*}[h]
    \centering
    \includegraphics[width=\graphwidth\linewidth]{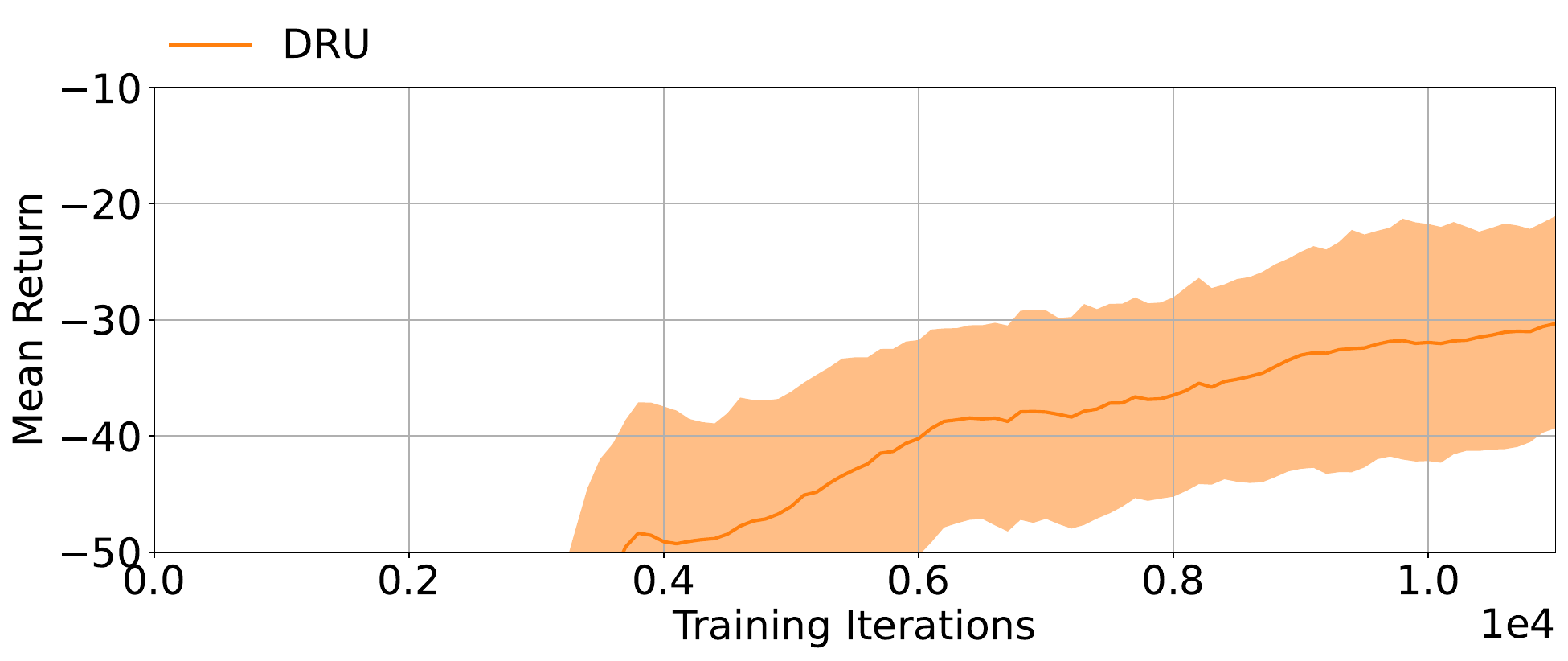}
    \caption{DRU in the parallel speaker listener environment}
\end{figure*}
\begin{figure*}[h]
    \centering
    \includegraphics[width=\graphwidth\linewidth]{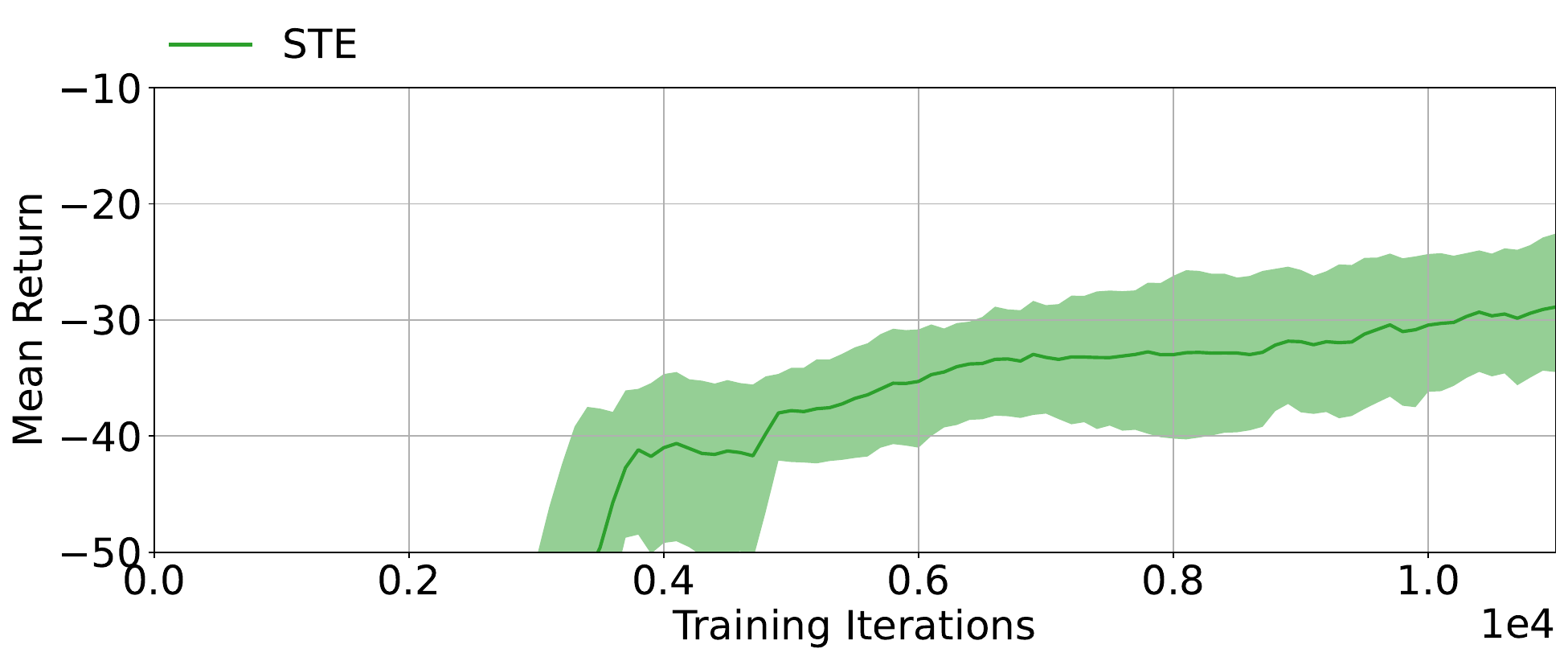}
    \caption{STE in the parallel speaker listener environment}
\end{figure*}
\begin{figure*}[h]
    \centering
    \includegraphics[width=\graphwidth\linewidth]{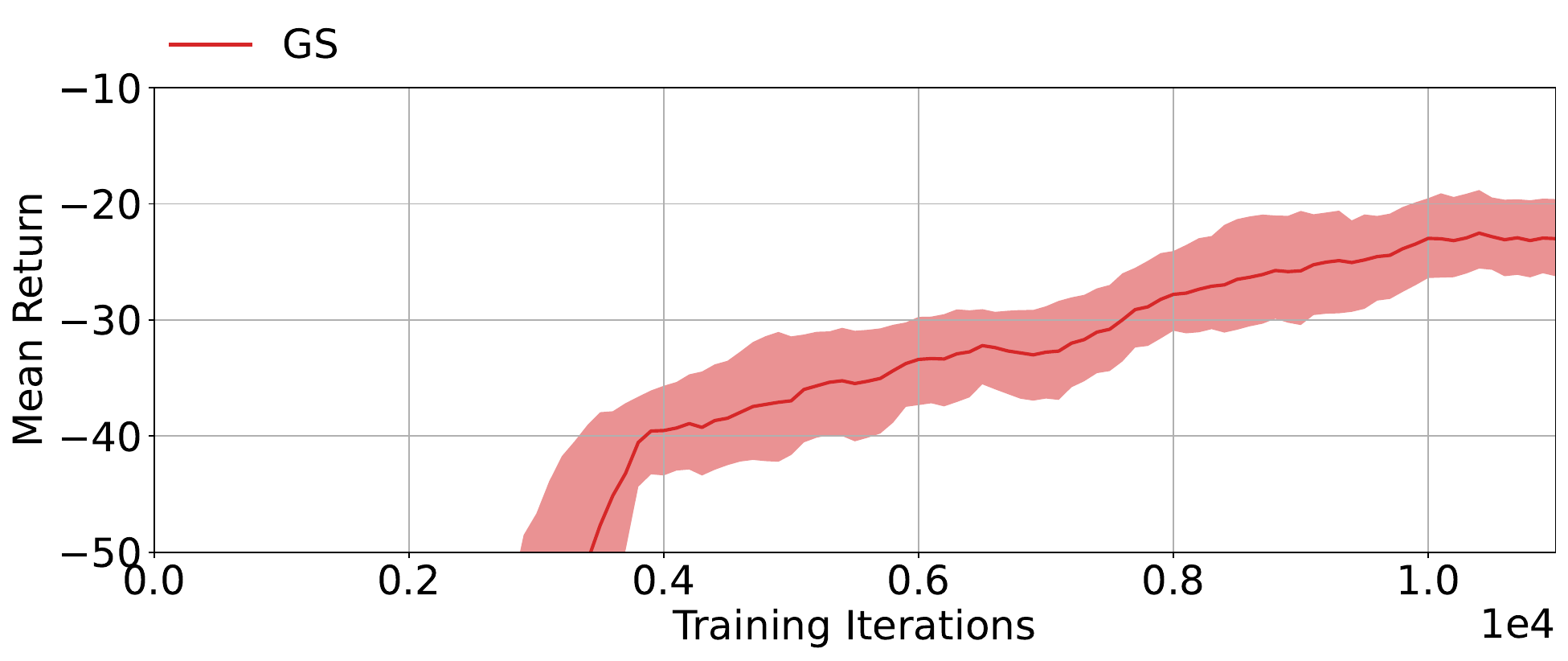}
    \caption{GS in the parallel speaker listener environment}
\end{figure*}
\begin{figure*}[h]
    \centering
    \includegraphics[width=\graphwidth\linewidth]{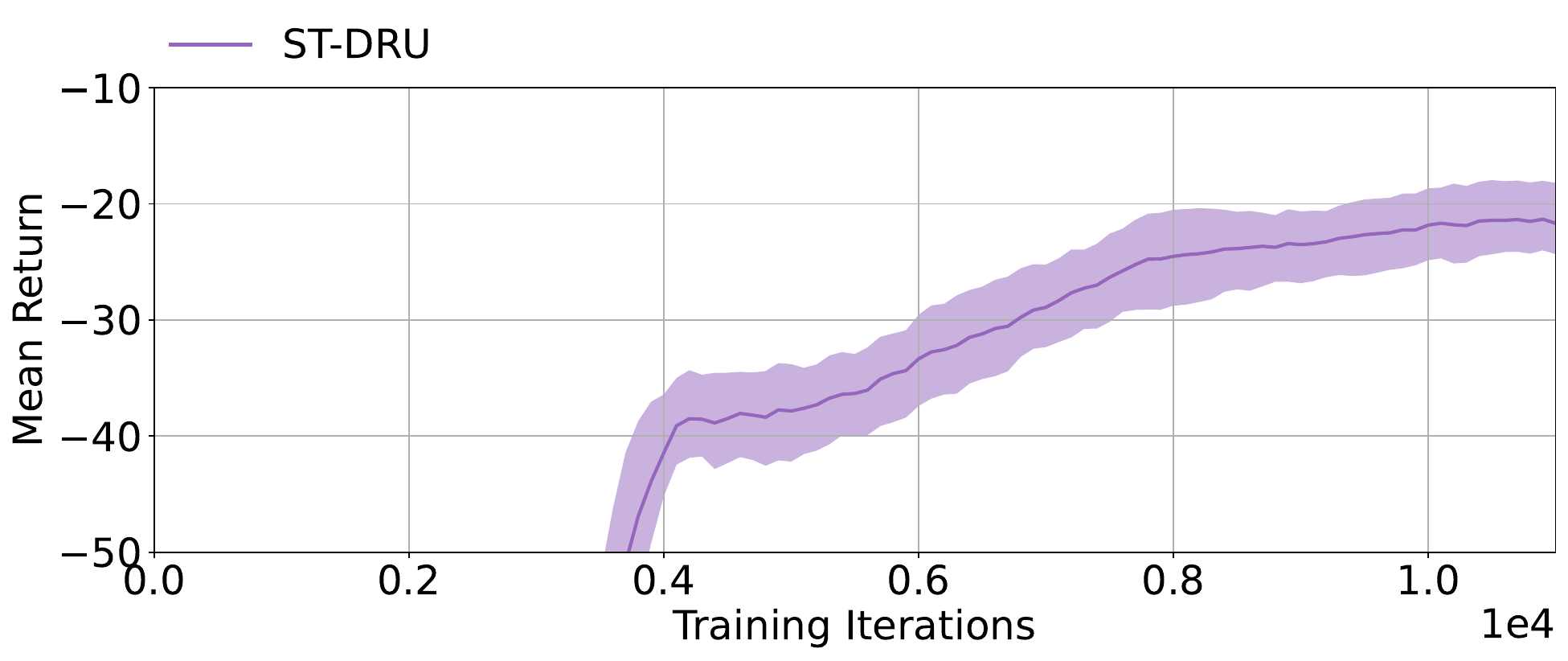}
    \caption{ST-DRU in the parallel speaker listener environment}
\end{figure*}
\begin{figure*}[h]
    \centering
    \includegraphics[width=\graphwidth\linewidth]{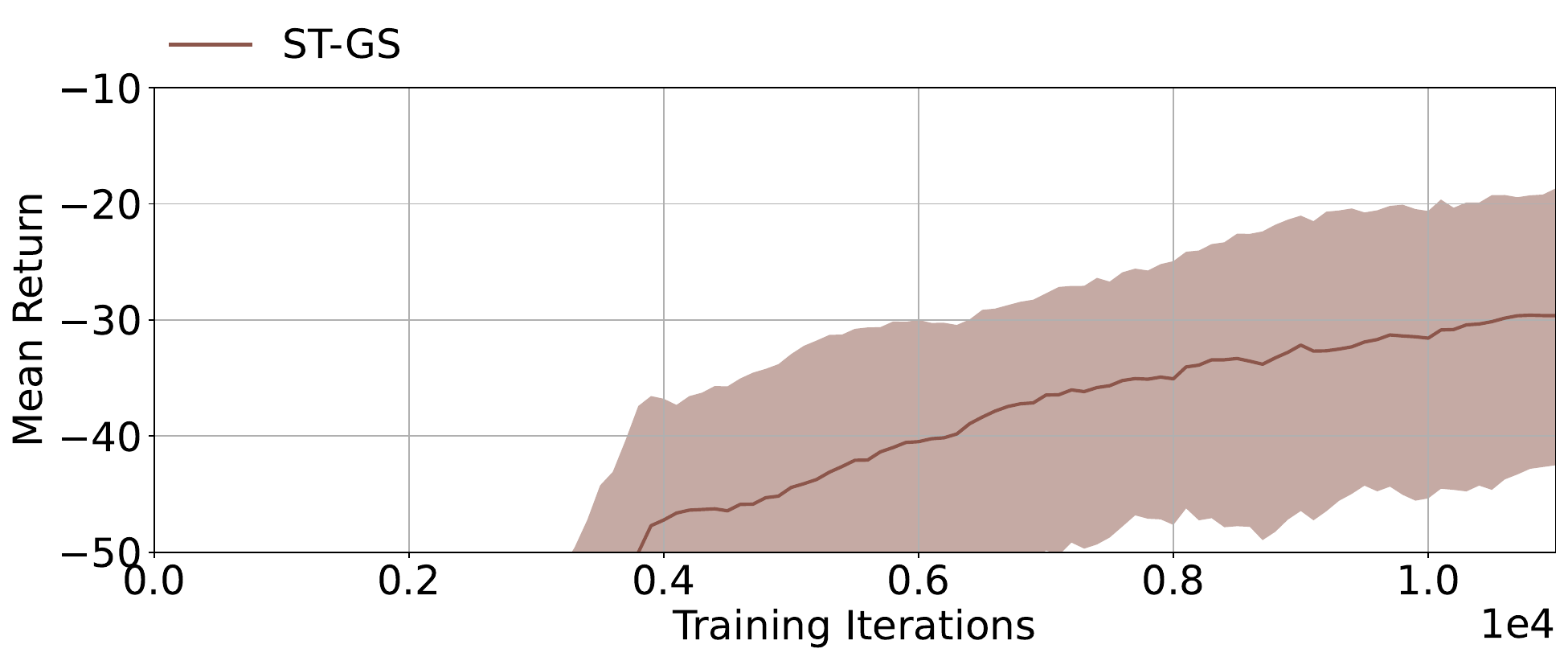}
    \caption{ST-GS in the parallel speaker listener environment}
\end{figure*}

\end{appendices}


\clearpage
\bmhead{Acknowledgments}
This work was supported by the Research Foundation Flanders (FWO) under Grant Number 1S12121N and Grant Number 1S94120N. We gratefully acknowledge the support of NVIDIA Corporation with the donation of the Titan Xp GPU used for this research.

\section*{Declarations}
\bmhead{Data availability}
Data sharing is not applicable to this article as no datasets were generated or analysed during this study.

\bmhead{Conflict of Interest}
The authors have no competing interests to declare that are relevant to the content of this article.
\bibliography{references}


\end{document}